\documentclass[11pt,letterpaper]{article}

\usepackage{microtype}
\usepackage{graphicx}
\usepackage{pdflscape}
\usepackage{booktabs}
\usepackage{longtable}
\usepackage{array}
\usepackage{multirow}
\usepackage{xcolor}
\usepackage{enumitem}
\usepackage{xspace}
\usepackage{float}
\usepackage{placeins}
\usepackage{tikz}
\usetikzlibrary{arrows.meta,positioning,shapes.misc,calc,fit,backgrounds}

\usepackage{survey_singlecolumn}

\usepackage{fontspec}
\usepackage{xeCJK}
\xeCJKsetup{AutoFallBack=true}
\setCJKmonofont{FandolSong-Regular.otf}
\setCJKfallbackfamilyfont{\CJKrmdefault}{HaranoAjiMincho-Regular.otf}

\usepackage{amsmath}
\usepackage{amssymb}
\usepackage{mathtools}
\usepackage{amsthm}
\usepackage[hidelinks]{hyperref}
\hypersetup{
  pdftitle={Brain-to-Language Decoding: Paradigms, Signals, Methods, Evaluation, Practical Use and Beyond},
  pdfauthor={Yiqian Yang; Yiqun Duan; Chenyu Liu; Yiqi Wang; Xinliang Zhou; Chin-Teng Lin; Qingsong Wen; Oiwi Parker Jones; Yu Zhang}
}
\usepackage[capitalize,noabbrev]{cleveref}
\IfFontExistsTF{Amiri-Italic.ttf}{%
  \newfontfamily\arabicfont[Script=Arabic,ItalicFont=Amiri-Italic.ttf]{Amiri-Regular.ttf}%
}{%
  \newfontfamily\arabicfont[Script=Arabic,ItalicFont=Amiri-Slanted.ttf]{Amiri-Regular.ttf}%
}
\newfontfamily\russianfont{DejaVuSerif.ttf}[ItalicFont=DejaVuSerif-Italic.ttf]
\definecolor{c1}{HTML}{2a78d6}
\definecolor{c2}{HTML}{eb6834}
\definecolor{c3}{HTML}{1baf7a}
\definecolor{c4}{HTML}{eda100}
\definecolor{c5}{HTML}{e87ba4}
\definecolor{c6}{HTML}{008300}
\definecolor{c7}{HTML}{4a3aa7}
\definecolor{c8}{HTML}{e34948}
\definecolor{ink}{HTML}{222222}
\definecolor{soft}{HTML}{f2f5fa}
\definecolor{uploadingInk}{HTML}{1A1C1E}

\AddToHookNext{shipout/foreground}{%
  \begin{tikzpicture}[remember picture,overlay]
    \node[anchor=north west,inner sep=0pt] at
      ([xshift=62.531761bp,yshift=-34bp]current page.north west)
      {\includegraphics[width=101bp]{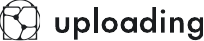}};
    \draw[uploadingInk,line width=0.45bp]
      ([xshift=62.531761bp,yshift=-62.5bp]current page.north west) --
      ([xshift=-62.531761bp,yshift=-62.5bp]current page.north east);
  \end{tikzpicture}%
}

\newcommand{\muecog}{$\mu$ECoG\xspace}
\newcommand{\wer}{WER\xspace}
\newcommand{\nrecords}{938\xspace}

\newcommand{\nidx}{920\xspace}
\newcommand{\nkey}{246\xspace}
\newcommand{\ncandidate}{674\xspace}
\newcommand{\nfulltext}{10\xspace}
\newcommand{\nabstract}{65\xspace}
\newcommand{\ninherited}{39\xspace}
\newcommand{\ntargeted}{132\xspace}
\newcommand{\nrecent}{511\xspace}

\newcommand{\nregistryadded}{125\xspace}
\newcommand{\registryfirstyear}{2008\xspace}
\newcommand{\registrylastyear}{2026\xspace}

\newcommand{\nrefsall}{495\xspace}
\newcommand{\nrefsbody}{364\xspace}
\newcommand{\nrefsappendix}{131\xspace}
\newcommand{\nrefsregistry}{355\xspace}
\newcommand{\nrefscontext}{140\xspace}
\newcommand{\nmethodfigures}{8\xspace}

\newcommand{\nEEG}{503\xspace}

\definecolor{ssvInk}{HTML}{493C36}
\definecolor{ssvRust}{HTML}{A64F34}
\definecolor{ssvOchre}{HTML}{B58A47}
\definecolor{ssvLight}{HTML}{FBF5EC}
\definecolor{ssvPink}{HTML}{F6E5DD}
\definecolor{ssvRule}{HTML}{D8C8B8}
\tikzset{ssv base/.style={font=\rmfamily\fontsize{9.5}{11.5}\selectfont,text=ssvInk},
  ssv box/.style={draw=ssvRule,fill=white,rounded corners=2pt,align=center,minimum height=8mm,inner sep=4pt},
  ssv learned/.style={ssv box,draw=ssvRust,fill=ssvPink},
  ssv reference/.style={ssv box,draw=ssvOchre,fill=ssvLight},
  ssv arrow/.style={-{Stealth[length=2mm]},draw=ssvInk,line width=.65pt},
  ssv supervision/.style={ssv arrow,draw=ssvOchre,dashed},
  ssv heading/.style={font=\rmfamily\bfseries\fontsize{10}{12}\selectfont,anchor=west}}

\newcommand{\SSAlignmentMiniDrawing}{%
\includegraphics[width=\linewidth]{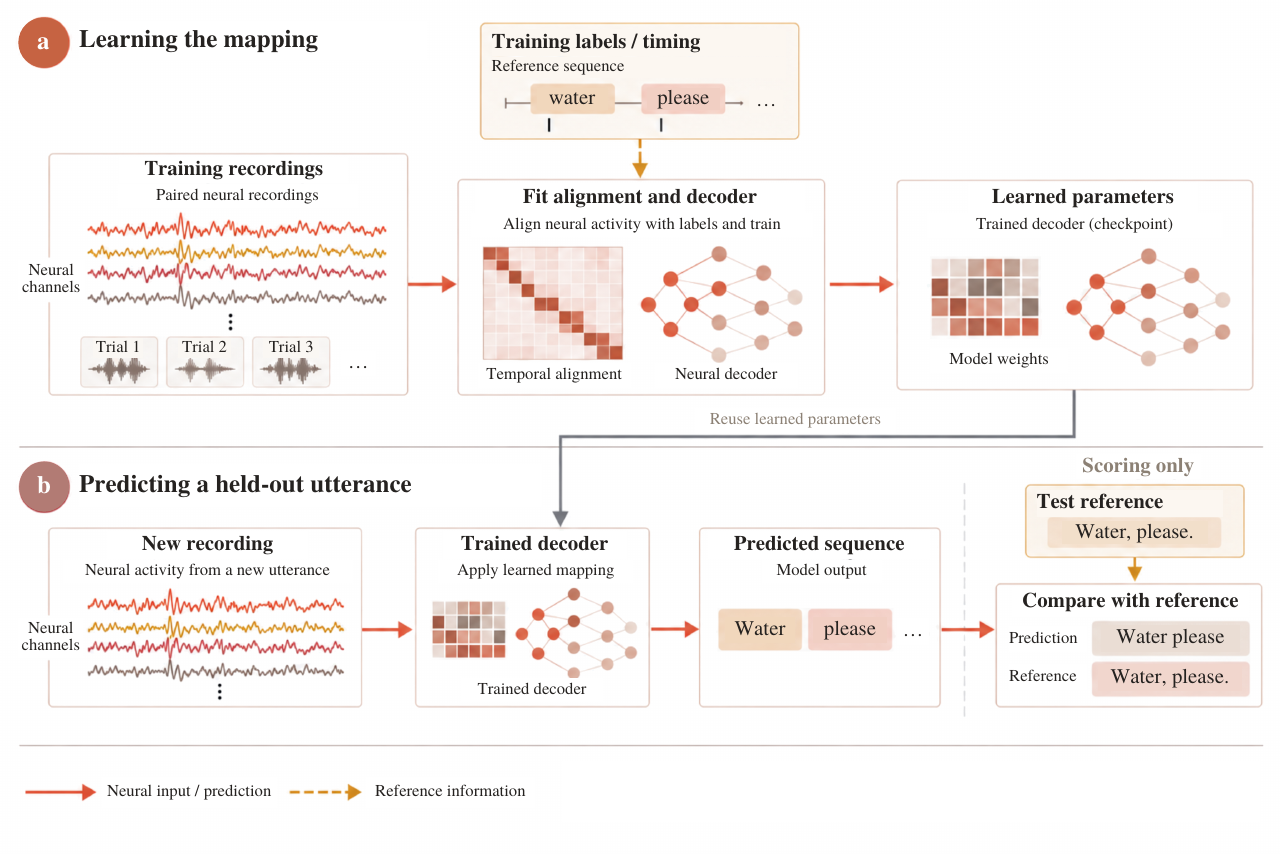}%
}
\newcommand{\SSAlignmentMiniFigure}{%
\begin{figure}[!htbp]\centering\SSAlignmentMiniDrawing
\caption{\textbf{Training supervision and test information occupy different paths.} (a) Training recordings and reference information fit an alignment and decoder; learned parameters are then reused. (b) A new recording produces a sequence, while the held-out reference enters only the scoring step. Iterative alignment discovery illustrates training-time alignment learning \citep{rabbani2024alignment}. Reference-dependent alignment during testing, as in silent-reading reconstruction with an Overt reference, supplies additional information \citep{martin2014s0032}. The illustrated test path excludes that input. Traces, weights and utterances are schematic; protocols must state any test-time references or boundaries.}
\label{fig:supervision_paths}\end{figure}%
}

\newcommand{\SSPretrainingMiniDrawing}{%
\includegraphics[width=\linewidth]{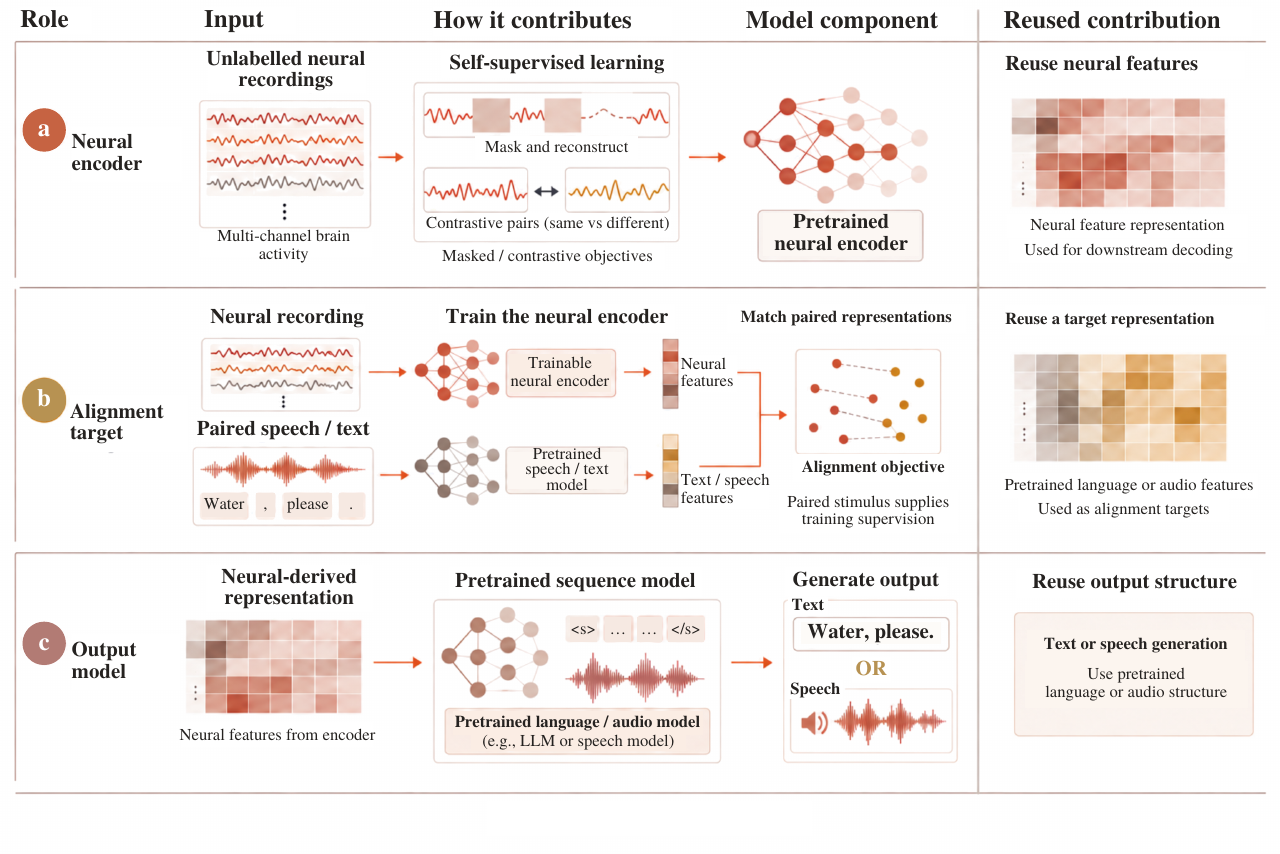}%
}
\newcommand{\SSPretrainingMiniFigure}{%
\begin{figure}[!htbp]\centering\SSPretrainingMiniDrawing
\caption{\textbf{Pretraining contributes at three different interfaces.} (a) Neural self-supervision learns reusable recording structure, as in BENDR and BrainBERT \citep{kostas2021bendr,wang2023brainbert}. (b) A pretrained speech or text model provides targets for aligning a neural encoder \citep{alexandre2023s0322}. The paired stimulus supplies training supervision. (c) An output model provides structure for constructing text or speech from neural evidence, as in language-guided semantic reconstruction \citep{tang2023s0356}. These roles can be combined without requiring the same training data or transfer setting. Feature patterns and messages are illustrative; the right column summarises the contribution being reused.}
\label{fig:pretraining_roles}\end{figure}%
}

\newcommand{\SSHoldoutMiniDrawing}{%
\includegraphics[width=\linewidth]{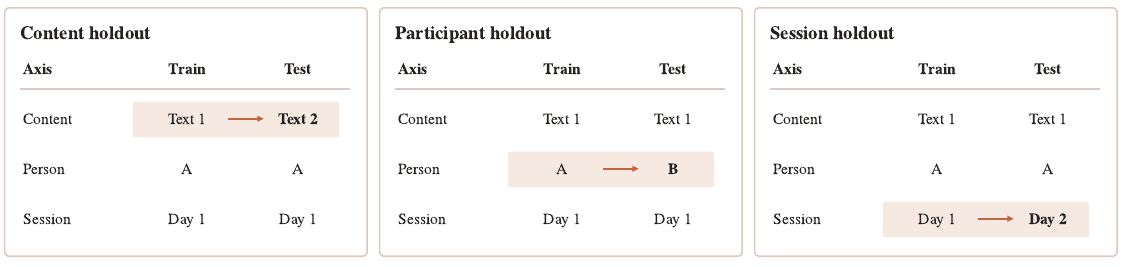}%
}
\newcommand{\SSHoldoutMiniFigure}{%
\begin{figure}[!htbp]\centering\SSHoldoutMiniDrawing
\caption{\textbf{A held-out unit defines the generalisation question.} Illustrative train/test pairs vary content, participant or session while keeping the other attributes familiar. Session indices are local to each participant; holding out person B does not establish future-session transfer. Jointly new participants and text require both forms of separation. Reading split analyses show why participant separation alone can retain familiar sentences \citep{congchi2025s0600}; cross-validation must preserve the independent unit relevant to the intended use \citep{varoquaux2017decoders}. These are different prediction settings rather than stages on a single robustness scale.}
\label{fig:holdout_axes}\end{figure}%
}

\newcommand{\SSAutoregressionMiniDrawing}{%
\includegraphics[width=\linewidth]{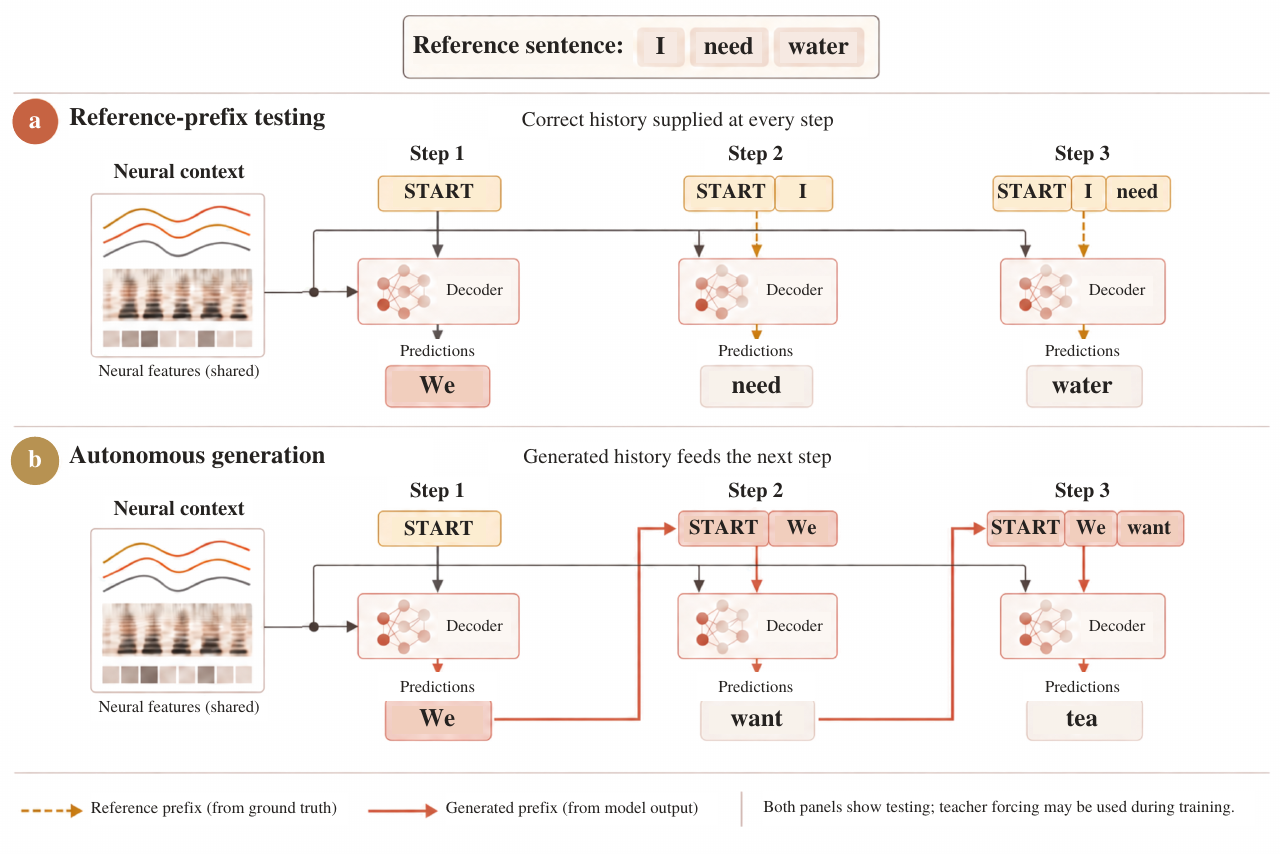}%
}
\newcommand{\SSAutoregressionMiniFigure}{%
\begin{figure}[!htbp]\centering\SSAutoregressionMiniDrawing
\caption{\textbf{The source of the preceding tokens changes what a text-generation test measures.} Both conditions receive neural context. (a) Reference-prefix testing supplies correct preceding tokens even after a prediction error. (b) Autonomous generation uses its own generated history. In the illustrative example, the first prediction, ``We'', enters the next prefix only in (b); subsequent outputs may consequently differ. Teacher forcing may be used during training in either case. The distinction is explicit in the ZuCo noise comparison and the split audit's with/without-teacher-forcing results \citep{jo2025noise,congchi2025s0600}. It concerns autoregressive generation, not every classifier or speech synthesiser.}
\label{fig:test_prefix_information}\end{figure}%
}

\begin{document}
\thispagestyle{plain}
\begin{center}
{\Large\bfseries Brain-to-Language Decoding: Paradigms, Signals, Methods,\\[3pt]
Evaluation, Practical Use and Beyond\par}
\vspace{9pt}
{\normalsize Yiqian Yang$^{1,\ast}$, Yiqun Duan$^{1,\ast}$, Chenyu Liu$^{1}$, Yiqi Wang$^{1}$, Xinliang Zhou$^{1}$,\\[2pt]
Chin-Teng Lin$^{2}$, Qingsong Wen$^{4}$, Oiwi Parker Jones$^{4}$ and Yu Zhang$^{1,3}$\par}
\vspace{5pt}
{\small $^{1}$Uploading Inc\\[1pt]
$^{2}$Human-centric Artificial Intelligence Centre, University of Technology Sydney\\[1pt]
$^{3}$Stanford Institute for Human-Centered Artificial Intelligence (HAI), Stanford University\\[1pt]
$^{4}$Department of Engineering Science, University of Oxford\par}
\vspace{3pt}
{\small \href{mailto:yiqian@uploading.tech}{yiqian@uploading.tech}\quad
\href{mailto:chenyu@uploading.tech}{chenyu@uploading.tech}\quad
\href{mailto:yiqiwang@uploading.tech}{yiqiwang@uploading.tech}\quad
\href{mailto:xinliang@uploading.tech}{xinliang@uploading.tech}\par}
{\small $^{2}$\href{mailto:chin-teng.lin@uts.edu.au}{chin-teng.lin@uts.edu.au}\quad
$^{4}$\href{mailto:qingsongedu@gmail.com}{qingsongedu@gmail.com}\\[1pt]
$^{4}$\href{mailto:oiwi.parkerjones@eng.ox.ac.uk}{oiwi.parkerjones@eng.ox.ac.uk}\quad
$^{3}$\href{mailto:yzhangsu@stanford.edu}{yzhangsu@stanford.edu}\par}
{\small $^{\ast}$Equal contribution. Correspondence: \href{mailto:yiqunduan@uploading.tech}{yiqunduan@uploading.tech}\par}
\end{center}
\vspace{2pt}
\afterpage{\clearpage\noindent
\begin{minipage}{\textwidth}
\captionsetup{type=figure}
\centering
\includegraphics[width=0.97\linewidth]{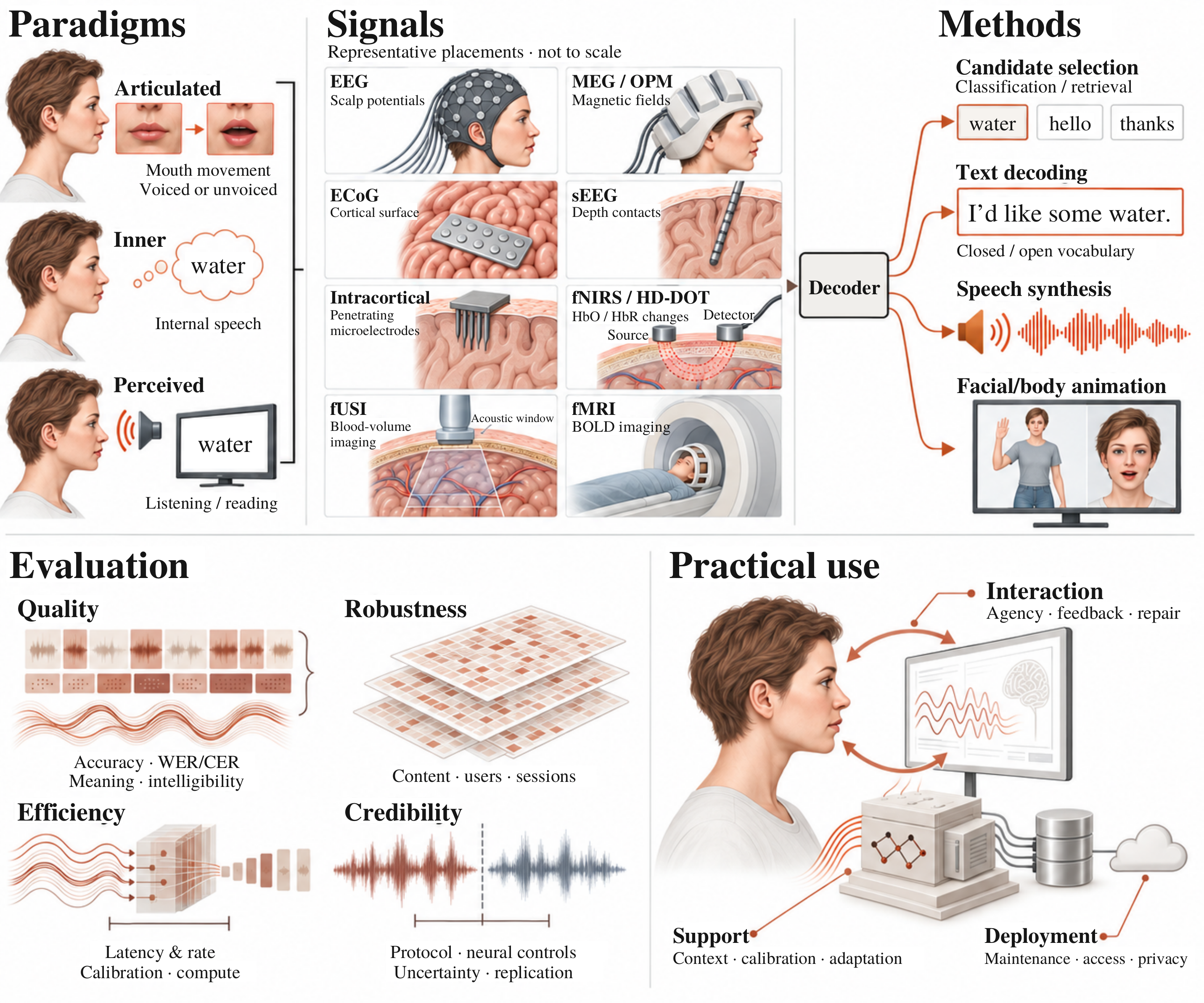}
\caption{\textbf{The survey connects what is decoded with the evidence and mechanisms required for useful communication.} The Paradigms layer groups participant conditions into Articulated, Inner and Perceived; the Signals layer maps the recording families; and the Methods layer connects neural measurements through a decoder to candidate selection, text decoding, speech synthesis and facial animation. Evaluation separates quality, robustness, efficiency and credibility; Practical use addresses interaction, system support and deployment. These layers distinguish successful output generation from evidence of reliable communication. The illustrations are conceptual.}
\label{fig:teaser}
\end{minipage}
\par\medskip}

\begin{abstract}
Brain-to-language decoding translates neural activity associated with language production, internal speech and perception into linguistic or expressive outputs. It offers a route to restoring communication after speech loss and a means of studying how the brain represents language. Advances in neural recording and representation learning have expanded the field from constrained recognition and acoustic reconstruction to text generation, streaming personalised speech and facial animation. This survey synthesises these developments across invasive and non-invasive measurements. We connect Articulated, Inner and Perceived paradigms to the neural populations they engage, the representations available to decoders and the outputs those representations can support. We examine model development, public resources and the evolution of evaluation, and compare published performance and communication costs within their reported protocols. The synthesis identifies complementary routes to progress: phonetic, acoustic and semantic targets preserve different aspects of a message; shared representations support reuse across recording conditions and tasks; and online communication increasingly depends on calibration, feedback and user control alongside decoding accuracy. Shared benchmarks enable algorithmic comparisons, while longitudinal studies reveal the demands of sustained use. We discuss these developments and their remaining limitations, then outline a prospective five-level trajectory from commands and language to meaning, scenarios and bidirectional cognitive exchange.
\end{abstract}

\section{Introduction}
\label{sec:intro}

Brain-to-language decoding seeks to recover linguistic information from recordings of language-related brain activity and translate it into interpretable outputs. It supports communication restoration for people with impaired speech and offers a computational approach to studying neural representations of language. This survey asks which aspects of language different neural measurements can recover, and how those representations can support a person's expression.

Progress spans phoneme recognition during overt speech and, for people with paralysis, sentence-level text decoding and continuous speech synthesis from attempted speech \citep{herff2015s0044,willett2023s0300,littlejohn2025s0496,wairagkar2025s0504}. Complementary non-invasive and inner-speech research investigates the recovery of perceived and internally generated language \citep{hollenstein2018s0778,tang2023s0356,alexandre2023s0322,kunz2025s0568}. Sustained communication nevertheless depends on stable recordings, manageable calibration and adaptation across sessions and operating conditions \citep{card2026s0729}. Advances in recording interfaces, representation learning, pretraining and adaptation address these demands at different stages of the system \citep{silva2024neuroprosthesisreview,dascoli2025s0623,he2026reviewr23}. A text output can represent candidate identification, verbatim transcription or semantic reconstruction, so its interpretation depends on the task and target. A synthesis connecting language activity, recording, decoding and recovered information is therefore needed to explain what each result establishes and how it is achieved.

Existing surveys establish foundations in EEG speech imagery, linguistic decoding and clinical speech interfaces \citep{rahman2024reviewr02,zhang2025reviewr04,murad2024review,wang2025linguisticreview,stavisky2025reviewr07,belfrouh2026reviewr21}. Recent reviews distinguish imagined-speech targets from output pathways \citep{sensors2026review} or integrate intracranial language BCIs from neural mechanisms to clinical deployment \citep{he2026reviewr23}. Our synthesis brings these perspectives into a common comparison across task paradigms and invasive and non-invasive recordings, relating supervision and test-time information to recoverable content, transfer and practical use. \Cref{tab:survey_comparison} positions the scope of this survey alongside prior reviews across paradigms, neural signals, evaluation and practical use, and \cref{sec:priorreviews} discusses the closest perspectives in more detail.

\begin{table}[!htbp]
\centering
\caption{Comparison with related surveys on brain-to-language decoding.}
\label{tab:survey_comparison}
\newcommand{\surveyyes}{\(\checkmark\)}
\newcommand{\surveypart}{\(\triangle\)}
\newcommand{\surveyno}{\textemdash}
\newcommand{\surveyhead}[1]{\rotatebox{90}{#1}}
\begingroup
\fontsize{9}{11}\selectfont
\setlength{\tabcolsep}{0pt}
\renewcommand{\arraystretch}{1.35}
\begin{tabular}{@{}p{65pt}>{\centering\arraybackslash}p{20pt}@{\hspace{4pt}}>{\centering\arraybackslash}p{24pt}>{\centering\arraybackslash}p{24pt}>{\centering\arraybackslash}p{24pt}@{\hspace{4pt}}>{\centering\arraybackslash}p{24pt}>{\centering\arraybackslash}p{24pt}>{\centering\arraybackslash}p{24pt}>{\centering\arraybackslash}p{24pt}>{\centering\arraybackslash}p{24pt}>{\centering\arraybackslash}p{24pt}>{\centering\arraybackslash}p{24pt}>{\centering\arraybackslash}p{24pt}@{\hspace{4pt}}>{\centering\arraybackslash}p{24pt}>{\centering\arraybackslash}p{24pt}>{\centering\arraybackslash}p{24pt}>{\centering\arraybackslash}p{24pt}@{\hspace{4pt}}>{\centering\arraybackslash}p{24pt}@{}}
\toprule
& & \multicolumn{3}{c}{Paradigms} & \multicolumn{8}{c}{Signals} & \multicolumn{4}{c}{Evaluation} & \\
\cmidrule(lr){3-5}\cmidrule(lr){6-13}\cmidrule(lr){14-17}
Survey & Year & \surveyhead{Articulated} & \surveyhead{Inner} & \surveyhead{Perceived} & \surveyhead{EEG} & \surveyhead{MEG} & \surveyhead{ECoG} & \surveyhead{sEEG} & \surveyhead{Intracortical} & \surveyhead{fMRI} & \surveyhead{fNIRS} & \surveyhead{fUSI} & \surveyhead{Quality} & \surveyhead{Robustness} & \surveyhead{Efficiency} & \surveyhead{Credibility} & \surveyhead{Practical use} \\
\midrule
\href{https://doi.org/10.3389/fnins.2016.00429}{Herff \& Schultz} & 2016 & \surveyyes & \surveypart & \surveypart & \surveyyes & \surveyyes & \surveyyes & \surveyno & \surveyyes & \surveyyes & \surveyyes & \surveyno & \surveypart & \surveypart & \surveypart & \surveypart & \surveypart \\
\href{https://doi.org/10.1007/s11571-024-10167-0}{Rahman et al.} & 2024 & \surveyno & \surveyyes & \surveyno & \surveyyes & \surveyno & \surveyno & \surveyno & \surveyno & \surveyno & \surveyno & \surveyno & \surveypart & \surveypart & \surveypart & \surveyno & \surveypart \\
\href{https://doi.org/10.1038/s41583-024-00819-9}{Silva et al.} & 2024 & \surveyyes & \surveypart & \surveypart & \surveypart & \surveypart & \surveyyes & \surveyyes & \surveyyes & \surveypart & \surveyno & \surveyno & \surveyyes & \surveyyes & \surveyyes & \surveyyes & \surveyyes \\
\href{https://doi.org/10.1109/TIM.2024.3472830}{Tang et al.} & 2024 & \surveyyes & \surveyyes & \surveyno & \surveyyes & \surveyyes & \surveyyes & \surveyyes & \surveypart & \surveyyes & \surveyyes & \surveyno & \surveypart & \surveypart & \surveypart & \surveyno & \surveyyes \\
\href{https://doi.org/10.1002/wcs.70016}{Almufareh et al.} & 2025 & \surveypart & \surveyyes & \surveypart & \surveyyes & \surveypart & \surveypart & \surveyno & \surveyno & \surveyyes & \surveypart & \surveyno & \surveypart & \surveypart & \surveypart & \surveypart & \surveyyes \\
\href{https://doi.org/10.1109/TCDS.2024.3462452}{Murad \& Rahimi} & 2025 & \surveypart & \surveyno & \surveyyes & \surveyyes & \surveypart & \surveypart & \surveyno & \surveyno & \surveypart & \surveyno & \surveyno & \surveypart & \surveypart & \surveypart & \surveyno & \surveyyes \\
\href{https://doi.org/10.3390/app15010392}{Qiu et al.} & 2025 & \surveyyes & \surveypart & \surveyyes & \surveyyes & \surveypart & \surveyyes & \surveyno & \surveypart & \surveypart & \surveypart & \surveyno & \surveypart & \surveypart & \surveypart & \surveyno & \surveyyes \\
\href{https://doi.org/10.1146/annurev-bioeng-110122-012818}{Stavisky} & 2025 & \surveyyes & \surveypart & \surveypart & \surveypart & \surveypart & \surveyyes & \surveyyes & \surveyyes & \surveypart & \surveyno & \surveyno & \surveyyes & \surveyyes & \surveyyes & \surveyyes & \surveyyes \\
\href{https://doi.org/10.7717/peerj-cs.2938}{Su \& Tian} & 2025 & \surveypart & \surveyyes & \surveypart & \surveyyes & \surveypart & \surveyno & \surveyno & \surveyno & \surveypart & \surveypart & \surveyno & \surveypart & \surveyno & \surveypart & \surveypart & \surveypart \\
\href{https://doi.org/10.1088/1741-2552/ade28e}{Tates et al.} & 2025 & \surveyno & \surveyyes & \surveyno & \surveyyes & \surveyno & \surveyyes & \surveyyes & \surveypart & \surveypart & \surveyyes & \surveyno & \surveypart & \surveypart & \surveyyes & \surveyyes & \surveyyes \\
\href{https://doi.org/10.1038/s42003-025-08511-z}{Wang et al.} & 2025 & \surveyyes & \surveyyes & \surveyyes & \surveyyes & \surveyyes & \surveyyes & \surveypart & \surveyyes & \surveyyes & \surveyno & \surveyno & \surveyyes & \surveypart & \surveypart & \surveypart & \surveypart \\
\href{https://doi.org/10.1016/j.neucom.2025.130928}{Yu et al.} & 2025 & \surveyyes & \surveypart & \surveyyes & \surveyyes & \surveyyes & \surveyyes & \surveyno & \surveyno & \surveyyes & \surveyno & \surveyno & \surveypart & \surveypart & \surveypart & \surveypart & \surveyyes \\
\href{https://doi.org/10.1109/TCDS.2024.3431224}{Zhang et al.} & 2025 & \surveypart & \surveyyes & \surveyno & \surveyyes & \surveyno & \surveypart & \surveyno & \surveyno & \surveyno & \surveypart & \surveyno & \surveyyes & \surveyyes & \surveypart & \surveyyes & \surveyyes \\
\href{https://doi.org/10.1186/s12911-026-03552-8}{Belfrouh et al.} & 2026 & \surveyyes & \surveyyes & \surveypart & \surveyyes & \surveyyes & \surveyyes & \surveyyes & \surveyyes & \surveypart & \surveypart & \surveyno & \surveyyes & \surveyyes & \surveypart & \surveyyes & \surveyyes \\
\href{https://arxiv.org/abs/2603.12279v2}{He et al.}\textsuperscript{*} & 2026 & \surveyyes & \surveyyes & \surveypart & \surveypart & \surveypart & \surveyyes & \surveyyes & \surveyyes & \surveyno & \surveyno & \surveyno & \surveyyes & \surveyyes & \surveyyes & \surveyyes & \surveyyes \\
\href{https://doi.org/10.3390/s26103212}{Zhang et al.} & 2026 & \surveypart & \surveyyes & \surveypart & \surveyyes & \surveypart & \surveypart & \surveypart & \surveyno & \surveypart & \surveypart & \surveyno & \surveyyes & \surveyyes & \surveyno & \surveyyes & \surveypart \\
\href{https://doi.org/10.1016/j.csl.2026.102020}{Sbaih et al.}\textsuperscript{\textdaggerdbl} & 2027 & \surveyyes & \surveyyes & \surveyyes & \surveyyes & \surveyno & \surveyyes & \surveyyes & \surveypart & \surveypart & \surveyno & \surveyno & \surveyyes & \surveyyes & \surveypart & \surveyyes & \surveyyes \\
\midrule
\textbf{This survey} & {} & \surveyyes & \surveyyes & \surveyyes & \surveyyes & \surveyyes & \surveyyes & \surveyyes & \surveyyes & \surveyyes & \surveyyes & \surveyyes\textsuperscript{\textdagger} & \surveyyes & \surveyyes & \surveyyes & \surveyyes & \surveyyes \\
\bottomrule
\end{tabular}
\endgroup
\par\vspace{5pt}
\begingroup
\fontsize{8.5}{10.2}\selectfont\raggedright
\surveyyes\ substantive synthesis; \surveypart\ limited discussion; \surveyno\ absent or mentioned only.
Articulated denotes executed articulation; Inner, imagery without attempted execution; Perceived, decoding externally presented language.
Attempts without confirmed articulation are not assigned to Articulated or Inner.
Evaluation rates the synthesis of assessment practices: Quality, output measures; Robustness, generalisation and stability; Efficiency, rate, latency and costs; Credibility, controls, statistical support and reproducibility.
Practical use covers interaction, adaptation and deployment.
\textsuperscript{*}Preprint.
\textsuperscript{\textdaggerdbl}Published online in 2026.
\textsuperscript{\textdagger}fUSI coverage concerns language mapping and feasibility, not established language-content decoding.
\par\endgroup
\end{table}

After defining the review method and scope (\cref{sec:method}), we organise the survey around four connected stages: language activity under a task paradigm, its measurement through a recording interface, neural decoding, and the recovery of language and meaning. This perspective links what the activity represents, what a measurement can capture and how the recorded information supports an output (\cref{fig:routes}).

\begin{enumerate}[leftmargin=*,label=\arabic*.,itemsep=6pt,topsep=6pt]
\item \textbf{Paradigms and neural activity.} We relate Articulated, Inner and Perceived paradigms to the neural representations associated with executed articulation, internal speech simulation and externally presented language, respectively (\cref{sec:scope,sec:signals}). These paradigms change both the information available to a decoder and its supervision: externally presented language supplies content and timing, whereas internal generation requires distinguishing experimental cues from the content being decoded.

\item \textbf{Recording interfaces and neural signals.} We compare the measurement principles, devices and recording configurations of EEG, MEG, ECoG, sEEG, intracortical recordings, fMRI, fNIRS and fUSI, with fUSI represented by language-mapping and feasibility studies (\cref{sec:signals}). Temporal resolution, anatomical coverage and signal quality shape the information that can be observed and constrain subsequent decoding.

\item \textbf{Neural decoding.} We examine how supervision, temporal alignment and representation learning map neural signals to linguistic targets, drawing on public datasets and shared benchmarks (\cref{sec:methods}). Signal properties and available labels shape the decoding route, while pretraining and shared representations provide opportunities for transfer across tasks and recording conditions.

\item \textbf{Language and semantic recovery.} We distinguish phonemic, lexical, sentence, acoustic and semantic targets and the routes through which they support text, speech and audio-linked facial outputs (\cref{fig:routes}). Exact wording, sound characteristics and meaning express different recovery goals; relating them to the preceding stages distinguishes the information recovered from the form used to present it.
\end{enumerate}

\begingroup
\clubpenalty=10000
\widowpenalty=10000
To support this account, we synthesise evaluation through \emph{Quality}, \emph{Robustness}, \emph{Efficiency} and \emph{Credibility}, interpreting reported performance in relation to its task, recording and decoding conditions (\cref{sec:eval}). Our discussion of practical use links interaction, adaptation and deployment to the feedback, correction, calibration effort and recording access needed for sustained communication (\cref{sec:practical}).\par
\endgroup

Together, these connections explain language recovery as a capability jointly shaped by task, measurement, decoding and the intended output. Bringing findings across paradigms, recording techniques and methods into this common account clarifies why capabilities differ and where they apply. This perspective supports method selection and research towards reliable communication by connecting shared limitations (\cref{sec:limits}) to prospective capabilities (\cref{sec:beyond}).

\FloatBarrier

\section{Review Method and Scope}
\label{sec:method}

\subsection{Scope and synthesis}

The empirical scope is neural recording used to recover language-related content in Articulated, Inner or Perceived paradigms. Reading is included as Perceived--Visual language. Clinical attempted-speech studies are retained even when residual articulator movement is not documented; an attempted-speech instruction alone does not assign the condition to Inner speech. We distinguish content decoding from auxiliary state detection, language mapping, representation analysis and dataset reports; adjacent evidence is retained only where it informs acquisition or interpretation. Interfaces based solely on surface electromyography (EMG) or camera-based lip-reading, and spelling-by-selection BCIs, fall outside this neural language-decoding scope. Multimodal systems with neural and peripheral inputs retain explicit input labels.

The unit of comparison is an experimental condition: the participant's task, recording, supervision and output together with the tested content, controls and use setting. Multiple conditions within one paper can support different conclusions. This is a critical narrative synthesis with a documented search and selected structured extractions, rather than an exhaustive full-text review of every indexed record. Registry coverage, structured extraction and claim-level source checking are reported separately.

\paragraph{Comparative approach.} Each comparison asks what was demonstrated, which information enabled it and what would have to change for another task or use. Candidate identification, semantic reconstruction and exact transcription require different tests even when all produce text. Scores are therefore interpreted within their participant, task, supervision and evaluation conditions rather than pooled. Missing controls or operating details remain unknown; they establish neither the presence nor the absence of a capability.

\subsection{Prior reviews and the synthesis developed here}
\label{sec:priorreviews}

\Cref{tab:survey_comparison} compares representative surveys across scope dimensions; \cref{app:recentreviews} lists all 30 directly relevant reviews and 16 contextual sources, with checking depth for each.

Pipeline-oriented reviews organise neural speech recognition, EEG processing and generative decoding around acquisition, representations and models \citep{herff2016review,rahman2024reviewr02,murad2024review}; imagery reviews additionally compare task instructions and linguistic targets \citep{tates2025reviewr11,sensors2026review}. Target granularity and output pathway complement the participant's operation: decoding a word during reading, listening or internal simulation tests different neural conditions even when the output vocabulary is identical.

Clinical reviews connect cortical mechanisms and recording interfaces to communication outcomes and continued use \citep{silva2024neuroprosthesisreview,belfrouh2026reviewr21,dohle2025reviewr45,swanson2026reviewr46}. The integrated intracranial perspective follows this chain from neural mechanisms to clinical deployment \citep{he2026reviewr23}. We extend the comparison across invasive and non-invasive recordings, relating each decoder's supervision and test-time information to the calibration, correction and assistance needed for communication.

\paragraph{Coverage and dates.} The search has no lower publication-year limit. Database retrieval closed on 15 September 2026; source-led additions and the registry were updated through 21 September. A targeted search on 10 October added primary sources on paradigm-specific neural processing for \cref{sec:paradigms}, outside the registry counts. Earlier foundational studies and reviews are included when they explain a relevant mechanism, recording choice or evaluation practice. The current registry spans \registryfirstyear--\registrylastyear, describing retrieved contents rather than an eligibility restriction. The scope covers Articulated, Inner and Perceived paradigms and invasive and non-invasive recordings. Of \nidx{} canonical entries, \nkey{} have selected structured extractions; registry size measures retrieval breadth, not full-text inclusion in a systematic synthesis.

\subsection{Search and study selection}

\paragraph{Retrieval.} Our reporting is informed by PRISMA-S \citep{page2021prismas} and the PRISMA 2020 distinction between records, reports and studies \citep{page2021prisma2020}; the review was neither preregistered nor screened independently in duplicate. Twelve Europe PMC and four OpenAlex query groups crossed signal synonyms with speech, phoneme, vowel, word and language terms and with decode, recognise, classify, synthesise, reconstruct and neuroprosthesis terms; separate queries targeted \emph{unspoken}, \emph{attempted}, \emph{watched} and \emph{functional ultrasound}. All query pages were exhausted. Forward citation tracking started from 16 seed reviews and key papers; backward tracking used prior surveys and fUSI primary-literature references, with DOIs completed through Crossref. A retrieval check used 44 studies linked to the earlier study list by the workbook's \texttt{Old\_survey\_ID}: 43 have database-query or citation-tracking provenance and one came from the ISCA publisher archive. This curated set is a coverage check, not a recall estimate.

\paragraph{Screening and coding.} Database queries retrieved 28{,}515 records, including cross-query duplicates; citation tracking and DOI/title matching produced 20{,}335 records. The screening log records automatic rule-assisted triage and curated retentions under the scope above. The workbook prioritises language-content decoding, speech-state detection and related datasets; other categories flag reviews, repository copies, corrections, non-target inputs or outputs and ultrasound-search noise. Screening retained 774 entries, mostly title-level candidates: 616 rule-prioritised and 158 curated retentions, including the 44 earlier-list studies. Rule exclusions are automatic outcomes, not manual or dual-reviewed decisions. Author tracing added three records (777 total), and a reading-route supplement added 36 (813 stable IDs); all three author-traced studies and 23 reading-route additions were among the 20{,}335 records but not retained, showing that the rules missed reports later included in the survey. Both supplements kept the original first-publication cut-off. Linking 18 duplicate reports and publication versions yielded the original 795 canonical entries; the source-led supplement added \nregistryadded{} eligible report/resource entries, bringing the registry to \nidx{}. Archived triage and supplements document recovered omissions, not screening recall. The original search, screening categories, supplements and version links appear in \cref{app:search}; archived aliases preserve their notes. Canonical entries need not represent independent participant cohorts.

\paragraph{Recent-review supplement.} A separate search on 18 September 2026 targeted reviews available from 2024 onwards, combining review terms with language tasks, speech neuroprostheses and recording modalities. Three fully paginated Europe PMC queries returned 3,970 hits and 3,942 unique records; fifteen Crossref queries, limited to their top 75 results, yielded 921 unique DOIs. Publisher and repository searches and backward tracking supplemented discovery. Screening titles, abstracts and targeted sections produced a 73-entry inventory: 30 directly relevant reviews, 37 adjacent reviews, two chapters, two perspectives and two candidates. These review-only records do not increase the primary-study counts. Exact queries, source links, publication status and checking depth are retained in the supplement.

\begin{figure*}[!htbp]
\centering
\includegraphics[width=\textwidth]{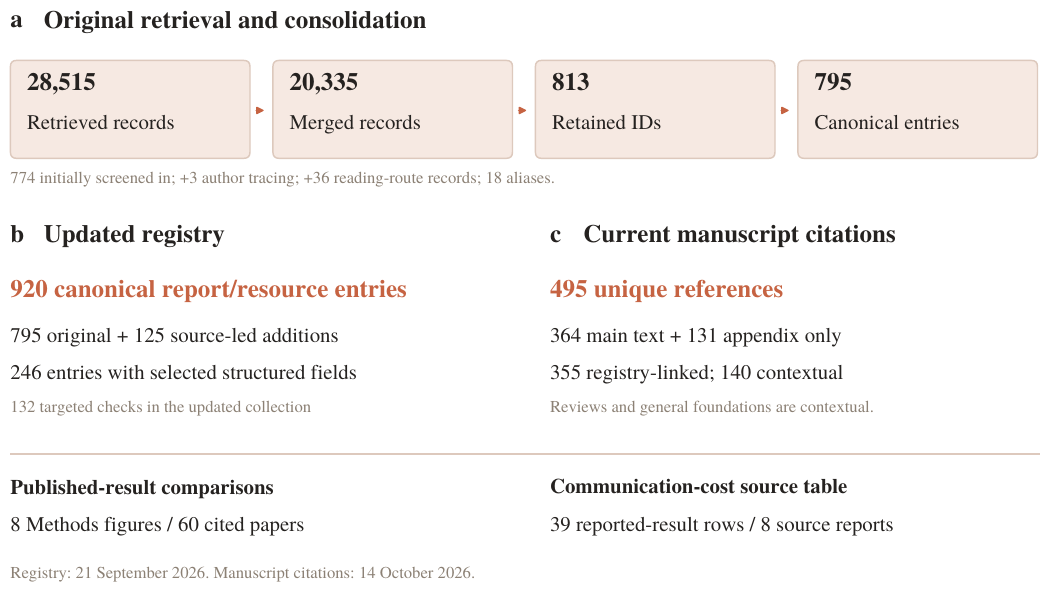}
\caption{\textbf{Literature retrieval, the updated registry and manuscript citation coverage.} (a) The original retrieval and supplements produced 813 retained IDs and 795 canonical entries after publication-version consolidation. (b) Source-led additions bring the registry to \nidx{} report/resource entries, including \nkey{} with selected structured fields. (c) The active manuscript cites \nrefsall{} distinct references: \nrefsbody{} in the main text and \nrefsappendix{} only in appendices; \nrefsregistry{} are linked to registry entries and \nrefscontext{} provide contextual reviews, foundations or adjacent evidence. The footer counts citations in the \nmethodfigures{} Methods comparison figures separately from the communication-cost result rows. References, reports, resources, experimental conditions and participants are different counting units. Publication versions are linked; companion data releases and articles can describe the same acquisition.}
\label{fig:search_flow}

\end{figure*}

\paragraph{Thematic primary-source supplement.} On 19--21 September 2026, source-led searches examined all 47 subsections and subsubsections from Paradigms through Limitations, together with the five prospective Beyond targets. Three research agents covered paradigms/signals, methods, and evaluation/use; each recorded topic-specific discovery and focused follow-up queries, source locations and reading depth. Backward and forward source-led searches sought foundational work, independent approaches and recent developments and updated the synthesis through 21 September 2026. Overlapping reports and publication versions were reconciled before integration. The first thematic pass added 48 cited sources; subsequent dataset, method and evaluation searches extended this set. Sources include neural-language reports and resources as well as methodological foundations and contextual BCI research. A source-led update on 20 September added targeted searches for neural language representations, generative MEG-to-text, shared decoding benchmarks and reported communication costs. Eligible reports and citable resources have been backfilled into the registry, with source locations and coding provenance preserved. Reviews, generic metrics and contextual user studies remain references rather than additions to the primary report/resource census. A unified citation ledger links each active reference to its registry identity where applicable (\cref{fig:search_flow}). Quantitative comparison tables retain the original paper, test condition, metric and source location; their figures reproduce reported results without new decoder experiments. Targeted citation tracking supplements the database search but is not dual independent human screening.

\subsection{Evidence extraction and source verification}

\paragraph{Structured extraction.} The \nkey{} selected entries contain task, recording, output or resource fields used in the synthesis. Their recorded provenance comprises \nfulltext{} full-text checks, \nabstract{} abstract checks, \ninherited{} inherited extractions with metadata rechecked, and \ntargeted{} targeted source-field checks. Targeted checks cover both newly added sources and previously indexed reports examined in greater detail. These categories describe the recorded depth of source use; checking selected fields does not amount to a whole-paper risk-of-bias appraisal. The remaining \ncandidate{} entries retain bibliographic and available label information. Selection was neither random nor a prespecified risk-of-bias stratum.

\paragraph{Claim-level use and remaining gaps.} Abstract-derived descriptions are treated as author-reported claims. The source-check table in \cref{tab:claimledger} separately states source locations and supporting evidence for selected conclusions; checking one claim does not verify every result in its paper or its complete methodological description. Cohort links are retained where known, with missing links treated as unknown rather than as proof of independent participants. The completed database retrieval used Europe PMC and OpenAlex, complemented by publisher, preprint and conference sources and source-led citation tracking. Native-platform access attempts on 20 September did not produce exportable Scopus, Web of Science, IEEE Xplore or ACM DL search results; the access outcomes are recorded with the queries. These platforms are therefore not counted as completed searches. Non-English coverage and citation tracking are incomplete. The documented search supports a broad critical synthesis, without establishing exhaustive retrieval.

\subsection{Corpus coverage}

\begin{figure*}[!htbp]
\centering
\includegraphics[width=\textwidth]{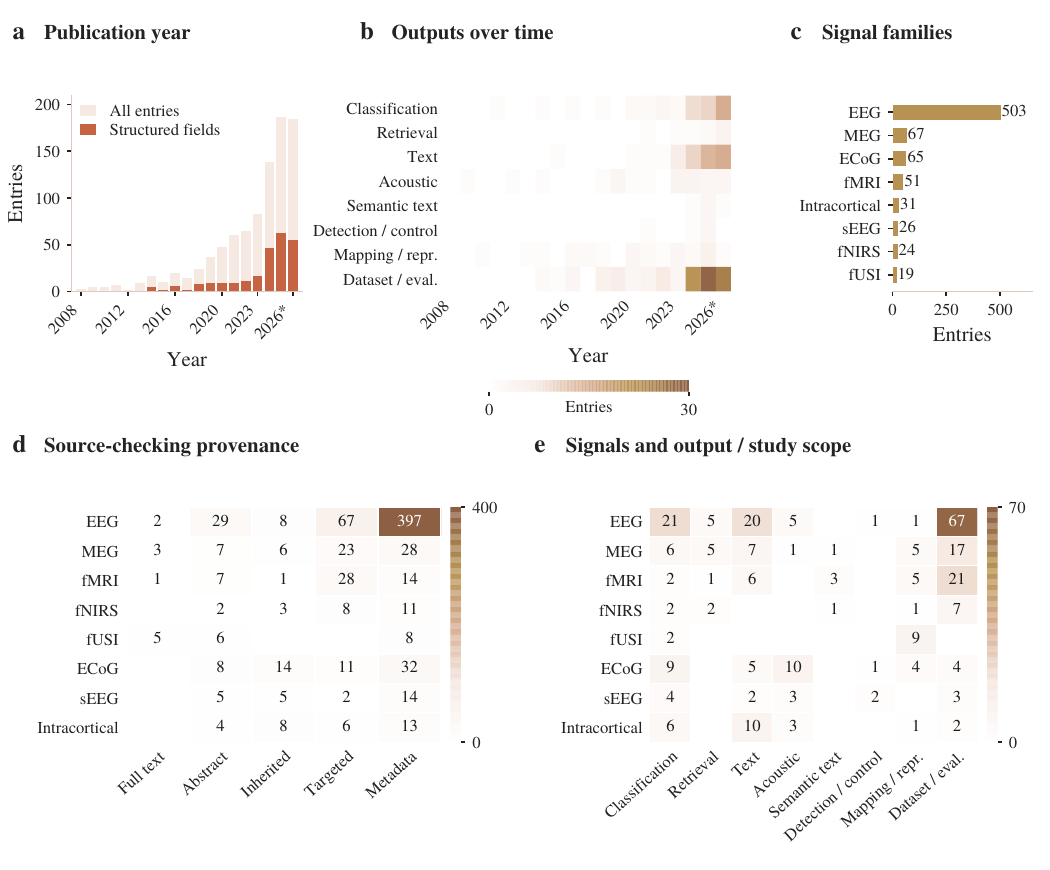}
\caption{\textbf{Publication years, signal families and output coverage in the updated registry.} (a) Annual counts of canonical report/resource entries, with structured fields highlighted. (b) Output and study-scope labels by year among the extracted entries. (c) Signal-family counts across the registry; ECoG includes micro-ECoG, fNIRS includes HD-DOT, and intracortical includes spike recordings. (d) Source-checking provenance by signal: full-text, abstract, inherited extraction, targeted source-field checks and bibliographic metadata. (e) Signal--output and signal--study-scope pairs among the extracted entries. The registry is updated through 21 September 2026; the asterisk marks a partial year. Heatmaps use separate count scales and blank cells denote zero assigned pairs. Labels overlap, and companion articles or data deposits may share acquisitions. These distributions describe the indexed collection, including adjacent evidence, rather than independent cohorts or field-wide success rates.}
\label{fig:landscape}
\end{figure*}

\paragraph{Corpus shape.} \Cref{fig:landscape} summarises the \nidx{} canonical entries by publication year, signal family and source-checking provenance, and the extracted entries by output scope. EEG is the most frequent signal label (\nEEG{} entries), and \nrecent{} entries date from 2024--2026. Labels are multi-valued and unassigned fields are not negative findings, so these counts describe the indexed collection rather than independent cohorts or the prevalence of approaches in the field.

\paragraph{Data availability.} The accompanying workbook and study exports contain the search queries, screening decisions, study descriptions and linked publication versions. The exports also identify where each study is discussed in the manuscript. The \href{https://github.com/Uploading-Inc/Neural-to-Text-Research/blob/main/silent_speech/paper/evidence/comprehensive_revision_20260921/ARTIFACT_INDEX.md}{artifact index} links the current workbook, citation ledger, source-value tables and search records. The reference list gives the cited works, and \cref{app:search} summarises the search and targeted supplements.

\section{Paradigms}
\label{sec:scope}

\label{sec:paradigms}

\subsection{Definitions, nomenclature and intended users}

Brain-to-language paradigms elicit language through executed articulation, internal speech simulation or perception of an external input. Production connects decoding to voluntary expression, while controlled presentation supplies content and timing for studying neural representations \citep{moses2021s0220,willett2023s0300,tang2023s0356,hollenstein2018s0778}.

Prospective users help determine which tasks matter. Interviews with people with locked-in syndrome identified communication, computer access and environmental control as desired applications, alongside preferences for active control; surveys of people with amyotrophic lateral sclerosis examined setup, training and interface requirements \citep{branco2021preferences,huggins2011preferences}. These findings make initiation, message choice and manageable operating effort central design considerations. Requirements vary across populations and available movements, so an experimental vocabulary should be chosen in relation to a communication goal.

We organise study-specific task names by the participant's operation during the decoded interval (\cref{tab:paradigms}); one study can therefore contribute more than one paradigm.

\begin{table}[!htbp]
\caption{Task names and procedures organised into three paradigms and nine subtypes. Assignment follows the operation during the decoded interval. Limited requires residual articulation; Mixed requires an explicitly combined imagery strategy.}
\label{tab:paradigms}
\vskip 0.1in
\centering\small
\begin{tabular}{@{}p{2.4cm}p{2.7cm}p{10.6cm}@{}}
\toprule
Paradigm & Subtype & Names and defining operation \\
\midrule
Articulated & Overt & Overt, vocalised or spoken speech; reading aloud, repetition and audible whispering. Articulation is executed and speech is audible. \\
 & Mouthed & Mouthing, mimed speech and silent articulation: executed speech movements without an audible utterance. Covert or subvocal conditions map here only when articulation is documented. \\
 & Limited & Impaired or dysarthric production and speech attempts with documented residual articulation. Audibility and voicing are separate descriptors. \\
\midrule
Inner & Articulatory & Articulatory imagery, motoric inner speech and imagined speaking: simulate speaking movements or sensations without trying to execute them. \\
 & Auditory & Hearing imagery and first- or third-person auditory inner speech: imagine one's own or another person's voice. \\
 & Mixed & Explicit combination of articulatory and auditory imagery. An unspecified imagery strategy does not establish Mixed. \\
\midrule
Perceived & Auditory & Listening, heard speech and auditory word or narrative perception. Attending to a speaker qualifies the listening condition. \\
 & Visual & Written-language Reading: careful comprehension, skimming, targeted search or spell checking; accompanying inner verbalisation is recorded separately. Lip-reading: watching visible speech without sound. \\
 & Audiovisual & Joint perception of heard speech and visible articulation, including audiovisual speech recognition. \\
\midrule
\multicolumn{3}{@{}p{15.7cm}@{}}{Inner, internal, imagined and speech imagery identify Inner only when the operation is internal simulation; covert, silent and subvocal labels require their execution conditions. A movement-free motor attempt remains explicitly attempted speech, not Inner imagery. Naming, repetition, rehearsal, cued or self-chosen content, and phoneme, word or sentence targets combine with the paradigms rather than adding groups.}\\
\bottomrule
\end{tabular}
\end{table}

\begin{figure*}[!htbp]
\centering
\includegraphics[width=0.80\textwidth]{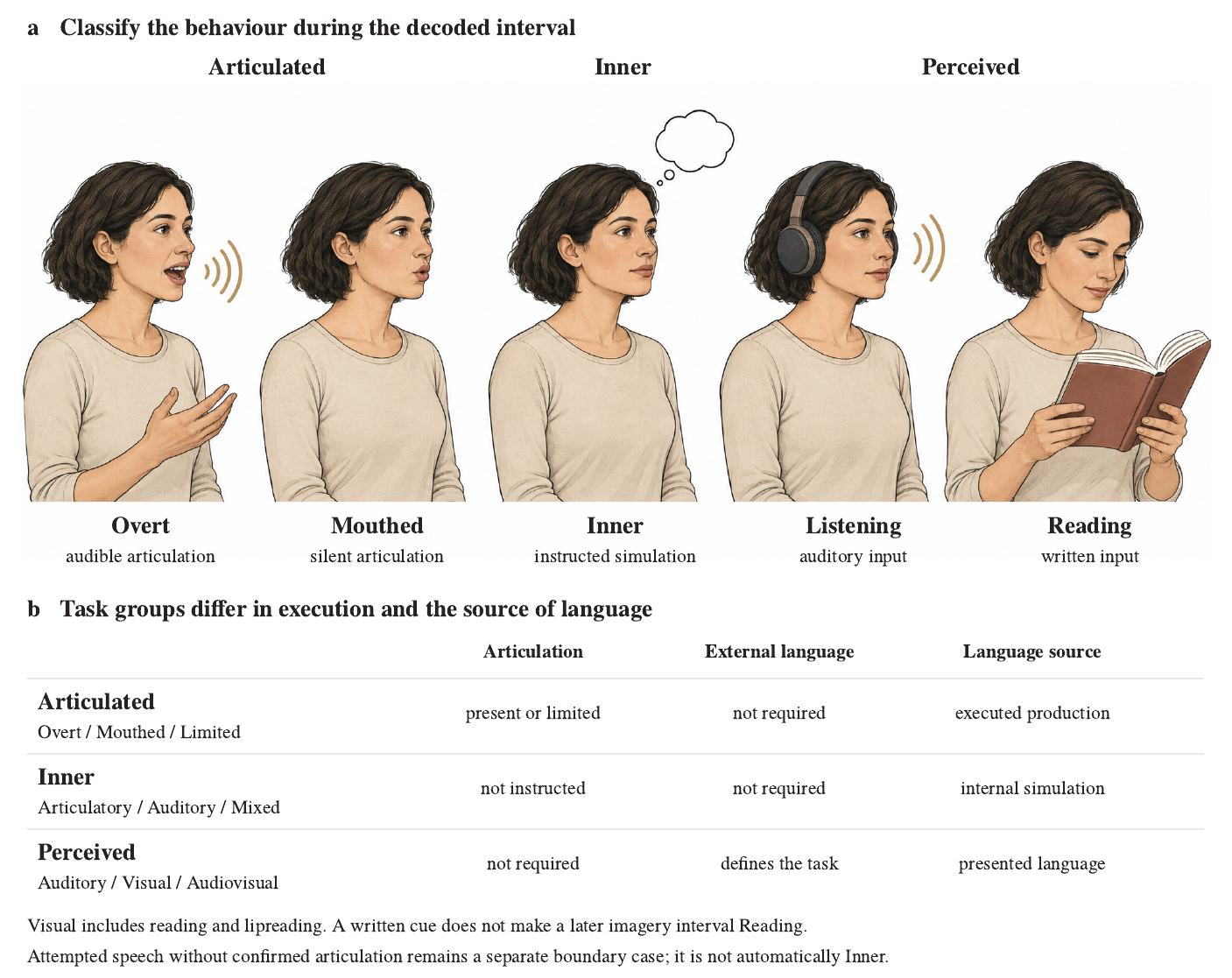}
\caption{\textbf{Task conditions distinguish producing language, simulating speech and perceiving a linguistic input.} Overt and Mouthed involve executed articulation; Inner involves instructed internal simulation; listening and reading provide external language. The lower panel compares the three groups by execution and content source. Clinical attempts without confirmed residual articulation remain explicitly qualified. Instructions alone do not establish absence of muscle activity; a written cue before imagery does not make that later interval a Reading task.}
\label{fig:residual}
\label{fig:taxonomy}
\end{figure*}

The neural distinction concerns both the activity elicited and the information it carries. Utterance planning and execution, internally reinstated speech and sensory input engage partially overlapping language and sensorimotor systems \citep{dataset_cai2026_6c6ce,kunz2025s0568,karthik2024s0391}. In \cref{fig:paradigm_neural_paths,fig:inner_neural_paths,fig:perceived_neural_paths}, each box names a brain region and its operation; labelled arrows specify the information and direction within the stated model. Task and bodily inputs/outputs remain outside brain-region boxes. Panel labels identify models informed by directed analyses, authors' candidate routes or survey-level syntheses; regional encoding and timing results appear as separate observations. Coloured identifiers link each region to the anatomy. Left and right lateral views appear above whole-brain dorsal and ventral views, with anatomical left at screen right in the ventral view.

\subsection{Articulated: Overt, Mouthed and Limited}

\label{sec:articulated}\label{sec:overt}\label{sec:mouthed}\label{sec:attempted}

\paragraph{Operation.} Articulated requires speech-articulator movement; its subtypes distinguish audibility and production limits (\cref{tab:paradigms}). Motor intention can persist without documented residual articulation, so these clinical conditions remain qualified as attempted speech rather than imagery \citep{kunz2025s0568,jude2026s0677}.

\paragraph{Planning, execution and feedback.} Preparing an utterance recruits populations distinguishable from those most responsive during articulation. Interactive question--answer tasks localise planning responses principally to caudal inferior and middle frontal cortex, whereas production responses concentrate in sensorimotor cortex \citep{castellucci2022planning}. Stimulation of these caudal frontal sites can disrupt planning during interaction \citep{castellucci2024cortical}. During natural speech, posterior middle frontal neurons show successive pre-utterance peaks for morphological, phonemic and syllabic information \citep{khanna2024s0467}. In cued pseudoword repetition, syllable-frame information becomes decodable before individual phonemes \citep{duraivel2026planning}. Recordings and stimulation also implicate middle precentral cortex in speech sequencing \citep{liu2025sequencing}.

\begin{figure*}[p]
\centering
\captionsetup{font=footnotesize}
\includegraphics[width=\linewidth]{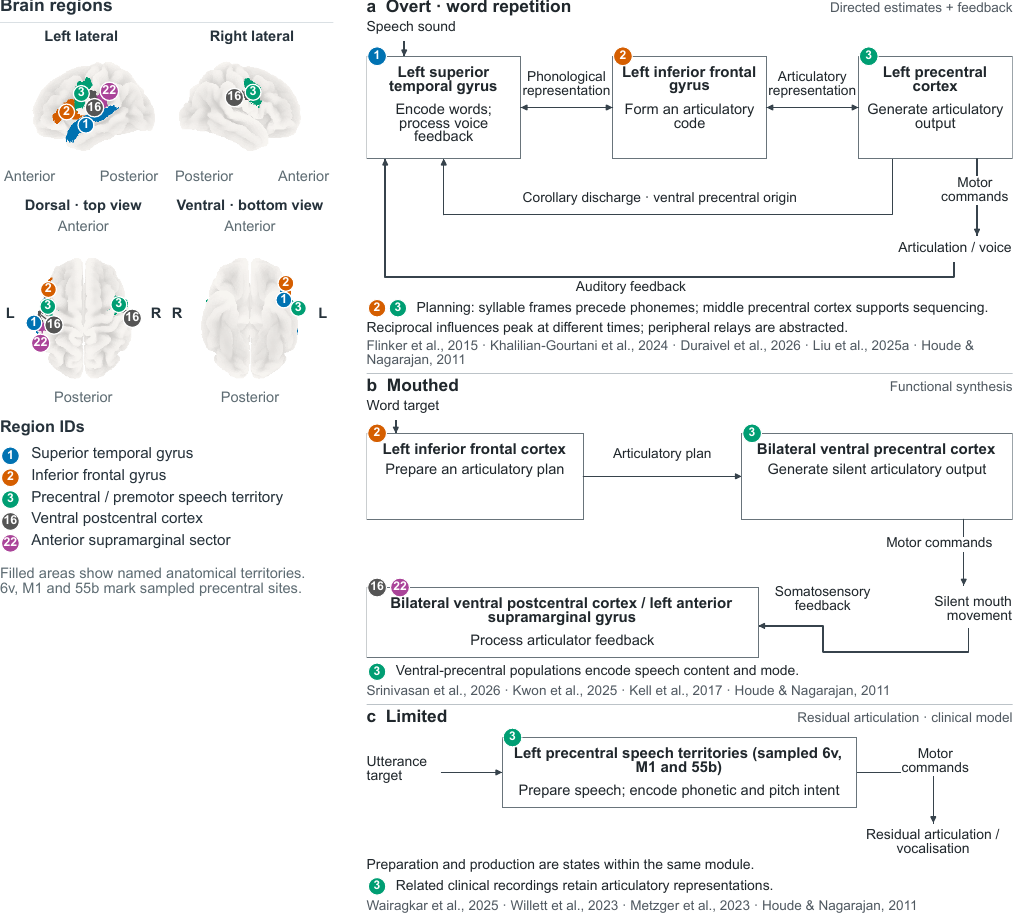}
\caption{\textbf{Articulated speech: cortical information flow, motor output and sensory feedback.} (a) During auditory word repetition, temporal--frontal and frontal--precentral exchanges support phonological and articulatory representation, respectively, in the authors' interpretation of four Granger directions; reciprocal influences peak at different times \citep{flinker2015cortical}. A separate ventral-precentral signal to superior temporal cortex supports corollary discharge \citep{khalilian2024corollary}. (b) The Mouthed synthesis transfers an articulatory plan to precentral execution and receives somatosensory feedback in postcentral and anterior supramarginal territories. (c) The Limited model places preparation and phonetic/pitch intent within the sampled precentral population, with modelled motor output supporting residual articulation \citep{wairagkar2025s0504}. Preparation, sequencing and content remain regional observations. Arrows name information within each stated model; bodily inputs/outputs summarise motor control and sensory feedback across peripheral relays \citep{houde2011state}. Colours locate schematic territories rather than activation extents; the four original views preserve handedness. Areas 6v, M1 and 55b identify sampled precentral territories.}
\label{fig:paradigm_neural_paths}
\end{figure*}

During execution, ventral sensorimotor populations encode articulatory configurations and trajectories, supported by studies using inferred kinematics and directly measured vocal-tract movements \citep{bouchard2013organization,chartier2018kinematics,conant2018cortical}; dorsal laryngeal motor cortex carries vocal-pitch information, and focal stimulation can elicit laryngeal movement and involuntary vocalisation \citep{dichter2018pitch}. Recordings and stimulation during Mandarin speech further implicate this region in lexical-tone control \citep{lu2023cortical}. Focal cooling during overlearned spoken sequences chiefly alters timing at Broca-region sites and articulatory quality at speech motor sites, providing intervention evidence for differentiated contributions \citep{long2016cooling}. \Cref{fig:paradigm_neural_paths} places these planning and execution findings alongside evidence from other articulated conditions.

The resulting sound supplies auditory feedback to temporal speech regions, while frontal and temporal activity also reflects feedforward processing \citep{wang2023feedforward}. During experimentally shifted vocal pitch, auditory error responses and later ventral premotor activity predict compensation; these responses are largely separable from suppression at normal voice onset \citep{chang2013feedback}. During auditory word repetition, Granger analyses identify reciprocal influences between left superior temporal and inferior frontal sites, and between inferior frontal and precentral sites \citep{flinker2015cortical}. The authors interpret the temporal--frontal exchange as supporting phonological representation and the frontal--precentral exchange as forming and implementing articulatory representation; reciprocal influences peak at different times (\cref{fig:paradigm_neural_paths}a). A separate analysis identifies a pre-articulatory influence from left ventral precentral cortex to superior temporal gyrus that predicts subsequent auditory suppression, supporting the authors' corollary-discharge interpretation \citep{khalilian2024corollary}. In the speech-motor-control model, motor commands drive articulation and the returning voice provides auditory feedback; the figure summarises these relations across peripheral and subcortical relays \citep{houde2011state}.

Mouthed speech preserves executed movements while removing audible output. An audible-trained speech synthesiser also decoded mimed sentences from right-hemisphere recordings in one participant \citep{anumanchipalli2019s0127}. Intracranial comparisons locate planning-related activity in left inferior frontal cortex and movement-related responses in precentral cortex; mouthing relative to covert reading additionally recruits inferior Rolandic, secondary somatosensory and anterior supramarginal territories \citep{kwon2025mimed,kell2017reading}. The survey synthesis in \cref{fig:paradigm_neural_paths} combines these findings with a motor-control account: inferior frontal cortex supplies an articulatory plan to ventral precentral cortex, and executed movement returns somatosensory feedback to ventral postcentral and anterior supramarginal regions \citep{houde2011state}. Recent ventral-precentral recordings encode articulation content and speech mode during miming \citep{srinivasan2026s0698}; fMRI comparisons also identify stronger inferior frontal, precentral and superior temporal responses for silent articulation than imagined speech \citep{teghipco2023prediction}.

Limited production can preserve cortical articulatory representations despite severely impaired output. In a participant with residual orofacial movement and unintelligible vocalisation, sampled ventral premotor area 6v carried phoneme-related patterns during attempted vocalised and silent speech \citep{willett2023s0300}; related clinical recordings identify retained articulatory organisation in left sensorimotor cortex \citep{metzger2023s0299}. The Limited example in \cref{fig:paradigm_neural_paths} follows recordings from left precentral territories 6v, M1 and 55b: population activity supports preparation and attempted production, carrying phonetic and pitch intent while residual articulation remains possible \citep{wairagkar2025s0504}. Preparation and production are successive population/task states within the sampled territories; the outgoing motor commands represent the functional link to the participant's residual articulation \citep{houde2011state}. \Cref{sec:neural_representations} relates this recorded content to decoding targets.

\paragraph{Execution, supervision and transfer.} Overt production can supply a participant's acoustic target for learning and alignment; Mouthed retains execution while removing audible speech. Limited can retain movement without intelligible production, as in clinical attempted-speech decoding \citep{willett2023s0300}. Transfer across these conditions must therefore account for differences in target availability and sensory feedback, alongside residual execution. Audible whispering is recorded as Overt with absent voicing: acoustic audibility and vocal-fold voicing are separate descriptors. Evidence from long-standing anarthria motivates studying retained speech representations, but participant comparisons also differ in neural coverage, training and disease history \citep{jude2026s0677}. Those comparisons cannot isolate the effect of movement by themselves. The clinical reports in \cref{tab:attempted} retain the source term attempted when residual movement is not established for that condition.

Direct comparisons help explain the relationship between these modes. Intracranial recordings during overt, mouthed and imagined speech reveal nested sets of channels informative for speech-activity detection \citep{soroush2023s0369}. This supports shared activity, not identical directed pathways; the fixed overt--mouthed--imagined order also limits separation of mode effects from priming.

\clearpage
\clearpage

\subsection{Inner: Articulatory, Auditory and Mixed}

\label{sec:inner}\label{sec:silent}\label{sec:covert}

\paragraph{Inner.} Inner denotes instructed speech simulation without attempted articulator movement; the table distinguishes articulatory, auditory and explicitly combined imagery. Because covert conditions can include mouthing, electromyography (EMG) or motion monitoring separately assesses execution and peripheral contributions \citep{martin2016s0067,bhadra2025s0575,kunz2025s0568}.

\begin{figure*}[p]
\centering
\captionsetup{font=footnotesize}
\includegraphics[width=\linewidth]{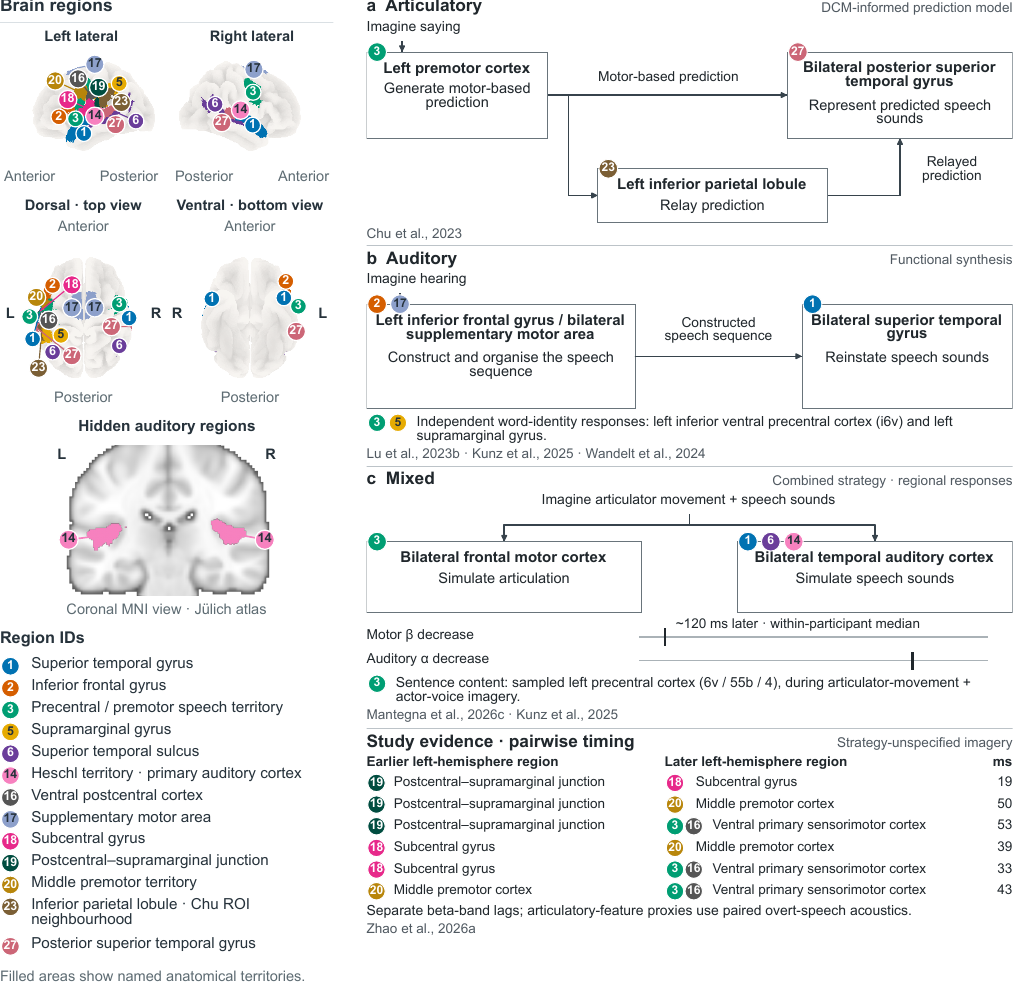}
\caption{\textbf{Inner speech: prediction, voice reinstatement and combined imagery.} (a) In the model, left premotor cortex supplies motor-based prediction to bilateral posterior superior temporal cortex directly and through left inferior parietal lobule, following estimated effective-connectivity directions \citep{chu2023dual}. (b) The auditory synthesis transfers a constructed speech sequence from frontal regions to superior temporal cortex \citep{lu2023corticalb}; separate recordings supply precentral and supramarginal content observations. (c) Combined movement-and-sound imagery engages motor and auditory regions under the same instruction; its approximately 120\,ms onset difference appears in a separate timing view \citep{mantegna2026mu}. Precentral sentence content is a distinct observation \citep{kunz2025s0568}. A separate table reports six pairwise beta-band lags during syllable imagery \citep{zhao2026motor_dynamics}. Timing views quantify measured precedence, while labelled arrows express functional information transfer. Coloured anatomy denotes schematic territories; Heschl anatomy and the separate J\"ulich TE1 coronal inset localise auditory cortex. The four original cortical views preserve handedness.}
\label{fig:inner_neural_paths}
\end{figure*}

\paragraph{Generation strategies and temporal organisation.} Articulatory imagery rehearses a speaking action without trying to execute it; auditory imagery reinstates an imagined voice. MEG and fMRI comparisons motivate motor-simulation and memory-retrieval accounts \citep{tian2010imagery,tian2016reactivation}. During imagined saying of written Chinese characters, dynamic causal modelling estimates stronger left-premotor influence on bilateral posterior superior temporal gyri through two routes: direct input and a relay through left inferior parietal lobule \citep{chu2023dual}. Each temporal target receives both influences. \Cref{fig:inner_neural_paths}a expresses these as direct and relayed motor-based prediction, the authors' functional interpretation of the estimated influences.

Auditory imagery constructs a remembered voice. Imagining previously heard poems recruits left inferior frontal and bilateral supplementary motor areas together with bilateral superior temporal cortex; the authors interpret this pattern as internal construction and sequencing that reinstates speech sounds \citep{lu2023corticalb}. The frontal-to-temporal flow in \cref{fig:inner_neural_paths}b is a survey synthesis of this account, labelled as a constructed speech sequence. Separate intracortical experiments identify imagined-word representations in left inferior ventral precentral cortex and, during a sound-imagination condition, left supramarginal gyrus \citep{kunz2025s0568,wandelt2024s0464}. These appear as independent content observations beside the voice-generation model.

Both strategies can recruit speech motor processes. Lip responses evoked by transcranial magnetic stimulation (TMS) are larger during inner speaking than inner hearing, with phonetic modulation also present during inner hearing \citep{nalborczyk2023speaking}. Recent MEG decoding finds motor representations before auditory representations for imagined syllables differing in consonants; syllables differing in vowels lack the same early motor discrimination \citep{orpella2026covert}. Their covert-onset reference is estimated from overt production, so this timing remains tied to the tested contrasts. At a broader scale, individually mapped networks place the overlap between cued speech imagery and listening predominantly in transmodal language regions rather than adjacent unimodal auditory cortex \citep{anderson2026imagery}.

Mixed combines movement and sound imagery explicitly. With instructions to simulate both the movements and sounds of syllables, bilateral frontal motor beta desynchronisation precedes bilateral temporal auditory alpha desynchronisation; the median within-participant onset difference is approximately 120\,ms \citep{mantegna2026mu}. \Cref{fig:inner_neural_paths}c shows both regions under the same combined-imagery instruction and compares their onset times separately. In a separate intracortical study, participant T15 combined imagined articulator movements with the imagined sound of an actor's voice; sampled left precentral populations in 6v, 55b and area 4 encoded internal sentence content \citep{kunz2025s0568}. The figure retains this encoding result as an independent regional observation.

\paragraph{Recorded regions and recoverable content.} Recent decoding from sampled precentral and supramarginal populations shows that internal speech can elicit content-bearing activity without executing an utterance \citep{wandelt2024s0464,kunz2025s0568}. Lesion associations also link left frontal and anterior supramarginal damage to impaired silent rhyme judgments after accounting for overt reading \citep{geva2011inner}. Low-beta features in bilateral superior temporal recordings distinguish imagined items \citep{proix2022s0267}, while imagined metrical-stress judgments recruit left posterior superior temporal cortex relative to a semantic decision control \citep{aleman2005stress}. Response contrasts and decodable content thus provide complementary evidence at the regional scale. \Cref{sec:neural_representations} discusses the recorded content and task-state distinctions; testing user-chosen expression requires the independent-content checks described below.

Temporal studies further distinguish operations within internally generated language. During silent inflection of prompted words, neighbouring recordings within Broca's area showed successive lexical, grammatical and phonological effects \citep{sahin2009sequential}. During syllable imagery, beta-band activity at the left postcentral--supramarginal junction leads activity in subcentral, middle-premotor and ventral primary sensorimotor populations; subcentral activity also leads the latter two, and middle-premotor activity leads ventral sensorimotor activity \citep{zhao2026motor_dynamics}. The first three populations share articulatory-feature coding across overt and imagined speech, while the ventral primary sensorimotor population shows imagery-specific coding. Here, articulatory features are trajectory proxies inferred from paired overt-speech acoustics. The independent timing table in \cref{fig:inner_neural_paths} reports the six pairwise cross-correlation lags. Each value describes its own regional comparison; the source experiment compared overt and imagined syllables with no separate strategy manipulation.

\paragraph{Relation to wider inner-speech terminology.} Articulatory and auditory imagery describe how an inner utterance is generated. Inner experience also varies in condensation, dialogality, intentionality and prosody, so rehearsing an expanded prompted sentence samples only part of this broader capacity \citep{grandchamp2019condialint,daho2025reviewr53,alexander2024reviewr73,kreiner2024reviewr52}. Speech-imagery reviews document heterogeneous elicitation and recording protocols \citep{tates2025reviewr11,almufareh2025reviewr13}. Retaining the instruction, content source and behavioural monitoring alongside the subtype connects these cognitive dimensions to reproducible decoding tasks.

\clearpage
\clearpage

\subsection{Perceived: Auditory, Visual and Audiovisual}

\label{sec:perceived}

Perceived uses externally presented language; Reading and Lip-reading are task descriptors within Perceived--Visual (\cref{tab:paradigms}). ZuCo reading EEG is one example \citep{hollenstein2018s0778,hollenstein2019s0779}. A written cue before an imagery or speech-attempt interval does not make the later decoded operation Reading (\cref{fig:taxonomy}).

Imagining a voice belongs to Inner--Auditory. Visible objects or silent scenes without a linguistic target remain outside Perceived--Visual; movie-based matching is language decoding only when the evaluated target is language. Audiovisual speech representation, for example, includes the interaction of heard phonemes and visible articulation \citep{li2026s0772}. These distinctions preserve the difference between input modality and output meaning.

\paragraph{From sound to language.} Auditory speech provides a sound-driven sensory entry: superior temporal populations encode spectrotemporal and acoustic--phonetic features \citep{mesgarani2014features,leonard2024single_neuron}. Middle superior temporal sites prominently encode rapid envelope rises, whereas posterior sites respond to speech onsets \citep{oganian2019cortical}. In competing-speaker conditions, attention reshapes the speech represented in these regions \citep{mesgarani2012cortical,osullivan2019cortical}. Simultaneous intracranial recordings show similarly early responses in posteromedial Heschl's gyrus, planum temporale and the posterior superior temporal onset zone, supporting parallel processing \citep{hamilton2021parallel}. In the same study, stimulation of Heschl's gyrus induced auditory experiences, whereas lateral superior temporal stimulation interfered with word perception.

\begin{figure*}[p]
\centering
\captionsetup{font=footnotesize}
\includegraphics[width=\linewidth]{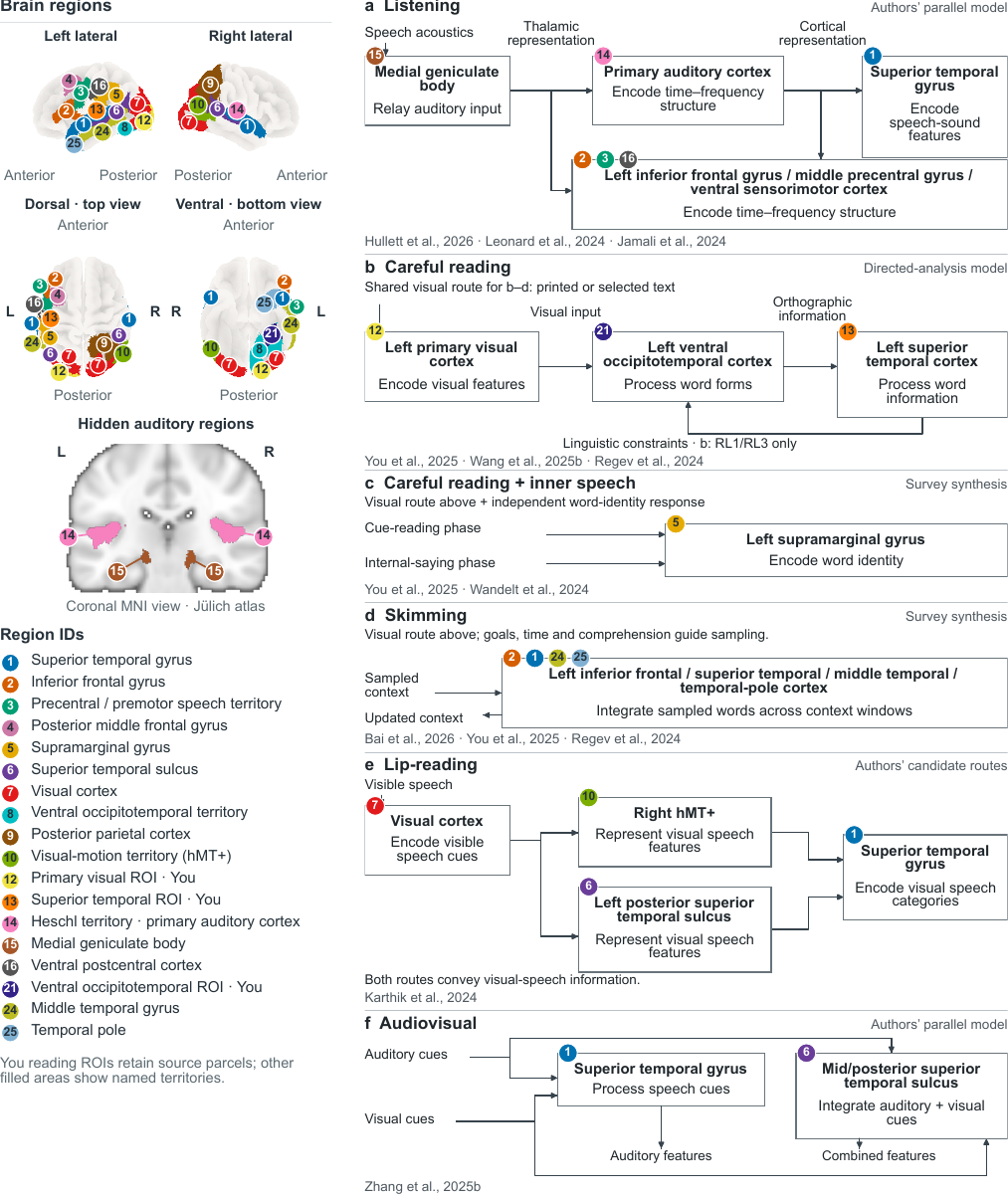}
\caption{\textbf{Perceived language: sensory entry and task-dependent processing.} The visual route is shared across (b--d); feedback applies to pseudoword base-word recovery (RL1/RL3) in (b). Word identity in (c) and context integration in (d) add independent study observations to the survey syntheses. Arrows specify information within each stated model; identifiers link modules to anatomy. Registered reading parcels retain source boundaries; other fills show schematic territories. Detailed regional findings accompany the figure in the text.}
\label{fig:perceived_neural_paths}
\end{figure*}

Recent passive-listening recordings identify short-latency spectrotemporal responses in left inferior frontal, middle precentral and ventral sensorimotor regions \citep{hullett2026parallel}. In the authors' candidate model, auditory information passes from medial geniculate body through primary auditory cortex to superior temporal gyrus, while medial-geniculate and primary-auditory inputs also supply frontal speech regions in parallel (\cref{fig:perceived_neural_paths}a). Thalamic and cortical identify the source levels of the modelled auditory representations; both carry speech time--frequency structure. Separate single-neuron recordings locate context-sensitive word-meaning representations in left posterior middle frontal gyrus \citep{dataset_jamali2024_002cf}. At larger linguistic scales, phrasal and sentential structure is tracked in superior temporal, temporoparietal and left inferior frontal regions \citep{ding2016cortical}; contextual language-model representations predict responses across frontal, temporal-pole, posterior superior temporal and angular regions during story listening \citep{goldstein2022cortical}. \Cref{sec:neural_representations} develops these distributed context-sensitive representations.

\paragraph{Reading operations and neural processing.} Reading instructions specify both a goal and an accompanying strategy. Careful comprehension, skimming for gist, targeted information search and spell checking impose different demands \citep{strukelj2018reading,hollenstein2019s0779}. Instructed inner verbalisation adds a strategy to the reading goal; careful reading alone does not establish whether the participant silently rehearses the words. One direct manipulation asks readers to imagine a familiar speaker saying each sentence and measures eye movements and comprehension \citep{zhou2016readingvoice}. Covert reading, mouthing while reading and overt reading must also remain distinct \citep{kell2017reading}. These labels describe operations and do not imply that each has a separately established neural pathway.

Word recognition involves ventral occipitotemporal cortex, with linguistic context engaging a wider frontal--temporal network \citep{hirshorn2016wordforms,dataset_fedorenko2016_69ce2}. Intracranial recordings distinguish two temporal effects: word/non-word selectivity emerges in mid-fusiform cortex before posterior sites, whereas word-frequency sensitivity appears earliest in both mid-fusiform and lateral occipitotemporal cortex before spreading posteriorly \citep{woolnough2021orthographic}. An earlier letter/word contrast also identified letter-selective posterior fusiform responses \citep{thesen2012cortical}. During visually presented word and pseudoword reading, MEG Granger analyses estimate influences from left primary visual to ventral occipitotemporal and then superior temporal cortex \citep{you2025connectivity}. The same analysis estimates later superior-temporal-to-occipitotemporal feedback for one- and three-letter-replacement pseudowords during base-word recovery (RL1/RL3). \Cref{fig:perceived_neural_paths}b selects this feedforward route and condition-specific feedback; orthographic information and linguistic constraints name the authors' functional interpretation of their content.

Natural reading also engages processes that overlap in time. MEG during self-paced sentence reading localises parafoveal word-form processing to left ventral occipitotemporal cortex and semantic processing to left inferior frontal and right posterior parietal cortex, with partly overlapping response windows \citep{wang2025parafoveal}. Separately, intracranial recordings identify spatially interleaved frontal and temporal language populations that integrate over different word windows \citep{regev2024windows}. These regional observations complement the selected reading circuit; their locations are included in the anatomy column of \cref{fig:perceived_neural_paths}.

Reading with instructed inner verbalisation combines external word processing with internally generated word representations. Left supramarginal neurons encode cued and internally spoken words, with shared word selectivity across the cue-reading and internal-saying phases \citep{wandelt2024s0464}. The survey synthesis in \cref{fig:perceived_neural_paths}c places this observation beside the selected reading branch \citep{you2025connectivity}; separate task labels identify the cue-reading and internal-saying conditions of the supramarginal recordings. Skimming changes which words enter language processing: goals and time pressure alter fixation, skipping and revisiting, while updated comprehension guides the next visual sample \citep{bai2026reading}. \Cref{fig:perceived_neural_paths}d combines this behavioural model with the word-processing route and distributed context integration at the survey-synthesis level.

Lip-reading begins with visible articulatory movements. Visual and posterior superior temporal responses accompany speech-category representations in superior temporal gyrus \citep{karthik2024s0391,brohl2022lipreading}, and visual responses can track aspects of the unheard acoustic signal \citep{hauswald2018cortical}. The authors' model in \cref{fig:perceived_neural_paths}e carries visual-speech information from visual cortex to superior temporal gyrus through two candidate routes: one via right hMT+ and the other via left posterior superior temporal sulcus \citep{karthik2024s0391}.

\paragraph{Combining auditory and visual speech.} Audiovisual speech combines heard speech with visible articulation. Intracranial recordings motivate a model that supplies both auditory and visual cues to superior temporal gyrus and mid/posterior superior temporal sulcus in parallel \citep{zhang2025audiovisual}. It assigns a leading role in auditory speech processing to the gyrus and in audiovisual integration to the sulcus. Relative to auditory-only presentation, audiovisual speech elicits earlier and stronger sulcal responses for clear speech and stronger responses in noise; word recognition also improves over auditory-only speech in noise (\cref{fig:perceived_neural_paths}f). Complementary evidence comes from incongruent syllables: inferior frontal and supramarginal fMRI responses depend on the perceived syllable beyond physical stimulus overlap \citep{hasson2007audiovisual}, and TMS over left posterior superior temporal sulcus reduces McGurk fusion within a restricted time window \citep{beauchamp2010mcgurk}. A September 2026 preprint further reports enhanced phoneme-level separability in superior temporal gyrus for congruent audiovisual relative to auditory-only words \citep{li2026s0772}.

\paragraph{A resource for learning and testing transfer.} Presented language provides a known target and timing, making Perceived tasks useful for studying neural representations and learning decoder components. The value can extend beyond stimulus reconstruction: an fMRI decoder trained during listening was also evaluated during imagined storytelling, with semantic agreement assessed against separately recorded accounts \citep{tang2023s0356}. That result motivates cross-task learning while locating the achievement at the tested semantic target and protocol. Shared information between tasks is a resource for transfer; the fidelity and user control required for a communication interface still need to be established in its intended task.

\clearpage
\clearpage

\subsection{Elicitation, cues and behavioural verification}

\paragraph{Independent task descriptors.} Content may be externally specified or self-generated; internal speech may be instructed or spontaneously occurring; items may be produced once or repeatedly. These dimensions can combine and are not additional subtypes. Cue modality, preparation intervals, residual sound and movement supply additional context; peripheral monitoring tests compliance with the instruction. A pre-articulatory window preceding an overt utterance is not automatically Inner or clinical attempted speech \citep{dash2020s0150,dash2024s0456}. \Cref{fig:residual} shows the task ingredients implied by the definitions, not measured neural strengths.

Auxiliary recordings help characterise how a participant performs the task. A preregistered experiment found that facial EMG distinguished phonetic classes during overt speech but did not establish comparable group-level discrimination during instructed inner speech; increased activity relative to rest was not itself specific to speech content \citep{nalborczyk2020emg}. The useful analysis therefore relates peripheral activity to the instruction and linguistic target. Exact cue sequences, imagery instructions and control channels allow later studies to reproduce that comparison.

\paragraph{What a task-transfer claim should test.} Task transfer can change the neural activity being decoded, the source of the message, or both. Each change requires its own test. To distinguish a cue-dependent result from decoding of a later attempt or imagery interval, useful tests vary the cue--target mapping and evaluate the later interval separately. Temporal separation should reflect the signal's physiological response and any carry-over from the cue; separating epochs alone does not establish independence (\cref{sec:signals,sec:eval}). A claim of self-generated expression additionally requires user-chosen content that is unavailable to the decoder, together with a means of checking whether the output conveys the intended message. Such tests target the dependency being removed, rather than treating a change of task name as evidence of transfer.

\section{Signals}
\label{sec:signals}

Neural recording provides complementary views of language processing. Electrical and magnetic measurements resolve rapid population dynamics; haemodynamic measurements capture spatially structured vascular responses over longer intervals. Within each family, placement and coverage determine which linguistic processes are most accessible. Choosing a signal therefore connects a scientific target---such as articulatory movement, speech sounds or contextual meaning---to a feasible recording configuration and timescale.

\begin{table}[!htbp]
\centering\small
\caption{Temporal and spatial scales in representative neural-recording configurations. Values describe the cited studies, rather than universal family limits. Sampling intervals, feature windows and vascular response dynamics describe different stages; contact pitch and voxel size specify sampling geometry, rather than independently measured functional localisation.}
\label{tab:signal_resolution_comparison}
\begingroup
\setlength{\tabcolsep}{4pt}
\renewcommand{\arraystretch}{1.14}
\begin{tabular}{@{}>{\raggedright\arraybackslash}p{0.18\linewidth}>{\raggedright\arraybackslash}p{0.25\linewidth}>{\raggedright\arraybackslash}p{0.21\linewidth}>{\raggedright\arraybackslash}p{0.30\linewidth}@{}}
\toprule
Family and example & Temporal sampling or analysis & Response being measured & Spatial sampling and coverage \\
\midrule
EEG \citep{nieto2022s0292} & 1,024-Hz acquisition ($0.98$-ms interval) & Scalp potentials; analysis depends on the frequency band and time window & 128 scalp electrodes; cortical localisation additionally depends on the head and source models. \\
MEG \citep{gwilliams2023s0341} & 1,000-Hz acquisition (1-ms interval) & Magnetic fields from neural currents; temporal filtering shapes the features & 208 axial gradiometers around the head; sensor spacing is not cortical localisation accuracy. \\
fMRI \citep{tang2023s0356} & Volume repetition time, TR $=2$ s & BOLD vascular response; TR is not the response duration & 2.6-mm isotropic voxels; distributed brain coverage, with voxel sampling distinct from functional specificity. \\
fNIRS / HD-DOT \citep{schroeder2023s0343} & Analysis resampled to 1 Hz after low-pass filtering & Haemoglobin changes mediated by vascular dynamics & Overlapping optical paths over superficial cortex; 13-mm nearest-neighbour spacing is optode geometry, not image resolution. \\
fUSI \citep{soloukey2019s0116} & 500--667-Hz compounded frames; 3.6--4.8-Hz live Doppler images & Blood-volume-sensitive Doppler power; neural activity is related through vascular coupling & Reported 300-$\mu$m imaging scale; plane and acoustic window determine coverage. \\
ECoG / $\mu$ECoG \citep{duraivel2023s0332} & Field-potential features depend on band selection and temporal averaging & Local cortical-surface electrical activity & Example $\mu$ECoG array: 200-$\mu$m contacts, 1.33-mm pitch; coverage follows the implanted surface array. \\
sEEG \citep{verwoert2025s0585} & 1,024- or 2,048-Hz acquisition; analysed at 1,024 Hz & Field potentials near depth contacts & Sparse sampling along clinically selected depth trajectories; access to deep sites does not imply dense whole-brain coverage. \\
Intracortical \citep{wairagkar2025s0504} & 30-kHz voltages; 1-ms feature extraction; 10-ms feature bins & Threshold crossings and spike-band power & 256 microelectrodes in ventral precentral cortex; local coverage rather than whole-cortex sampling. \\
\bottomrule
\end{tabular}
\endgroup
\end{table}

\subsection{Neural representations of language}
\label{sec:neural_representations}

\begin{figure*}[!htbp]
\centering
\includegraphics[width=0.97\textwidth]{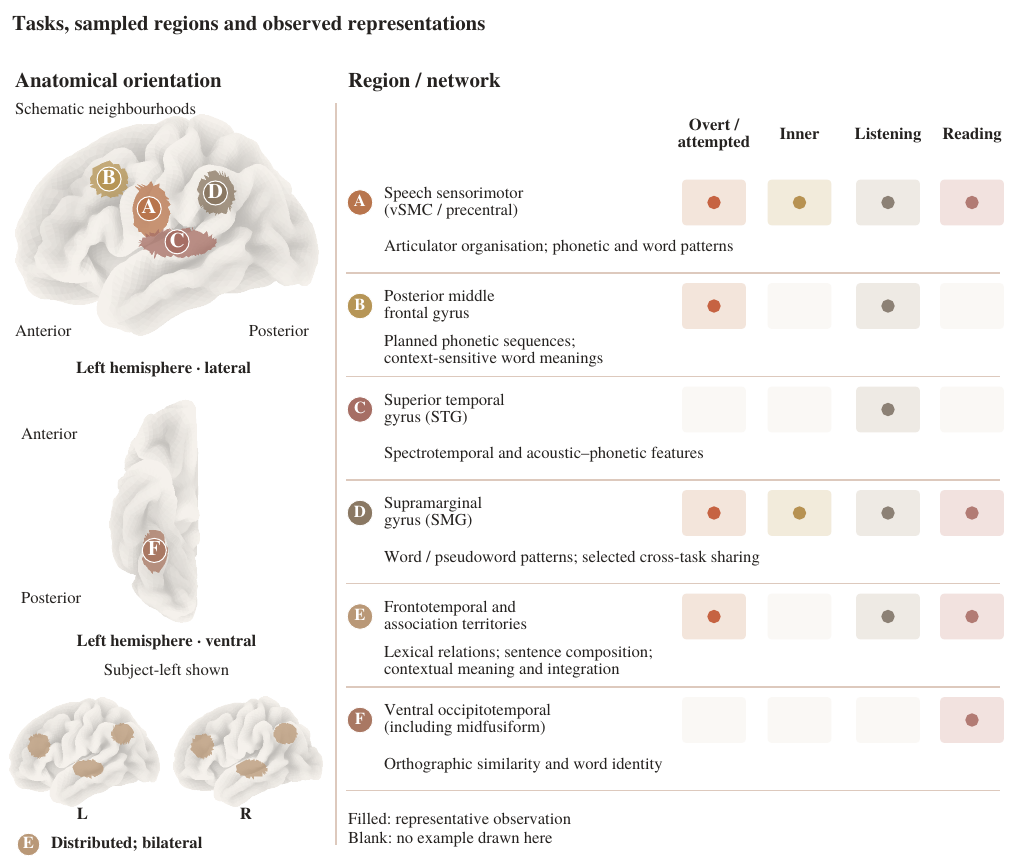}
\caption{\textbf{Neural representations across language tasks.} The fsaverage5 surfaces provide anatomical orientation \citep{fischl1999surface}; coloured neighbourhoods are schematic, not electrode coordinates or functional boundaries. E denotes bilateral distributed territories. Filled cells mark representative observations; blank cells mean no example is drawn. Overt and attempted speech remain distinct tasks despite sharing a column. Measurements and sources: \textbf{A}, sensorimotor surface and array recordings \citep{bouchard2013organization,kunz2025s0568}; \textbf{B}, posterior frontal single neurons during speech planning and contextual listening \citep{khanna2024s0467,dataset_jamali2024_002cf}; \textbf{C}, superior temporal surface recordings during listening \citep{mesgarani2014features}; \textbf{D}, supramarginal arrays during vocalised and internal words in two participants, with auditory/written cues and cross-task sharing tested in participant 1 \citep{wandelt2024s0464}; \textbf{E}, frontotemporal single neurons and ECoG, plus fMRI of semantic maps and reading--listening timescales \citep{dataset_cai2026_6c6ce,dataset_fedorenko2016_69ce2,dataset_huth2016_c0291,chen2024timescales}; \textbf{F}, ventral temporal iEEG and stimulation, plus source-localised MEG of parafoveal reading \citep{hirshorn2016wordforms,wang2025parafoveal}.}
\label{fig:task_network_representation}
\end{figure*}

Language tasks recruit overlapping systems while placing different demands on them: producing a word requires a plan for its form and articulation, hearing it requires interpreting an acoustic event, and reading it requires identifying a visual form. Across these tasks, words acquire meaning through their relationships to surrounding words and the broader situation. The useful unit for neural decoding is therefore a task-dependent representation in a sampled population, rather than a fixed correspondence between one brain region and one linguistic function. \Cref{fig:task_network_representation} compares representative observations across tasks; \cref{fig:routes} explains how decoded representations support communication outputs.

\paragraph{From linguistic plans to articulation.}
Speech production combines the construction of an utterance with coordination of the vocal tract. Cortical-surface recordings over ventral sensorimotor cortex reveal overlapping representations of the lips, tongue, jaw and larynx; their population activity changes as these articulators combine into speech sounds \citep{bouchard2013organization}. This organisation provides a basis for predicting articulatory or acoustic trajectories from neural activity. Planning also carries structure before a movement or sound occurs. Acute single-neuron recordings in the language-dominant posterior middle frontal gyrus distinguish the phonetic composition, order and syllabification of upcoming words \citep{khanna2024s0467}. More extensive frontotemporal recordings resolve grammatical relationships, phrase structure and contextual semantic information before word utterance \citep{dataset_cai2026_6c6ce}. These studies distinguish the information needed to construct a word sequence from the information needed to execute its articulation, while showing that both involve distributed, mixed population codes. Accordingly, recording placement influences whether a decoder has direct access to phonetic planning, articulatory dynamics or higher-level sentence organisation.

\paragraph{From speech sounds to contextual meaning.}
During listening, superior temporal gyrus responses distinguish acoustic--phonetic features through sensitivity to spectrotemporal cues in continuous speech \citep{mesgarani2014features}. Such responses motivate acoustic reconstruction and speech-feature alignment, because the predicted target preserves aspects of the sound that elicited the activity. Comprehension additionally represents what a word means in context. Voxelwise models of narrative-listening fMRI reveal semantic selectivity across bilateral temporal, parietal and prefrontal territories \citep{dataset_huth2016_c0291}. At a different scale, single neurons in left posterior middle frontal cortex respond to word meanings in ways that change with the sentence context, rather than simply following phonetic form \citep{dataset_jamali2024_002cf}. Thus, contextual semantic information is accessible both in local spiking and in distributed haemodynamic measurements; it is not a property exclusive to one recording family. Conversely, a shared recording region need not imply identical task codes: the posterior frontal neurons selective for heard phonetic content were largely distinct from those selective during speech planning in a within-study task comparison \citep{khanna2024s0467}.

\paragraph{Reading links visual forms to language composition.}
Reading begins with a visual input whose orthographic structure must be connected to lexical and sentence information. Intracranial recordings in left midfusiform cortex distinguish an early representation organised by orthographic similarity from later word-specific activity. Electrical stimulation at the tested sites, and resection in one participant, impaired visual word or letter perception, providing evidence for a causal contribution to these reading operations \citep{hirshorn2016wordforms}. Sentence understanding extends beyond recognition of individual forms: high-gamma responses across frontal and temporal sites build up as meaningful sentences unfold, with different response profiles for word lists, nonword lists and syntactically structured nonwords \citep{dataset_fedorenko2016_69ce2}. The contrasts link this build-up to the construction of sentence meaning. Recent natural-reading MEG with eye tracking further identifies parafoveal orthographic similarity in left ventral occipitotemporal cortex and semantic similarity in left inferior frontal cortex before the target word is fixated \citep{wang2025parafoveal}. These source-localised associations show why a fixation-aligned neural segment can contain information about neighbouring words and context as well as the currently fixated word. A reading decoder consequently samples a mixture of visual recognition, linguistic integration and, depending on the reader's strategy, internal verbal activity \citep{kunz2025s0568}.

\paragraph{Inner and attempted speech share content but differ in state.}
Internally saying a word, imagining its sound and attempting its articulation provide related routes to linguistic content. Their overlap creates opportunities for transfer, while their differences determine which supervision remains appropriate. Microelectrode recordings from supramarginal gyrus support classification of internally and overtly produced words and pseudowords in two participants; shared representations with word reading and vocalised speech were demonstrated in one participant \citep{wandelt2024s0464}. Motor-cortex recordings in four participants subsequently showed correlated word-pattern geometry across attempted, inner and perceived speech. Inner speech was more weakly modulated, and a population dimension associated with motor intention helped distinguish inner from attempted speech \citep{kunz2025s0568}. This combination is useful for an interface: content information can be shared while the intended mode of communication remains distinguishable. It also explains why an absence of visible movement does not by itself identify the neural task. Transfer between listening, attempted and inner speech depends on the particular population, instructions and representation, rather than on a shared word label alone.

\paragraph{Matching the representation to the measurement and output.}
Cross-task reuse also extends to longer linguistic context. Models fitted to fMRI responses during matched reading and listening narratives reveal similar cortical organisation of language-context timescales across temporal, parietal and prefrontal regions \citep{chen2024timescales}. This supports a common substrate for integration beyond the modality-specific input, without requiring identical fine-scale dynamics. Measurement scale still matters: the linguistic selectivity of individual neurons can differ from that of field potentials at the same recording sites \citep{dataset_cai2026_6c6ce}. Broad coverage and local precision therefore answer complementary questions about accessible information.

The decoding target selects which of these properties the interface preserves. Acoustic representations can support reconstruction of heard speech \citep{pasley2012s0021}; phoneme sequences can connect attempted speech to text \citep{willett2023s0300}; contextual representations can constrain a generated description of perceived or imagined language \citep{tang2023s0356}. A single speech-cortex recording can also feed text, speech synthesis and audio-linked facial animation through different prediction targets \citep{metzger2023s0299}. These are many-to-many relationships between neural information and output forms. Representation studies explain why a target is plausible; held-out and online decoding establish how effectively a particular system can recover it.

\subsection{Electrophysiological signals}

Electroencephalography (EEG) measures scalp potentials arising from the combined electrical activity of neuronal populations; around-ear placement is one recording configuration. Magnetoencephalography (MEG), including conventional systems and optically pumped magnetometers (OPMs), measures magnetic fields associated with neural activity. Invasive recordings include cortical-surface electrocorticography (ECoG) and its high-density or micro-ECoG configurations, stereo\-electro\-encephalography (sEEG) depth macro-contacts, intra\-cortical arrays and acute Neuropixels probes. Intracranial EEG (iEEG) encompasses ECoG and sEEG. Extracellular field potentials reflect pooled local activity, whereas spikes and threshold crossings provide features associated with neuronal firing. Intracortical decoders use these alongside spike-band power and local field potentials. These families differ in spatial coverage and signal access; a common model architecture does not make their neural inputs equivalent \citep{herff2016review,duraivel2023s0332,khanna2024s0467,willett2023s0300}.

These measurements offer different forms of information access. Scalp and ear EEG can track features of attended external speech, and source-localised MEG can characterise auditory and visual speech information over time \citep{geirnaert2025ear,brohl2022lipreading}. Cortical-surface recordings resolve local population activity with articulatory and acoustic-phonetic organisation \citep{bouchard2013organization,mesgarani2014features}. Intracortical arrays add features derived from neuronal spiking and local fields, which have supported large-vocabulary speech decoding \citep{willett2023s0300}. Thus, the feature representation and recording location explain a decoder's input more precisely than the device name alone.

Beyond content reconstruction, distributed sEEG recordings support speech-state detection and identification of intent-related contacts \citep{verwoert2025s0585,prakash2025s0517}. These targets help identify when and where language-related activity can be read out.

\subsection{Haemodynamic signals}

Functional magnetic resonance imaging (fMRI), functional near-infrared spectroscopy (fNIRS), high-density diffuse optical tomography (HD-DOT) and functional ultrasound imaging (fUSI) measure vascular responses associated with neural processing. BOLD fMRI uses magnetic-resonance contrast sensitive to blood oxygenation; fNIRS estimates haemoglobin changes optically, with dense sampling and reconstruction in HD-DOT; fUSI uses Doppler measurements sensitive to microvascular blood-volume changes. Simultaneous electrophysiological and fMRI experiments in nonhuman primates provided an empirical basis for interpreting this relationship \citep{logothetis2001fmri}. Optical reconstruction and Doppler acquisition provide different views of vascular activity \citep{schroeder2023s0343,soloukey2019s0116}. The response integrates activity over time, making temporal alignment and target duration important parts of a language-decoding model.

Language applications illustrate the range of targets available through this response. fNIRS has supported concept discrimination from responses to paired pictures and spoken labels, while fMRI decoders reconstruct the gist of extended language and high-density fNIRS systems investigate cue-conditioned sentence generation \citep{cao2021fnirs,tang2023s0356,suyi2024s0450}. HD-DOT combines dense optical sampling with reconstruction to map tasks such as silent reading, silent verb generation and overt production, and to retrieve stimulus segments \citep{schroeder2023s0343,tripathy2025s0586}. These approaches exploit information integrated over longer intervals. Faster sampling can improve the measurement, but the vascular response must still be included when interpreting output latency.

\subsubsection{Functional ultrasound: mapping and decoding boundaries}\label{sec:fus}

\begin{figure}[!htbp]
\centering
\includegraphics[width=\linewidth]{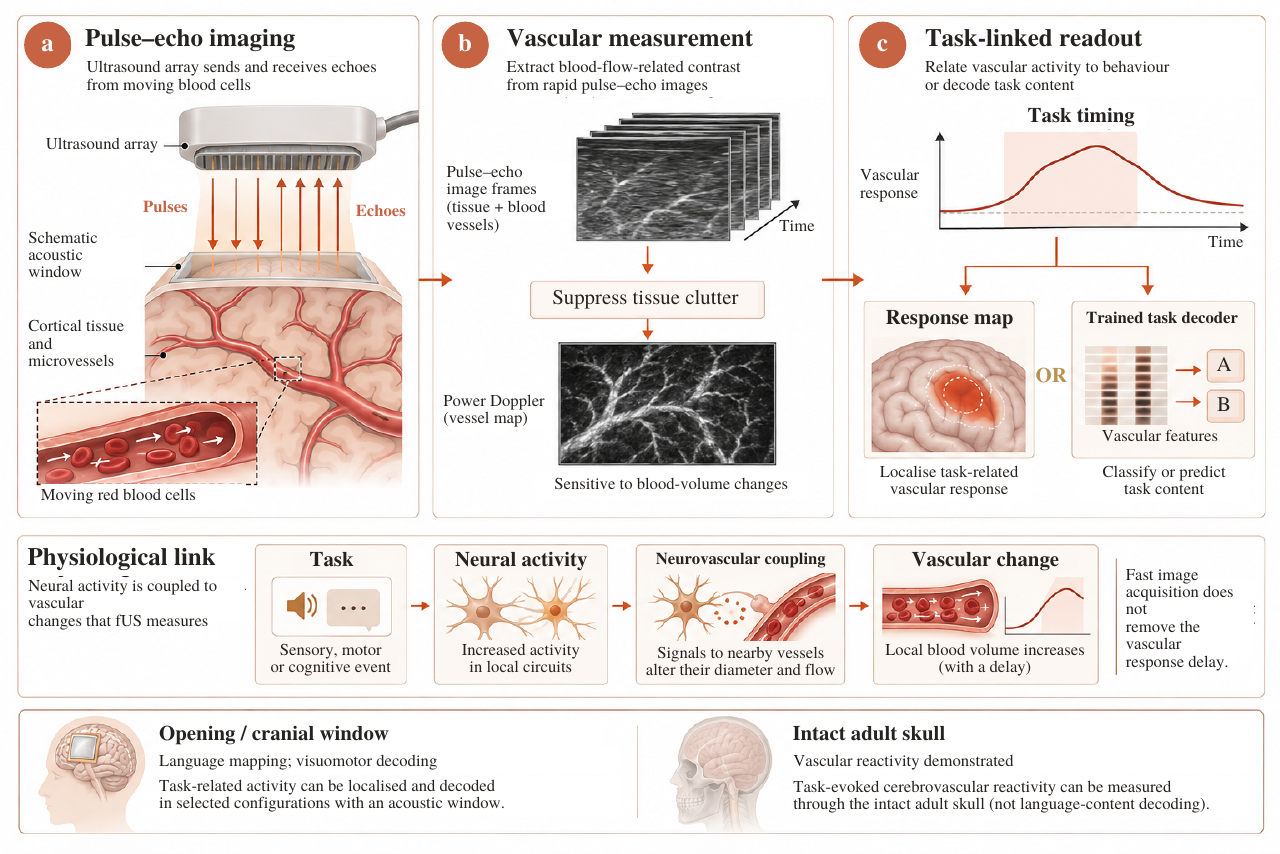}
\caption{\textbf{How functional ultrasound imaging (fUSI) relates neural activity to a vascular measurement.} (a) An array transmits pulses and receives echoes, including those scattered by moving blood cells. (b) Tissue-clutter suppression and power-Doppler processing yield a blood-volume-sensitive image sequence. (c) Task-linked vascular changes support response mapping or a trained decoder. Neurovascular coupling provides the physiological link; rapid acquisition does not remove the vascular response delay. Intra-operative language mapping and window-based visuomotor decoding have been reported \citep{soloukey2019s0116,rabut2024s0437}. The cited intact-skull study instead measured breath-hold vascular reactivity \citep{vienneau2026s0769}. All anatomy, vessel images and traces are schematic. The figure depicts imaging, not ultrasound stimulation or direct spike measurement.}
\label{fig:fus_explainer}
\end{figure}

An ultrasound array sends acoustic pulses into tissue and records returning echoes. Rapid image sequences allow tissue motion to be suppressed and echoes from moving blood cells to be retained; power-Doppler processing then yields a measurement sensitive to local blood volume. Neural activity is related to this vascular signal through neurovascular coupling, so image acquisition speed and the underlying physiological response are different timescales \citep{soloukey2019s0116,rabut2024s0437}. \Cref{fig:fus_explainer} follows this measurement from pulse transmission to task-linked readout.

Functional ultrasound imaging offers sub-millimetre vascular measurements in selected configurations. Its potential for language interfaces depends on joining three advances that currently address different problems: localising language-related activity, predicting content from single trials and maintaining recording access (\cref{tab:fus}).

\paragraph{Localising a response is the first step.} Intra-operative fUSI establishes that task-related haemodynamics can be measured in awake patients \citep{imbault2017s0073}. Language evidence includes aloud and source-labelled covert repetition in a ten-patient series with 18 language measurements, with both responses shown for one patient \citep{soloukey2019s0116}, and overlap with fMRI and electrical stimulation mapping (ESM) \citep{soloukey2023s0333}. Agreement between maps supports localisation of a task response. It leaves open whether that response distinguishes the words or sentences within the task.

\paragraph{Decodability and sustained access are separate advances.} Human single-trial finger and effector decoding across sessions demonstrates usable motor information through acoustically transparent cranial windows \citep{rabut2024s0437,lin2026s0707}. Natural-behaviour imaging over more than 20 months and miniaturised four-dimensional systems address access and recording practicality \citep{soloukey2025s0584,verhoef2025s0583}. Neither establishes language-content recovery: imagined lip licking in the former study is non-speech motor imagery, outside Inner. Non-human primate evidence progresses from propagation of task-related activity to movement-intention decoding and closed-loop ultrasonic brain--machine interfaces \citep{dizeux2019s0117,norman2021s0225,griggs2024s0412}.

\paragraph{The access route limits extrapolation.} Intact-skull human fUSI measured breath-hold vascular reactivity without contrast agents in 13 participants \citep{vienneau2026s0769}; ex-vivo human-skull acoustic transparency is additional feasibility evidence \citep{wang2025s0498}. These findings show that a cranial window is not universally required, but the motor-decoding results obtained through windows cannot simply be assigned to transcranial acquisition. Auditory-response work in ferret, crow and marmoset, and click responses in comatose patients, similarly establishes sensitivity to stimulation rather than language identity \citep{bimbard2018s0095,landemard2025s0552,liao2025s0543,takahashi2021s0226,chen2025s0548}. Cross-species comparisons and mouse temporal-resolution measurements help explain the signal while retaining their species and modality boundaries \citep{landemard2021s0203,bian2026s0657}.

Together, human studies link language-related vascular mapping, single-trial motor decoding and repeated recording access \citep{soloukey2019s0116,lin2026s0707,soloukey2025s0584}. They motivate testing linguistic distinctions under a specified access route, with the target duration matched to the vascular response.

\subsection{Sensors, placement and recording configurations}

\paragraph{Density, coverage and alignment solve different problems.} Denser sampling can recover distinctions lost when nearby signals are pooled: 128- and 256-contact intra-operative \muecog improved decoding of non-words built from nine phonemes in four participants \citep{duraivel2023s0332}. Across patients, the challenge changes because the sampled populations differ; shared latent dynamics improve decoding with \muecog and sparse sEEG \citep{spalding2026s0759,singh2025s0629}. Temporal modelling is another design choice: recurrent and Transformer models outperformed linear regression for sEEG speech reconstruction, with no significant difference detected between the two deep-learning methods in that study \citep{wu2024s0469}. These examples support improving spatial sampling, representational alignment and temporal modelling for different reasons; none shows that model choice compensates for missing coverage. Acute Neuropixels findings resolve articulatory planning, phonemes and sequences at single-neuron scale \citep{khanna2024s0467}, but do not establish the same information access in a chronic communication system.

Placement determines which tissue and neural populations are observed. Scalp, cortical-surface, depth and intracortical configurations therefore differ in coverage as well as access requirements (\cref{tab:signal_resolution_comparison}). Source localisation and cross-participant latent alignment can reduce mismatch between recordings, but cannot create direct measurements of populations that were not sampled. The clinical configurations in \cref{tab:attempted} illustrate why implant choice and participant condition must be considered together.

Simultaneous recording allows configurations to be compared under the same behavioural conditions. Scalp, around-ear and in-ear EEG recorded during one auditory-attention task showed that recording configuration and reference choice affect decoding \citep{geirnaert2025ear}. Complementary intraoperative work demonstrated speech decoding with a dense cortical-surface microelectrode interface \citep{lehner2026layer7}. The first comparison informs sensor placement and referencing; the second establishes feasibility for an acute cortical recording route.

\paragraph{Comparing recording routes.} Signal selection couples a linguistic target to an access and supervision requirement. Reviews of imagined-speech reconstruction and EEG--fNIRS acquisition explain the range of non-invasive options \citep{jiye2024s0442,motaqi2026reviewr72}; implantable-speech reviews relate cortical access to patient selection and training \citep{jhilal2025reviewr15}. Hybrid sensing adds another design choice: peripheral information may improve communication even when it contributes no neural evidence. Broader silent-speech sensing reviews include these peripheral routes \citep{tang2026reviewr56}. Channel-specific comparisons are needed to determine what combining them with neural recording actually adds.

\paragraph{Choosing acquisition around the target.} Three questions narrow a recording choice before model selection: does the configuration sample the relevant population, does its temporal support match the target, and can that access be sustained in the intended setting? A phoneme-timed speech output and a semantic account of a long passage require different temporal information. Greater spatial coverage can aid representation learning while introducing setup and access costs; denser sampling of one region cannot substitute for sampling the relevant process. The conditional trade-offs in \cref{tab:acquisitionchoices} therefore concern configurations and objectives, not a global ordering of signal families.

\begin{table*}[!htbp]
\centering\small
\caption{Acquisition choices change the learning and use problem. The trade-offs synthesise the cited configurations; they are not controlled estimates of modality superiority.}
\label{tab:acquisitionchoices}
\begin{tabular}{@{}p{2.5cm}p{4.0cm}p{4.1cm}p{4.7cm}@{}}
\toprule
Configuration & Useful information access & Main trade-off & Consequence for a language system \\
\midrule
Intracortical / cortical-surface recording & Local speech-related activity with fast temporal features \citep{willett2023s0300,metzger2023s0299} & Surgical access and participant-specific coverage; chronic and acute data are distinct & Test calibration and stability for the actual implant and user, rather than extrapolating from acute decoding. \\
Scalp EEG & Time-resolved potentials; reading data with eye tracking \citep{hollenstein2018s0778} & Distributed mixtures, electrode contact and peripheral activity & Pair neural decoding with eye/muscle controls and calibration under the intended behaviour. \\
MEG & Time-resolved magnetic responses; reusable listening recordings \citep{alexandre2023s0322,miran2025s0576} & Sensor geometry, head movement and system-specific magnetic environment & Test content and participant transfer with the acquisition setup explicitly specified. \\
fMRI & Distributed BOLD patterns carrying contextual language information \citep{tang2023s0356} & Delayed response and scanner-constrained acquisition & Match targets to response support; distinguish semantic fidelity from feedback latency. \\
Optical haemodynamics & Cortical haemoglobin responses through fNIRS / HD-DOT \citep{suyi2024s0450,schroeder2023s0343} & Optical coverage, motion and delayed vascular response & Separate cue-conditioned language information from reference inputs and operating delay. \\
Functional ultrasound & Selected human language mapping and motor-decoding configurations \citep{soloukey2019s0116,rabut2024s0437} & Access differs between cranial-window recording and intact-skull feasibility & First establish held-out language-content decoding for the proposed access route. \\
\bottomrule
\end{tabular}
\end{table*}

\subsection{Signal quality and acquisition constraints}

Confounds depend on both task and acquisition. Articulated conditions introduce movement, muscle activity and possible acoustic contamination; Reading adds gaze and fixation timing. Inner instructions reduce intended movement but do not measure compliance. Auxiliary EOG, EMG or motion channels can test these explanations, while deliberate EEG--EMG fusion uses peripheral activity as part of the communication signal \citep{nieto2022s0292,hollenstein2018s0778,inoue2026s0725}. The same channel can thus serve as a control or a useful input, depending on the claim. Neural specificity requires separating these roles.

Acquisition controls are most informative when they follow the physical measurement. Multicentre analyses and recording-chain experiments show that sound-related vibration can introduce speech-correlated components into invasive signals during production or sound perception \citep{roussel2020acoustic,bush2022artifacts}. Synchronous sound and movement measurements, frequency-specific analyses and hardware controls help separate these components from physiology. Among 22 reports in selected decoding and dataset comparisons, six of 13 involving participants with intact speech report an acoustic-contamination check \citep{angrick2021s0221,verwoert2022s0248,meng2023s0312,zhang2024s0376,chen2024s0381,feng2025s0497}, as do four of nine involving audible attempts by participants with impaired speech \citep{moses2021s0220,card2024s0386,angrick2024s0460,wairagkar2025s0504}. The count includes checks explicitly presented by their authors as addressing contamination: audio--neural correlation, separate recording chains, non-audible controls or a statement that contamination was ruled out. The \href{https://github.com/Uploading-Inc/Neural-to-Text-Research/blob/main/silent_speech/paper/evidence/yz_comments_20261012/acoustic_controls.md}{accompanying report lists} identify the selected comparisons, exclusions and source passages; reports may share participants. Reporting a check does not establish its effectiveness, and absence of a report does not establish contamination. Optical measurements require a different separation: short-separation channels can estimate superficial physiological activity, and controlled comparisons show how optode density, noise regression and image reconstruction affect decoding \citep{fischer2026fnirs}. These adjacent methodological results guide signal preparation without serving as language-performance estimates.

\paragraph{Implications for method choice.} A model must accommodate both the information in the recording and the timescale on which it arrives. Electrophysiological decoders in the reviewed systems use time-varying features for phoneme, acoustic or sequence targets, whereas haemodynamic decoding must account for the delayed vascular response when aligning language and measurements \citep{willett2023s0300,anumanchipalli2019s0127,tang2023s0356}. Recording coverage and access also shape the available supervision: clinical production systems, naturalistic listening corpora and reading datasets provide different paired observations. A useful decoder must therefore match its target representation and training supervision to the information and timing of the chosen recording.

The supplementary mechanism map (\cref{fig:signal_mechanisms}) distinguishes physiological response and access constraints that the sampling configurations in \cref{tab:signal_resolution_comparison} alone cannot express.

\section{Methods}
\label{sec:methods}

Decoding methods distribute the reconstruction problem between neural measurements, paired training data and a model of plausible outputs. Their central design choice is which information the neural representation must preserve: distinctions among candidates, a linguistic sequence, a voice trajectory or a semantic relationship. Supervision determines which of these mappings can be learned, and the model prior supplies structure that the recordings do not fully specify. Shared benchmarks now support comparisons across successive methods in clinical text decoding, non-invasive retrieval and generation, speech synthesis and neural pretraining. These comparisons reveal which design choices improve a common task, while differences in test inputs and training resources explain why the same method can perform differently across settings.

\subsection{Data and supervision}
\label{sec:data}

\subsubsection{Experimental protocols and datasets}

\label{sec:datasets}

Datasets supply both the examples from which a decoder learns and the experimental contrasts through which its behaviour can be understood (\cref{tab:dataset_design}). Multi-session recordings support study of day-to-day variation; matched overt conditions supply acoustic or linguistic supervision; peripheral EOG/EMG channels help characterise task execution \citep{nieto2022s0292}. Recording vocalised, mouthed and internal production of the same items adds a direct basis for cross-task learning \citep{ma2025s0486,he2025s0633}. These contrasts depend on the elicited behaviour: silent articulation includes attempted articulatory movement, whereas imagined speech need not. SS-EEG, for example, instructs articulation attempts without sound \citep{dataset_zhou2025_6cc0f}.

\begin{table*}[!htbp]
\caption{Selected datasets provide complementary task contrasts and forms of supervision. Counts refer to the cited releases; the two Moreira experiments and two ZuCo releases are listed separately. Task conditions, auxiliary measurements and annotations determine which comparisons each resource supports. The complete inventory appears in \cref{tab:dataset_inventory}.}
\label{tab:dataset_design}
\centering\small\setlength{\tabcolsep}{4pt}
\renewcommand{\arraystretch}{1.13}
\begin{tabular}{@{}p{3.7cm}p{1.2cm}p{10.7cm}@{}}
\toprule
Resource & N & Task contrasts, auxiliary measurements and supervision \\
\midrule
Thinking Out Loud \citep{nieto2022s0292} & 10 & EEG with EOG and lip EMG; overt speech, imagined speech and visualised-direction trials share four direction classes, allowing task contrasts and movement checks. \\
Chisco \citep{zhang2024s0394} & 3 & High-density EEG during reading followed by imagined production of the same Chinese text; paired phases and 39 semantic categories connect comprehension with prompted production. \\
Moreira articulation corpus \citep{moreira2025s0505} & 8; 16 & Two 64-channel EEG experiments use phonemes, syllable pairs, real words and pseudowords, with perception/production tasks and control versus transcranial magnetic stimulation conditions. \\
3M-CPSEED \citep{ma2025s0486} & 20 & EEG for the same ten Pinyin items during overt, mouthed and imagined production; shared class labels support comparisons across articulatory conditions. \\
Aguilera paradigm comparison \citep{aguilerarodrguez2025s0502} & 15 & EEG during the same four imagined commands under conventional cues and a maze-game protocol; user-experience and affect questionnaires characterise the elicitation context. \\
VocalMind \citep{he2025s0633} & 1 & sEEG from Mandarin words and sentences in vocalised, mimed and imagined conditions; overt speech supplies an acoustic reference for cross-condition decoding. \\
LibriBrain \citep{miran2025s0576} & 1 & Multi-session MEG during audiobook listening, with aligned audio, word and phoneme annotations and session-based training, validation and test partitions. \\
ZuCo 1.0 / 2.0 \citep{hollenstein2018s0778,hollenstein2019s0779} & 12 / 18 & Simultaneous EEG and eye tracking during natural and task-directed English sentence reading; fixations link neural responses to the presented words. \\
SS-EEG \citep{dataset_zhou2025_6cc0f} & 12 & EEG during instructed silent articulation of 24 English words in six semantic groups; 16 sessions per participant support word/semantic classification and cross-session evaluation. \\
\bottomrule
\end{tabular}
\end{table*}

\begin{figure*}[!htbp]
\centering
\input{evidence/dataset_duration_redesign_20260921/figure7_onepage.tex}
\caption{\textbf{Task-related neural recordings and paired-material coverage across datasets.}
Each dataset or study occupies one row, ordered by release year (or report year); N describes the shown neural cohort.
Left: task intervals summed across participants and repetitions. Right: filled markers show material coverage, counting each source once; text is duration-equivalent coverage using the stated window per distinct prompt, rather than measured stimulus-audio duration.
The open CI/NH marker estimates one-session stimulus exposure (about 0.14 h), including repeated sentences; it is not unique-source duration.
Coloured squares denote signals; multiple colours share a row without summing their durations. BABA has separate MEG and fMRI cohorts: N and neural hours are stated per cohort. Both axes use identical logarithmic hour scales. Lower bounds can represent a verified component; audio includes aligned task windows.
Shared material across participants and simultaneous modalities is not additive. The full resource inventory appears in \cref{tab:dataset_inventory}.}
\label{fig:dataset_duration}
\end{figure*}

Perceived tasks provide stimulus transcripts, acoustic timing or visible text. Listening corpora and repeated reading datasets support reusable benchmarks, whereas clinical production data may require participant-specific collection. ZuCo combines EEG with eye tracking; LibriBrain pairs one participant's MEG with 52.32 hours of listening stimuli \citep{hollenstein2018s0778,hollenstein2019s0779,miran2025s0576}. Across signal families, task-related recording totals and paired-material coverage span several orders of magnitude (\cref{fig:dataset_duration}). Repeated presentations and additional participants increase the available neural observations without necessarily introducing new linguistic material. Material coverage therefore complements recording hours and participant counts when assessing the resources available for learning. The inventory in \cref{tab:dataset_inventory} provides the task designs, recording scopes and access conditions.

Recording depth and participant breadth support different kinds of learning. LibriBrain100 extends listening-MEG data along both dimensions, enabling prolonged within-person training and comparison across listeners \citep{francesco2026s0727}. Protocol diversity adds another resource: post-imagery overt responses can help check task compliance, while repeated reading and recall sessions allow stability to be studied with familiar content \citep{aguilerarodrguez2025s0502,zhang2026s0792}. These resources support complementary research questions. Long recordings from one person support fitting a detailed personal mapping; many participants support testing shared representations; repeated sessions expose recording drift; distinct stories or sentences support tests of new content. Repetitions improve estimation of a response but add less linguistic diversity than new material. For transfer, the availability of matched tasks and languages can consequently matter as much as total hours.

\subsubsection{Target labels and temporal alignment}

Overt audio provides acoustic and phonetic targets, although alignment remains an explicit processing step. Mouthed and Inner conditions do not provide the same simultaneous audible reference; reviewed systems use prompted content, paired Overt recordings or transferred models \citep{anumanchipalli2019s0127,angrick2021s0221,youngeun2023s0372}. In Reading, fixation markers can define word-aligned inputs, while marker-free methods pose a different segmentation problem \citep{duan2023s0782}. An expected prompt identifies the requested content; it does not prove that every trial followed it.

Alignment can be learned during model fitting. Iterative alignment discovery uses dynamic time warping within neural voice-activity training to accommodate uncertain response onset and duration \citep{rabbani2024alignment}. This offers a constructive way to learn from imperfectly timed labels. Because the alignment is part of training, it need not supply a reference waveform for the test utterance; reference-dependent test alignment has a different information requirement.

Training supervision must be distinguished from information required during testing. Overt-reference temporal alignment in silent-reading reconstruction \citep{martin2014s0032} and DTW reference alignment in EEG synthesis \citep{xiong2025s0617} constrain conclusions about autonomous operation. Haemodynamic models must additionally account for the relation between stimulus time and measured response \citep{tang2023s0356}. The reported test protocol should state which boundaries, alignments and transcripts are supplied to the decoder.

\SSAlignmentMiniFigure

\subsubsection{Preprocessing and artifact handling}

Filtering, referencing, epoching and artifact handling define the input available to a model, rather than being interchangeable implementation details. EEG reviews describe heterogeneous feature and preprocessing pipelines \citep{panachakel2021review,lopezbernal2022review}; the selected invasive and MEG systems likewise use different neural features and temporal windows \citep{willett2023s0300,alexandre2023s0322}. The appropriate pipeline follows the signal physiology, the target timescale and the information permitted during inference.

Preprocessing determines which regularities reach the learner. A multiverse analysis of EEG event-related-potential experiments showed that artifact correction could lower classification accuracy, consistent with models exploiting structured non-neural variation \citep{kessler2025preprocessing}. This explains why signal cleaning and prediction performance should be assessed together: the best pipeline for a neural interpretation may differ from the one that most easily separates the recorded conditions. Speech-specific recording-chain controls address acoustic contamination through a separate physical mechanism \citep{roussel2020acoustic}.

A reproducible comparison specifies filter causality and temporal support, referencing or channel selection, rejection criteria, auxiliary channels and normalisation. Learned transforms and feature selection should be fitted within the training partition. Artifact removal should be accompanied by residual-confound controls, rather than treated as proof of neural origin. A pipeline using future samples or reference-dependent alignment may remain useful offline, but its latency and information access differ from a causal communication system.

\subsubsection{Data partitioning and leakage prevention}

Data partitioning specifies the kind of reuse a model is expected to support. Within-person fitting, transfer across sessions and prediction for a new user are distinct learning settings. Cross-validation and hyperparameter selection can be organised around the corresponding independent unit, keeping validation separate from the final test \citep{varoquaux2017decoders}. This makes a performance estimate answer a useful deployment or scientific question.

Partitions should be defined before overlapping windows, repeated items or learned preprocessing can link training and test observations. The split unit follows the claim: trial-held-out tests assess new observations, stimulus-held-out tests assess new material, and session- or participant-held-out tests assess transfer to those conditions. Each split removes a different source of familiarity, so the split unit is part of the generalisation claim (\cref{sec:eval}). Audits show how such links and related shortcuts arise in practice: participant splits of reading EEG and narrative fMRI kept test sentences in training \citep{congchi2025s0600}, and in a variable-length MEG decoding diagnostic, signal-blind noise exploited sentence-duration cues in next-token prediction \citep{zhang2026s0793}. \Cref{tab:leakage_routes} summarises these and other documented routes, including some that bypass the partition itself, such as checkpoint selection, recording order and test-time prefixes.

\begin{table}[!htbp]
\caption{Leakage routes documented in decoding studies. Examples illustrate each route rather than tally its prevalence; one study can contain several routes, and the safeguards address the listed routes.}
\label{tab:leakage_routes}
\centering\footnotesize\setlength{\tabcolsep}{3pt}
\begin{tabular}{@{}>{\raggedright\arraybackslash}p{2.9cm}>{\raggedright\arraybackslash}p{7.3cm}>{\raggedright\arraybackslash}p{6.0cm}@{}}
\toprule
Route & Example in the literature & Safeguard \\
\midrule
\multicolumn{3}{@{}l}{\textit{Shared material between training and test}}\\
Repeated stimuli across partitions & Participant splits of ZuCo reading EEG and Narratives fMRI kept test sentences in training; UniCoRN's ZuCo BLEU-1 fell from 58.1 on the original test set to 18.5 on an additional leakage-free one \citep{congchi2025s0600}. & Assign each sentence or stimulus segment to one partition across participants, by text hashing or a stimulus-identity key \citep{dascoli2025s0623,zhang2026s0793}. \\
Participant overlap in a cross-participant claim & Story-level or pooled-recording splits kept the test participants' recordings in training \citep{congchi2025s0600}. & Group folds by participant, e.g.\ leave-one-participant-out \citep{li2026s0726}; also hold out text for a joint claim. \\
Overlapping or adjacent samples & Overlapping multi-frame fMRI samples placed in different partitions shared text \citep{congchi2025s0600}; within-participant leave-one-sample-out estimates were optimistic because samples are correlated \citep{varoquaux2017decoders}. & Split before windowing, keeping whole stories in one partition \citep{congchi2025s0600} or pruning training and validation windows that overlap test windows \citep{zhang2026s0793}; leave out whole blocks, sessions or participants \citep{varoquaux2017decoders}. \\
\addlinespace[3pt]
\multicolumn{3}{@{}l}{\textit{Test data used in fitting or selection}}\\
Tuning, early stopping or normalisation & A leave-one-participant-out speech-EEG benchmark used the held-out participant's accuracy for early stopping \citep{owais2026s0687}; regularisation tuned in the scoring loop \citep{varoquaux2017decoders} and normalisation fitted with held-out data \citep{li2026s0726} bias the estimate. & Fit scaling and alignment within training folds, report every seed and correct across candidate pipelines \citep{li2026s0726}; select on separate validation data or in a nested loop \citep{varoquaux2017decoders}. \\
\addlinespace[3pt]
\multicolumn{3}{@{}l}{\textit{Shortcuts that bypass the neural response}}\\
Recording order aligned with class & In an EEG study of unspoken (imagined) words, blocked presentation gave 45.5\% recognition among five words, which fell to chance with sequential or random order \citep{anne2009s0006}. & Randomise or interleave classes within runs; class-homogeneous blocks can confound class with recording time. \\
Input format revealing the target & In a variable-length MEG decoding diagnostic, padding and attention masks exposed sentence duration, so Gaussian noise reached 66.3\% next-token R@1 against 90.6\% for real MEG \citep{zhang2026s0793}. & Fixed-duration windows; signal-blind inputs passed through the identical pipeline \citep{zhang2026s0793}. \\
Correct prefixes at test time & Test-time teacher forcing markedly inflated ZuCo EEG-to-text scores \citep{jo2025noise}. & Report autonomous generation as the primary result (\cref{fig:test_prefix_information}). \\
Language prior or memorised text & With text leakage, shuffled brain input left reading-EEG and fMRI scores largely unchanged \citep{congchi2025s0600}; several EEG-to-text models scored similarly or better on noise \citep{jo2025noise}. & Require a significant margin over noise- and shuffled-input baselines \citep{jo2025noise,congchi2025s0600}. \\
\bottomrule
\end{tabular}
\end{table}

Preprocessing, supervision and architecture interact: a model can exploit differences introduced by the pipeline before it learns a linguistic distinction. EEG and end-to-end reviews accordingly connect features, task design and neural assumptions with model choice \citep{rahman2024reviewr02,zhang2025reviewr04,jin2025reviewr12,k2026s0678,yang2025reviewr29,estrellaibarra2026reviewr19}. Extending this perspective from imagery classification to language generation requires identifying what each output route asks the neural representation to preserve \citep{qiu2025reviewr08,yu2025reviewr17}.

\subsection{Output types and decoding objectives}

\label{sec:outputs}

Content decoders present candidate selections, text, audible speech or facial animation. Text may target verbatim wording or semantic paraphrase; open vocabulary describes the task, while spectrograms, phoneme posteriors and semantic embeddings are intermediate targets. Facial animation conveys speech-related expression rather than arbitrary images, and systems can combine forms \citep{metzger2023s0299}. \Cref{tab:outputs} summarises the associated evaluation questions.

Output form, neural attribution and practical use remain separate axes: controlled classification may provide stronger evidence of neural information than poorly controlled generation, and intelligibility is a quality of speech output rather than its form.

\begin{table}[!htbp]
\caption{Content-output forms and associated evaluation questions. Auxiliary control and study scope are recorded separately.}
\label{tab:outputs}
\centering\footnotesize\setlength{\tabcolsep}{3pt}
\begin{tabular}{@{}p{3.0cm}p{4.0cm}p{9.2cm}@{}}
\toprule
Form & Objective & Interpretation details \\
\midrule
Candidate selection & classification or retrieval & class/candidate count, chance, top-$k$, repeated observations \\
Text decoding & character, word or sentence sequences & WER/CER; vocabulary and content limits; verbatim versus semantic target \\
Speech synthesis & audible speech, via acoustic or articulatory targets & intelligibility, listener protocol, naturalness, prosody and latency \\
Facial animation & speech-related visual expression & separate visual and audiovisual assessment; not text accuracy \\
\midrule
\multicolumn{3}{@{}p{16.2cm}@{}}{Speech-state detection and device control are auxiliary outputs. Mapping, representation analysis, dataset release and methodological audits describe study scope, not content-output forms.}\\
\bottomrule
\end{tabular}
\end{table}

\subsection{Neural features and learned representations}

\begin{figure*}[!htbp]
\centering
\includegraphics[width=\textwidth]{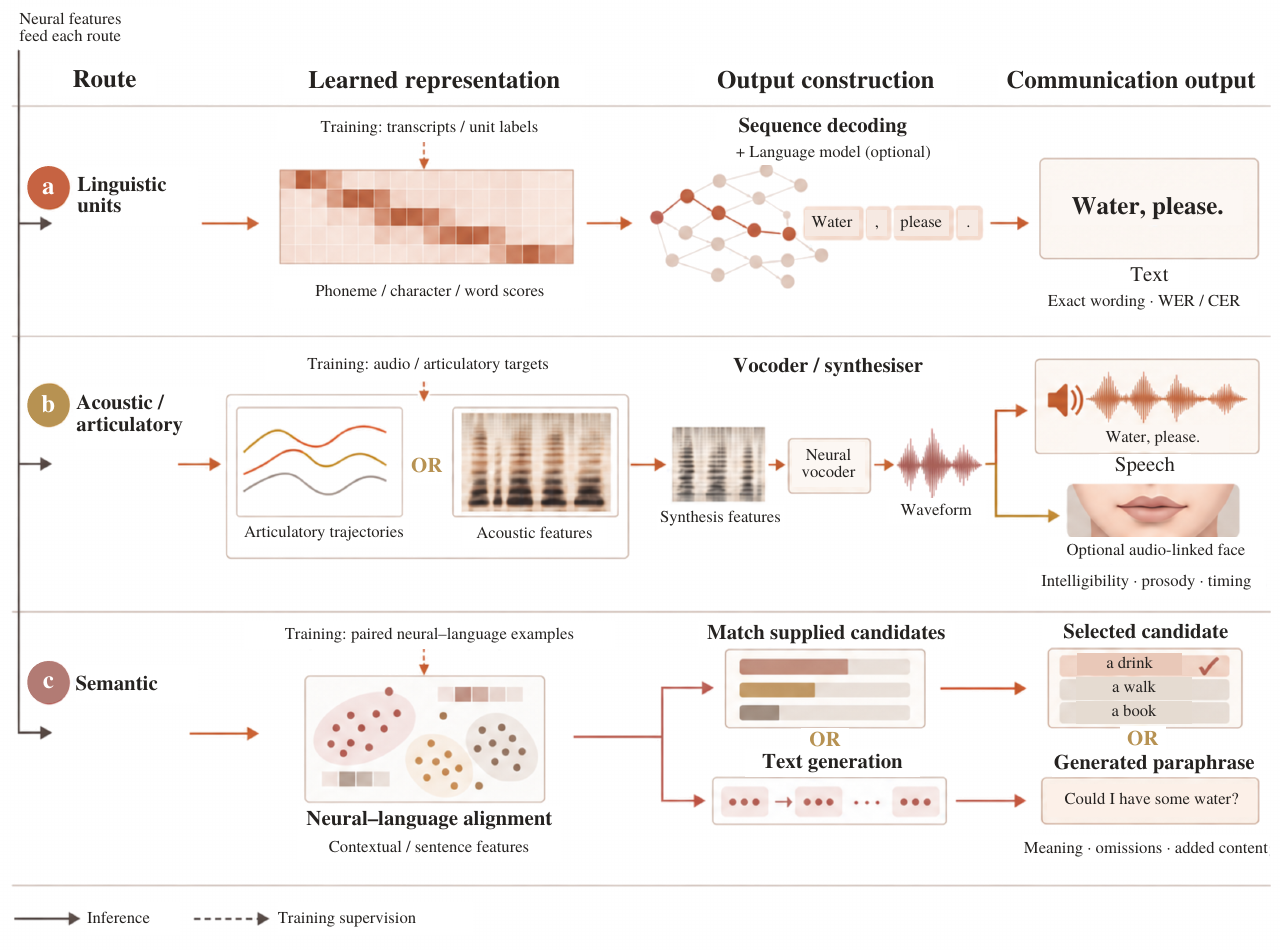}
\caption{\textbf{Three intermediate-representation routes connect neural features to communication outputs.} (a) Linguistic-unit scores support sequence decoding and word or character fidelity \citep{willett2023s0300}. (b) Articulatory or acoustic features support speech synthesis and optional audio-linked facial output \citep{anumanchipalli2019s0127,metzger2023s0299}. (c) Semantic alignment supports matching supplied candidates or generating text that can differ in wording while preserving meaning \citep{tang2023s0356}. Solid arrows indicate inference; dashed arrows mark training supervision, not test-time reference inputs. Patterns, trajectories, scores and messages are illustrations, not reported results. The evaluation terms identify distinct output qualities; routes can be combined.}
\label{fig:routes}

\end{figure*}

\label{sec:routes}

\Cref{fig:routes} summarises three recurring intermediate-representation routes, and \cref{tab:arch} compares their learned structure, supervision and trade-offs. They are alternative interfaces between neural measurements and output constraints, not successive stages of technical maturity; one architecture can serve different reconstruction objectives.

Intermediate targets also connect model design to physiology. Population recordings reveal articulator organisation and coordinated articulatory trajectories, while phoneme classifiers expose distinctions that can be composed into words \citep{bouchard2013organization,chartier2018kinematics,mugler2014s0033}. These findings motivate production-related targets for learning. When an articulatory trajectory is inferred from audio, it remains model-derived supervision rather than a direct measurement of the participant's movement.

Pretrained models supply useful structure at these interfaces. Speech representations such as wav2vec~2.0 provide alignment targets for MEG/EEG, whereas sequence decoders supply linguistic structure for Perceived--Auditory text generation \citep{alexandre2023s0322,wang2026s0711,yiqian2026s0748,yang2024neugpt}. This can reduce what must be learned from scarce paired neural data. It also makes two outcomes separable: a better completed output and a more informative neural representation. Shared latent spaces align neural populations across patients, whereas long recordings within one participant provide repeated observations for learning temporal structure \citep{spalding2026s0759,singh2025s0629,miran2025s0576,zhang2024s0394}. These approaches address different sources of data scarcity: variation in the neural populations sampled and limited coverage of linguistic experience.

\begin{table*}[!htbp]
\caption{Intermediate representations distribute the learning problem differently. The same neural architecture can serve several routes; the useful comparison is what structure it learns, which supervision it needs and how the output is constructed. Clinical attempted remains a source-task description.}
\label{tab:arch}
\centering\footnotesize\setlength{\tabcolsep}{3pt}
\begin{tabular}{@{}p{2.5cm}p{4.0cm}p{4.3cm}p{4.5cm}@{}}
\toprule
Mechanism / route & Structure made available & Supervision and principal trade-off & Representative conditions and systems \\
\midrule
Linguistic-unit decoding & Phones, phonemes or characters connect neural dynamics to a compositional lexicon. & Transcripts and sequence alignment; the language model resolves uncertainty but can alter wording. & Overt / clinical attempted ECoG and intracortical text \citep{herff2015s0044,moses2021s0220,willett2023s0300,card2024s0386,card2026s0729}; cued Inner \citep{kunz2025s0568}. \\
Sequence-to-sequence text & Learned temporal encoding connects recordings directly to linguistic sequences. & Paired sequences; learned alignment reduces hand specification while placing greater demands on data and partition design. & Overt ECoG / sEEG \citep{makin2020s0165,zhang2024s0376,feng2025s0497}; pretrained MEG sequence interfaces \citep{yiqian2026s0748,yang2024mad,yang2024neugpt}. \\
Articulatory / acoustic synthesis & Movement trajectories, spectra or acoustic tokens retain a route to voice and expression. & Audio or articulatory targets, often supplied by Overt training; transfer must bridge missing audible targets. & Articulatory synthesis \citep{anumanchipalli2019s0127,chen2024s0381}; spectral/vocoder routes \citep{jinuk2024s0420,youngeun2023s0372,xiong2025s0617,angrick2021s0221}. \\
Streaming voice & Causal sequence models couple ongoing neural input to incremental audio. & Paired participant recordings and synthesis targets; temporal context trades off against responsive feedback. & Clinical ECoG / intracortical synthesis \citep{littlejohn2025s0496,wairagkar2025s0504,wairagkar2026s0653}. \\
Acoustic / semantic alignment & Shared embeddings support candidate matching or constrain a language generator. & Paired stimuli or messages; the target embedding and candidate inventory determine which distinctions are preserved. & Speech-representation alignment \citep{alexandre2023s0322,wang2026s0711}; embedding retrieval \citep{dascoli2025s0623}; semantic generation \citep{tang2023s0356,ye2025s0549,suyi2024s0450,sparsh2025s0626}. \\
Discriminative features & Covariance, spatial filters, learned temporal features or neural populations distinguish a finite inventory. & Class labels; simpler objectives can suit small datasets but fix the target distinctions in advance. & Riemannian / SVM \citep{nguyen2018s0091,hons2026s0713,ikeda2014s0038}; CNN \citep{dash2020s0150,cooney2022s0236,ko2025s0558}; regional tokens \citep{jiang2026s0672}; population decoding \citep{wandelt2024s0464,khanna2024s0467}. \\
\bottomrule
\end{tabular}
\end{table*}

The task-dependent representations in \cref{fig:task_network_representation} provide the physiological basis for these computational interfaces. Local motor recordings can supply fast articulatory or phonemic structure; auditory recordings can align with the presented speech; distributed and slower responses can constrain meaning over longer context. The decoding objective selects which part of that information must survive the mapping to an output.

\paragraph{What the intermediate representation buys.} Route choice changes the errors users receive: language priors can alter wording; acoustic or articulatory routes preserve voice characteristics but depend on paired audio or movement targets that are difficult to obtain during silent production; semantic outputs can preserve gist while losing a negation, name or intended action. The selected route therefore determines what the model must learn and which errors matter to communication \citep{willett2023s0300,metzger2023s0299,anumanchipalli2019s0127,wairagkar2025s0504,tang2023s0356}.

\paragraph{Selection and generation solve different bottlenecks.} Retrieval makes the alternative set explicit and can be useful when only a small number of decisions is needed; its success is conditional on that set. Generation permits combinations beyond a supplied candidate list, but shifts part of the prediction problem into a learned prior. Moving from retrieval accuracy to generated text is consequently a change of task, not just a stronger score on the same task. A useful comparison holds the intended content and test information fixed, then asks whether the extra output freedom preserves the user's distinctions.

\subsection{Classification and retrieval}

\label{sec:imagined}

\paragraph{Which distinction is being decoded?} Classification is useful when a small set of alternatives matches the intended decision. It also tests whether a representation contains a particular distinction: selected ECoG electrodes separate internally articulated vowels, and low- and cross-frequency ECoG features carry Inner item or attribute information \citep{ikeda2014s0038,proix2022s0267}. The strength of that evidence is condition- and participant-specific. Eight-class cued Inner decoding from supramarginal single neurons reached online accuracies of 79\% and 23\% in two participants \citep{wandelt2024s0464}; mean and best accuracies for pairwise ECoG inner-word decoding are reported in \cref{tab:imagined} \citep{martin2016s0067}. A population code demonstrated in one configuration therefore does not establish uniformly usable control.

Spatial structure offers another architectural choice for classification. BrainStack combines seven regional convolutional experts with a global convolution--Transformer expert, using learned routing and global-to-local distillation to coordinate their predictions \citep{zhao2026s0795}. This makes regional specialisation and cross-region integration explicit within one model. Its EEG word-decoding task selects among predefined classes, so the relevant contribution concerns how those alternatives are discriminated.

\paragraph{Task contrasts are more informative than isolated peaks.} Reading versus Inner, preparation versus execution, and neural-only versus fused inputs test different explanations of a score. A five-word EEG/MEG comparison decoded silent reading more successfully than its mostly near-chance Inner conditions \citep{csaky2025s0624}. MEG preparation-window results, including 93\% for five phrases and 67.2\% for self-paced yes/no compared with 90.4\% during overt production, remain tied to forthcoming speech \citep{dash2020s0150,dash2024s0456}. Repeated-word EEG windows must additionally separate content from cues and overlapping observations \citep{jiang2026s0672}. For EEG--fNIRS, group means of 34.3\% and 77.3\% in different four-class studies are not a controlled improvement; the latter study's within-protocol finding of no fusion benefit over EEG alone answers a narrower, more useful acquisition question \citep{cooney2022s0236,hons2026s0713}.

Non-invasive classification can likewise probe a specific target without establishing sequence reconstruction. A 7-T fMRI study classified pairs of spoken phonemes at 53\% against 33\% chance in 15 participants \citep{vitria2024s0414}, while an OPM-MEG/EEG comparison examined time-resolved decoding of six Mandarin vowels in ten participants \citep{xu2026s0638}. Their value lies in testing access to these distinctions and their timing; sentence generation requires additional evidence about how distinctions compose.

\paragraph{Retrieval exposes the granularity of the target.} Contrastive speech alignment can identify a multi-second segment without resolving its individual words. Across four datasets and 175 participants, MEG/EEG alignment retrieved the correct 3-second segment among more than 1{,}000 candidates with top-1 accuracy of 41\% in the best MEG setting \citep{alexandre2023s0322}. Word-level retrieval in a larger pooled corpus of 723 participants, nine datasets, three languages and five million words reached top-10 of 37\% among 250 candidates against 4\% chance, including words absent from training \citep{dascoli2025s0623}. MEG, silent reading and test-time averaging were generally easier under that protocol. Hierarchical alignment to spectrograms, wav2vec2 and GPT-2 further tests representations at several linguistic or acoustic levels, with reported top-1 of 35.6\% for MEG and 20.0\% for EEG on SMN4Lang and SparrKULee \citep{wang2026s0711}. These studies broaden representation learning, but their candidate units and observation budgets remain part of the result: neither a larger candidate pool nor a higher top-$k$ score alone establishes finer content recovery. HD-DOT movie-segment retrieval provides an adjacent example \citep{tripathy2025s0586}; without a speech-specific target, movie identification is not Perceived--Audiovisual speech decoding.

\begin{figure}[!htbp]
\centering
\includegraphics[width=\textwidth]{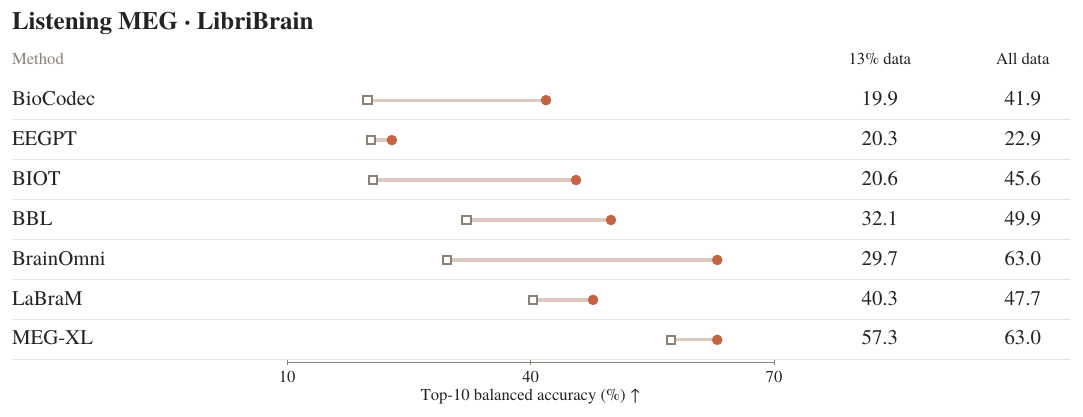}
\caption{\textbf{Listening-MEG word retrieval under two training-data budgets.} Pairs connect the 13\% and full-data conditions in MEG-XL's 50-word LibriBrain benchmark \citep{jayalath2026megxl}; squares and circles denote these conditions, respectively. Method references are BioCodec \citep{avramidis2025biocodec}, EEGPT \citep{wang2024eegpt}, BIOT \citep{yang2023biot}, BBL \citep{dulhan2024s0480}, BrainOmni \citep{xiao2025brainomni}, LaBraM \citep{jiang2024labram} and MEG-XL \citep{jayalath2026megxl}. Word onsets are supplied; pretraining resources and context lengths differ. Top-10 chance is 20\%. Exact values are printed beside each pair; three-seed SEMs appear in \cref{tab:meg_retrieval_values}.}
\label{fig:method_meg_retrieval}
\end{figure}

Long-context pretraining extends the retrieval route by reconstructing masked neural tokens from extended surrounding context, before the encoder is adapted to predict contextual word embeddings. MEG-XL uses this strategy for nearest-neighbour word selection with supplied word onsets and fixed candidate vocabularies \citep{jayalath2026megxl}. Its contribution is improved data efficiency and contextual representation learning within retrieval. This illustrates how a broader temporal context can improve a neural encoder without changing the output task into sentence generation. The shared LibriBrain comparison in \cref{fig:method_meg_retrieval} shows the resulting data-efficiency difference: MEG-XL reaches 57.3\% top-10 balanced accuracy with 13\% of the training data and 63.0\% at the full budget, where BrainOmni also reaches 63.0\%. These results favour long-context pretraining at the smaller budget while showing that the ordering can change as more paired data become available.

\paragraph{Negative evidence.} A leakage-audited benchmark of cross-subject auditory-evoked EEG vowel perception in 16 participants found a best balanced accuracy of 21.5\% against 20\% chance, with no implementation significant after correction \citep{li2026s0726}. A source-attribution audit of MEG decoding found that, in a variable-length next-token diagnostic, signal-blind noise reached 66.3\% R@1 against 90.6\% for real MEG; fixed-length inputs reduced these scores to 1.53\% and 5.10\%, respectively, against 0.05\% random chance. A separate test of MEG-to-audio retrieval with fixed-duration windows and stimulus-identity splits still preserved measurable stimulus-locked discriminability \citep{zhang2026s0793}.

Visual language separates tracking a representation from identifying content. Cortical tracking of inferred acoustic features during unfamiliar silent lip-reading supports a link between visual input and acoustic representations \citep{bourguignon2020s0163}. Word-identity decoding in MEG \citep{keitel2020s0174} and syllable or consonant decoding in auditory cortex, using fMRI in 64 participants and iEEG in 14 \citep{karthik2024s0391}, additionally test discrimination among linguistic alternatives. Their convergence motivates asking which representation carries the content; different target sets and recording conditions preclude a common accuracy comparison. These lip-reading tests also remain distinct from decoding written language.

\begin{table*}[!htbp]
\caption{Inner research develops content discrimination, cross-condition output generation and reusable protocols. Rows are grouped by contribution; task instructions, participant counts and chance levels define the tested condition. Study N can exceed condition N. Quantities are author-reported under heterogeneous protocols and are not a common ranking.}
\label{tab:imagined}
\vskip 0.1in
\centering\footnotesize\setlength{\tabcolsep}{3pt}
\begin{tabular}{@{}p{2.5cm}p{1.65cm}p{0.80cm}p{4.0cm}p{6.65cm}@{}}
\toprule
Study & Signal & N & Task and output & Reported result and its boundary \\
\midrule
\multicolumn{5}{@{}l}{\textit{Content discrimination}}\\
\citep{martin2016s0067} & ECoG & 5 & six words, pairwise classification & Inner mean 58\% (best 88\%) vs. Overt 86\% and Perceived--Auditory 89\%; chance 50\% \\
\citep{nguyen2018s0091} & EEG (64 ch) & 15 & Inner vowels and words; Riemannian features & up to 70\% for three-word and 95\% for two-word classification; task-specific maxima, not group means \\
\citep{dash2020s0150} & MEG (306 ch) & 8 & five phrases; cue $\to$ prepare $\to$ speak; CNN & 93\% in the source-labelled imagery/preparation window and 96\% in Overt; preparation limits an Inner interpretation; chance 20\% \\
\citep{proix2022s0267} & ECoG (3\newline datasets) & 13 & Inner word-feature discrimination; decoding subset N=8 & median binary accuracy 60\% across 12 significant participant--representation pairs using cross-frequency features; chance 50\% \\
\citep{cooney2022s0236} & EEG + fNIRS & 19 & four-class Overt and Inner; bimodal CNN & group mean Overt 46.3\%, Inner 34.3\% (chance 25\%); best individual 87.2\% / 53\% \\
\citep{hons2026s0713} & EEG + fNIRS & 22 & Inner and Perceived /a, i, b, k/; SVM & Inner 77.3\%, perceived 76.1\% (chance 25\%); fusion not better than EEG alone \\
\addlinespace[3pt]
\multicolumn{5}{@{}l}{\textit{Cross-condition generation and auxiliary detection}}\\
\citep{angrick2021s0221} & sEEG (119 contacts) & 1 & overt-trained (100 words) synthesis transferred online to whispered and Inner & real-time audio produced but explicitly not yet intelligible; $r=0.62\pm0.15$ is overt cross-validation \\
\citep{youngeun2023s0372} & EEG (64 ch) & 6 & NeuroTalk: 12 words/phrases; GAN + vocoder; one unseen word & Inner CER 68.3\%, MOS 2.78; unseen-word CER 83.1\% \\
\citep{suyi2024s0450} & HD-fNIRS & 4 & MindSpeech: word-cloud-cued sentences; text-prefix-conditioned continuation & BLEU-1 26.5 / 30.8 / 19.1 / 23.0 versus shuffled-brain 22.4 / 25.3 / 18.9 / 20.6 (P1--P4; 0--100); significant in 3 of 4 \\
\citep{sparsh2025s0626} & EEG (64 ch) & 56 & 30 Spanish sentences, listen then imagine; masked autoencoder + BERT & 48.9\% is training masked-token accuracy; validation 27.3\%; not test-set sentence accuracy \\
\citep{xiong2025s0617} & EEG (45 of 64 ch) & 13 & four Mandarin disyllables; overt bridging; DTW alignment & 4-alternative listener accuracy 91.2\%, MOS 3.50; closed-set, not open-vocabulary intelligibility \\
\citep{ko2025s0558} & EEG & 19/6 & Inner versus idle detection; word/sentence datasets & mean offline accuracy 79.9\% / 81.3\%, macro-F1 0.754 / 0.770; additional pseudo-online tests, not word decoding \\
\addlinespace[3pt]
\multicolumn{5}{@{}l}{\textit{Paired resources, protocol comparison and replication}}\\
\citep{he2025s0633} & sEEG & 1 & VocalMind: Mandarin words/sentences; Overt, Mouthed, Inner & Inner mel correlation 0.358 / 0.608 (words/sentences), rising to 0.785 / 0.793 after DTW; six-fold CV with overt acoustic references \\
\citep{ma2025s0486} & EEG & 20 & 3M-CPSEED: pinyin in Overt, Mouthed, Inner & 80 raw EEG recordings; cross-mode comparison \\
\citep{aguilerarodrguez2025s0502} & EEG & 15 & paradigm-design comparison dataset & attention, fatigue and engagement effects \\
\citep{tates2026s0682} & EEG & 9--16 & replication across four Inner datasets (9, 15, 10 and 16 participants) & 36.1\% of participants reach significance at 99\% confidence, averaged across datasets; replicated scores 2--39 percentage points below original reports \\
\addlinespace[3pt]
\bottomrule
\end{tabular}
\end{table*}

\subsection{Text decoding}

Text decoding composes neural evidence into linguistic sequences. One family predicts phonemes or characters and uses a language model to assemble them into words; another learns an encoder--decoder mapping from recordings to a text sequence. Early chronic phoneme discrimination and a 50-word clinical ECoG system established useful building blocks and a sentence-level application, respectively (\mbox{\citealp{brumberg2011s0014}}; \citealp{moses2021s0220}). Larger-vocabulary systems combine these building blocks with richer sequence models. The distinction between an explicit linguistic intermediate and an end-to-end text objective helps explain their supervision and generalisation behaviour.

\paragraph{Composing text from neural units.} ECoG phone representations combined with an automatic speech recognition (ASR) back-end and language model enabled continuous phrase-to-text decoding in seven participants \citep{herff2015s0044}. Encoder--decoder learning subsequently treated neural-to-text mapping as a sequence-translation problem, reaching a best \wer of 3\% in a four-participant study with repeated sentences \citep{makin2020s0165}. Self-supervised pretraining on separate unlabelled overt ECoG recordings further improved decoding in a repeated 30--50-sentence setting \citep{yuan2024ecogpretraining}. These developments improve representation and sequence learning under a controlled inventory; held-out repetitions, unseen sentences and new participants remain distinct generalisation tests.

\paragraph{Language changes the useful intermediate.} A representation that separates articulatory units must also preserve distinctions that change lexical identity in the target language. Mandarin illustrates the resulting choice of targets: sEEG in two speakers classified seven places of articulation at 86.5\% from superior temporal gyrus and four lexical tones at 58.3\% from thalamus \citep{wu2024s0470}. These local classifications identify potentially useful features, whereas a five-participant iEEG study tested sentence recovery, decoding ten Mandarin sentences composed of 40 characters at 21\% \wer with 93\% tone classification \citep{zhang2024s0376}. Acoustic-inspired sentence decoders further connect speech features with logosyllabic text \citep{feng2025s0497,liu2023s0316}. The design implication is to test both the relevant linguistic distinction and composed text: tone accuracy cannot substitute for word accuracy, and a closed sentence inventory does not establish new-sentence composition.

Clinical text systems connect sequence decoding to communication without normal audible production. Phoneme-based models have expanded from constrained lexicons to large vocabularies, while rapid calibration makes participant-specific fitting more practical (\cref{tab:attempted}). Multi-user intracortical models address a complementary objective: reusing a decoder across people by learning user-specific projections and calibrating it for held-out users \citep{fogg2026s0639}. This shift from fitting each model in isolation to sharing linguistic structure is a concrete route towards reducing initial training requirements.

\begin{table*}[!htbp]
\caption{Clinical speech interfaces expand from compositional text to audible, expressive and sustained communication. Rows are grouped by contribution, retaining the source term attempted where residual articulation is unconfirmed. N is study- or condition-specific. T15 recurs across Card and Wairagkar reports, and BRAVO3 across Metzger and Littlejohn; rows cannot be summed as independent cohorts. Vocabulary, WER and word accuracy refer to the stated tasks.}
\label{tab:attempted}
\vskip 0.1in
\centering\footnotesize\setlength{\tabcolsep}{3pt}
\begin{tabular}{@{}p{2.5cm}p{1.75cm}p{0.65cm}p{4.1cm}p{6.6cm}@{}}
\toprule
Study & Signal & N & Task and output & Reported result and its boundary \\
\midrule
\multicolumn{5}{@{}l}{\textit{Compositional text and language access}}\\
\citep{moses2021s0220} & ECoG & 1 & attempted words $\to$ text; 50-word vocabulary & median 15.2 words/min, median \wer 25.6\%; median correctly decoded throughput 12.5 words/min; anarthria after brain-stem stroke \\
\citep{metzger2022s0266} & ECoG & 1 & attempted phonetic code-words $\to$ spelling & 29.4 characters/min, median CER 6.13\% with a 1{,}152-word vocabulary; online spelling \\
\citep{willett2023s0300} & intracortical (4 arrays) & 1 & phonemes $\to$ LM $\to$ text & 125{,}000-word \wer 23.8\%; study-wide mean 62 words/min; 50-word \wer 9.1\%; source-reported unvoiced attempts 24.7\% at 125{,}000 words \\
\citep{silva2024s0375} & ECoG & 1 & bilingual attempted speech $\to$ text; 178-word vocabulary & median online \wer 25.0\% overall (Spanish 26.7\%, English 22.2\%); language classification 87.5\% \\
\citep{card2024s0386} & intracortical & 1 & rapidly calibrating large-vocabulary text & 97.5\% word accuracy (aggregate \wer 2.5\%) and mean 31.6 words/min over the final five copy-task sessions with a 125{,}000-word vocabulary; sustained about 8 months \\
\citep{jude2026s0677} & intracortical & 1 & long-standing anarthria, locked-in, ventilated & online \wer 22\% (50 words, second session); best unconstrained-vocabulary block 48\%; offline \wer 39\% on day 68 \\
\addlinespace[3pt]
\multicolumn{5}{@{}l}{\textit{Audible output and expressive control}}\\
\citep{guenther2009s0004} & neurotrophic electrode & 1 & online formant control; vowel synthesis & first direct audio feedback; output space limited to vowels \\
\citep{angrick2024s0460} & ECoG & 1 & online synthesis of six self-chosen keywords & listeners identified 80\% of synthesised keywords; participant retained some voicing \\
\citep{littlejohn2025s0496} & high-density ECoG & 1 & streaming brain-to-voice (RNN transducer, RNN-T) & 80 ms decoding increments; increment size is not end-to-end latency \\
\citep{wairagkar2025s0504} & intracortical & 1 & instantaneous voice synthesis with prosody & median listener-transcription \wer 43.75\%; six-choice sentence matching 94.34\%; real-time emphasis and melody \\
\citep{wairagkar2026s0653} & intracortical & 1 & Brain2voice 2.0 multitask synthesis & mean listener \wer 5.24\% (continuous) / 5.65\% (tokenised); offline causal benchmark; 10 ms output step; preprint \\
\citep{srinivasan2026s0698} & intracortical & 2 & mouthed / whispered / normal / loud modes & mode and loudness classification 94\% and 89\%; not word accuracy \\
\addlinespace[3pt]
\multicolumn{5}{@{}l}{\textit{Multimodal expression and sustained use}}\\
\citep{metzger2023s0299} & 253-ch ECoG & 1 & text, personalised voice, avatar & median 78.3 words/min and median \wer 25.5\% in a 1{,}024-word condition; audio and avatar evaluated separately \\
\citep{brosler2026s0760} & 253-ch ECoG & 2 & parallel phrase text and preset avatar gestures & Bravo-6: vocalised attempts and imagined gestures, with 70\% phrase and 66\% gesture accuracy in real-time copy trials; 10 phrases and 10 gestures plus rest; 3 participants in movement mapping \\
\citep{card2026s0729} & intracortical & 1 & independent home use, speech and cursor & $>$3{,}800 h including idle; 1.96 M words; 56 words/min; 99.2\% word accuracy on cued copy; 92\% conversational ratings at least mostly correct, including user-corrected sentences \\
\addlinespace[3pt]
\bottomrule
\end{tabular}
\end{table*}

\begin{figure}[!htbp]
\centering
\includegraphics[width=\textwidth]{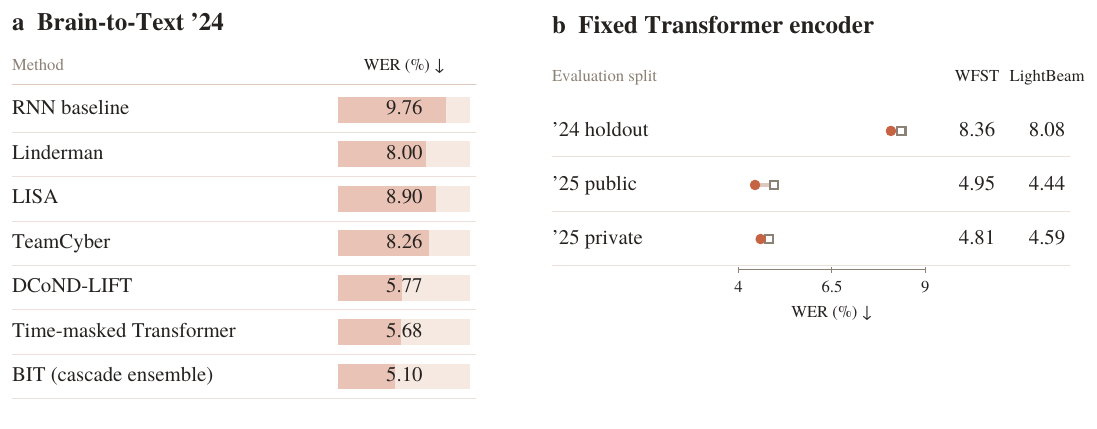}
\caption{\textbf{Attempted-speech comparisons separate complete systems from paired language decoders.} (a) Brain-to-Text '24 private-test WER: the RNN baseline follows BIT's comparison \citep{zhang2025s0489}; Linderman, LISA and TeamCyber use the reported benchmark values \citep{willett2024lessons}; DCoND-LIFT \citep{li2025s0508}, the time-masked Transformer \citep{feghhi2025timemasked} and BIT \citep{zhang2025s0489} use their strongest reported cascaded ensembles. These systems share 1,200 test sentences but differ in training and language resources. (b) LightBeam \citep{feghhi2026lightbeam} holds a Transformer encoder fixed within each pair: squares use a weighted finite-state transducer (WFST) to combine lexical and language-model constraints; circles use LightBeam. Language resources change from 5-gram/OPT to 4-gram/Llama, with benchmark-text adaptation; public and private partitions remain separate. Values are means across ten encoder seeds.}
\label{fig:method_attempted_benchmark}
\end{figure}

\paragraph{What shared attempted-speech benchmarks reveal.} Brain-to-Text '24 comparisons on a common private test set separate algorithmic gains from differences in participant recordings (\cref{fig:method_attempted_benchmark}). The retrospective attributes the strongest gains to complementary decoder predictions and language-model rescoring, alongside improved training and phonemic objectives \citep{willett2024lessons}. A later time-masked Transformer ensemble reports 5.68\% WER, and BIT lowers the cascaded ensemble result to 5.10\% after cross-task and cross-species encoder pretraining \citep{feghhi2025timemasked,zhang2025s0489}. BIT's end-to-end ensemble reaches 10.22\%, compared with 24.69\% for the earlier end-to-end system in its comparison. A stronger pretrained encoder therefore helps both frameworks, but the cascaded language-decoding stage remains advantageous on this benchmark. LightBeam subsequently compares the language-decoding stage with the neural encoder held fixed, improving WER across its paired GRU and Transformer settings while changing the language resources \citep{feghhi2026lightbeam}. Together, these comparisons locate progress at the interface between neural evidence and sequence inference, rather than attributing every gain to a larger neural encoder.

Perceived--Auditory MEG provides a complementary test of whether these interfaces can work with non-invasive recordings. Phoneme-level sentence reconstruction from ten hours per participant reported phoneme error of 62.2\% and sentence CER of 65.8\% \citep{vlez2026s0749}; a joint text-and-acoustic model proposes coupling two output targets \citep{jin2026s0743}. These approaches make the intermediate linguistic or acoustic constraint explicit. Generative sequence models additionally learn how to complete linguistic content, making held-out text and matched input controls central to the comparison.

\begin{figure}[!htbp]
\centering
\includegraphics[width=\textwidth]{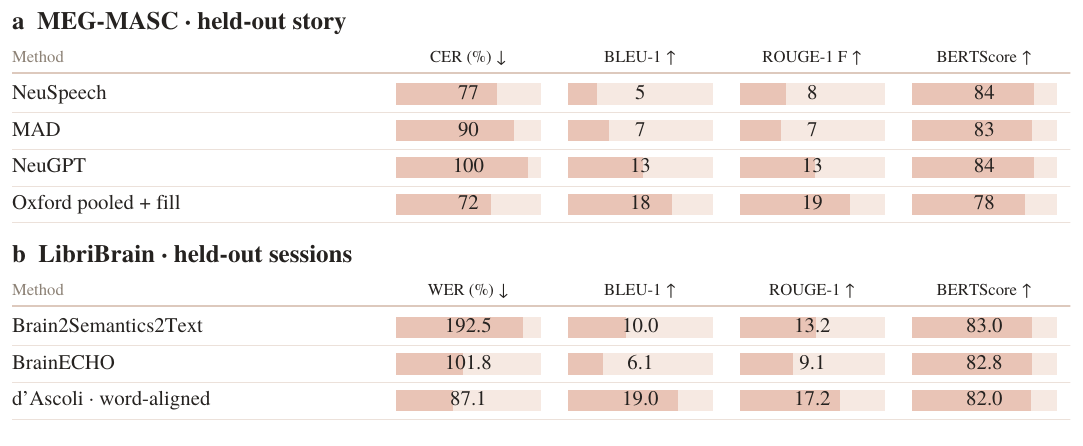}
\caption{\textbf{MEG-to-text methods share benchmark lineages but use different linguistic interfaces.} (a) NeuSpeech \citep{yang2024neuspeech}, MAD \citep{yang2024mad}, NeuGPT \citep{yang2024neugpt} and Oxford \citep{jayalath2025unlocking} follow Oxford's held-out-story comparison. Earlier scores are quoted by that paper; Oxford's pooled beam-and-fill configuration adds LibriBrain/Armeni training data and uses supplied word onsets and out-of-vocabulary positions. (b) Brain2Semantics2Text \citep{landau2026semanticbottleneck}, BrainECHO \citep{li2025brainecho} and d'Ascoli's word-aligned method \citep{dascoli2025s0623} follow the later LibriBrain comparison \citep{landau2026semanticbottleneck}. All receive sentence boundaries; the last also receives word timing. Metrics are displayed on a 0--100 scale, with WER allowed above 100. Each row uses one configuration. Matched noise results and SDs are in \cref{tab:meg_noise_values}.}
\label{fig:method_meg_generation}
\end{figure}

\paragraph{Generative MEG-to-text.}
Generative MEG models connect a neural encoder to a language or speech generator through three main interfaces: continuous neural features, discrete neural tokens, or predicted linguistic representations. The last route can operate on individual words or whole sentences. These choices determine how much timing information the decoder needs and whether its output is primarily constrained by lexical evidence or sentence meaning (\cref{fig:method_meg_generation}).

Direct sequence models adapt an existing generator to neural input. NeuSpeech uses a Whisper-style decoder and reports BLEU-1 of 60.30 on Gwilliams without test-time teacher forcing \citep{yang2024neuspeech}. This sentence--recording-pair split permits familiar stimulus sentences across partitions; holding out an entire story asks the model to recover new linguistic content. Under story holdout, MAD aligns neural and speech features and reports BLEU-1 of 6.86 and CER of 89.82\% \citep{yang2024mad}. NeuGPT instead compresses the recording into discrete neural tokens and incorporates these tokens into a Qwen2 language model. Its MEG-MASC comparison reports BLEU-1 of 12.92 and ROUGE-1F of 13.06, compared with 5.49 and 8.43 for NeuSpeech in the same table \citep{yang2024neugpt}. NeuGPT's CER is 99.8\%, showing that the gain in overlapping words does not translate into correspondingly accurate character sequences. The choice of neural interface and the held-out linguistic content therefore matter alongside the generator's capacity.

A word-level interface makes the lexical constraints on generation explicit. The Oxford system predicts contextual word embeddings from onset-aligned MEG, converts their vocabulary similarities into word probabilities, and rescores candidate sequences with a language model \citep{jayalath2025unlocking}. Its comparison extends the NeuGPT benchmark on the same held-out-story split, placing NeuSpeech, MAD, NeuGPT and word-based generation together (\cref{fig:method_meg_generation}a). The rounded BLEU-1 scores rise from 5 and 7 for NeuSpeech and MAD to 13 for NeuGPT and 18 for Oxford beam-and-fill decoding; CER falls from NeuGPT's 100\% to 71\%. Earlier baseline scores are quoted, whereas Oxford evaluates its own models with supplied word onsets and out-of-vocabulary positions. Pooling additional LibriBrain and Armeni training data retains BLEU-1 of 18 but changes CER to 72\%, so more training data does not improve every measure. On LibriBrain, a separate comparison replaces supplied out-of-vocabulary positions with a learned detector: BLEU-1 rises from 19 for greedy decoding to 22 after beam rescoring and 24 after in-filling, while CER falls from 77\% to 76\% and 69\%. Together, these results show how composing uncertain lexical predictions can improve text reconstruction, while making the timing and vocabulary information supporting that improvement explicit.

A sentence-level interface changes the target from individual words to their combined meaning. Brain2\allowbreak Semantics2\allowbreak Text maps MEG through a temporal encoder and attention pooling into a sentence-embedding space, then inverts the predicted vector into text \citep{landau2026semanticbottleneck}. This later Oxford study compares three methods on the same 172 held-out LibriBrain sentences from one participant (\cref{fig:method_meg_generation}b). Brain2\allowbreak Semantics2\allowbreak Text and BrainECHO retain sentence boundaries but remove word-level alignment, reaching BERTScores of 0.830 and 0.828, respectively; their matched noise controls score 0.819 and 0.825. A word-aligned implementation of d'Ascoli's method reaches 0.820, versus 0.794 with noise. WER orders the methods differently: 192.5\%, 101.8\% and 87.1\%, respectively. The semantic route's longer reconstructions contribute to its higher WER. These descriptive scores suggest a trade-off between sentence-level similarity and control of exact wording and length; the word-aligned route benefits from a more explicit lexical constraint. Both semantic quality and the improvement over matched input controls are needed to understand this trade-off.

\begin{figure}[!htbp]
\centering
\includegraphics[width=\textwidth]{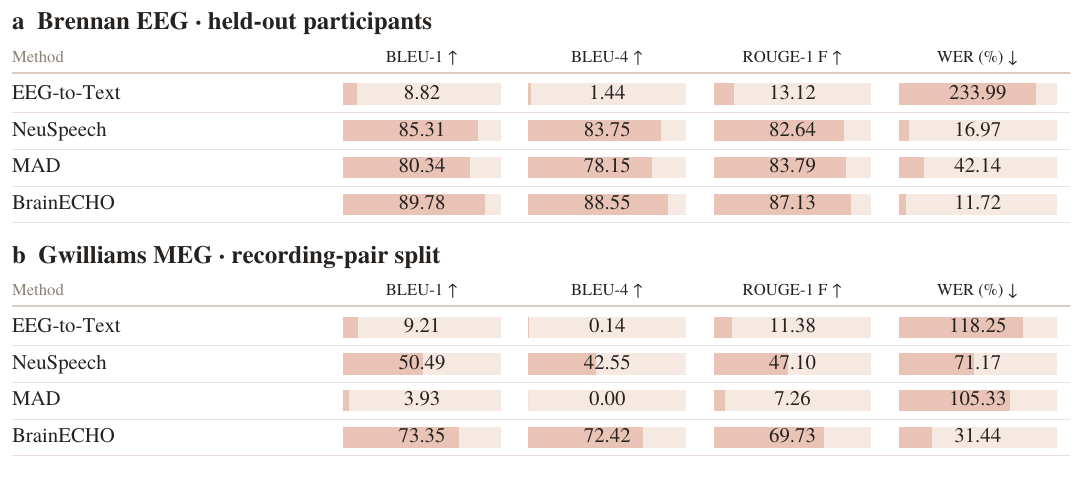}
\caption{\textbf{Listening benchmarks compare lexical overlap and edit errors within each recording split.} BrainECHO's published comparison \citep{li2025brainecho} evaluates EEG-to-Text \citep{wang2022s0780}, NeuSpeech \citep{yang2024neuspeech}, MAD \citep{yang2024mad} and BrainECHO \citep{li2025brainecho}. (a) Brennan EEG uses participant holdout; (b) Gwilliams MEG uses shuffled recording-pair splits. Values are the comparison authors' implementations, including their NeuSpeech rerun. Familiar stimulus texts can occur across participants or recordings, unlike the story holdout in \cref{fig:method_meg_generation}a. Each row reports a single configuration. Scores are displayed as percentages; WER can exceed 100\%.}
\label{fig:method_listening_generation}
\end{figure}

\paragraph{Acoustic intermediates across listening benchmarks.} BrainECHO inserts a vector-quantised spectrogram reconstruction between the neural encoder and Whisper, comparing this route with EEG-to-Text, NeuSpeech and MAD under shared evaluations \citep{li2025brainecho}. On Brennan EEG with held-out participants, the four implementations yield BLEU-1 scores of 8.82, 85.31, 80.34 and 89.78; the corresponding WERs are 233.99\%, 16.97\%, 42.14\% and 11.72\% (\cref{fig:method_listening_generation}). On shuffled Gwilliams MEG recording pairs, BrainECHO reaches BLEU-1 of 73.35 and WER of 31.44\%, compared with 50.49 and 71.17\% for the authors' NeuSpeech implementation. The acoustic intermediate thus improves both lexical overlap and edit accuracy in these comparisons. The story-held-out Oxford benchmark asks a different generalisation question: participant or recording-pair holdout can retain familiar stimulus text, whereas story holdout removes that source of linguistic familiarity. Its lower absolute scores cannot isolate the effect of this split because the models, implementations and supplied timing information also differ.

\subsubsection{Perceived--Visual: Reading EEG-to-text}
\label{sec:reading}

Reading-based decoding targets text that the participant sees, rather than speech that they hear or intend to produce. Visual stimulation, eye movements and language processing can all contribute to the recordings. Much EEG-to-text research relies on ZuCo, which combines EEG and eye tracking during English reading: 12 participants and 1{,}107 sentences in ZuCo 1.0 \citep{hollenstein2018s0778}, and 18 participants and 739 sentences in ZuCo 2.0 \citep{hollenstein2019s0779}. Fixation-based segmentation supplies word-aligned features in some pipelines; other methods process continuous segments without fixation markers. Reuse of these corpora across algorithm papers does not provide new participant cohorts.

\paragraph{Where the methods differ.} Methods address three recurring mismatches between EEG and a pretrained text generator (\cref{tab:reading}). First, neural recordings and linguistic tokens have different structure: contrastive learning, discrete encoders, masked autoencoding, cross-modal codebooks and byte-pair alignment seek a more usable interface \citep{feng2023s0781,duan2023s0782,wang2024s0785,tao2024s0786,zhou2024s0787}. Second, recordings vary across people and datasets; participant-dependent representations and EEG pretraining address this learning problem, while shared fMRI/EEG architectures and Chinese reading data broaden the settings examined \citep{amrani2024s0783,liu2024s0784,xi2023s0790,lu2025s0794}. Third, the language generator completes information missing from the neural representation. BART-based decoding and GPT-4 refinement illustrate that dependence on a text prior \citep{wang2022s0780,amrani2024s0783}. Each intervention changes a different part of the information chain: improved text overlap alone cannot show which mismatch has been resolved.

\begin{figure}[!htbp]
\centering
\includegraphics[width=\textwidth]{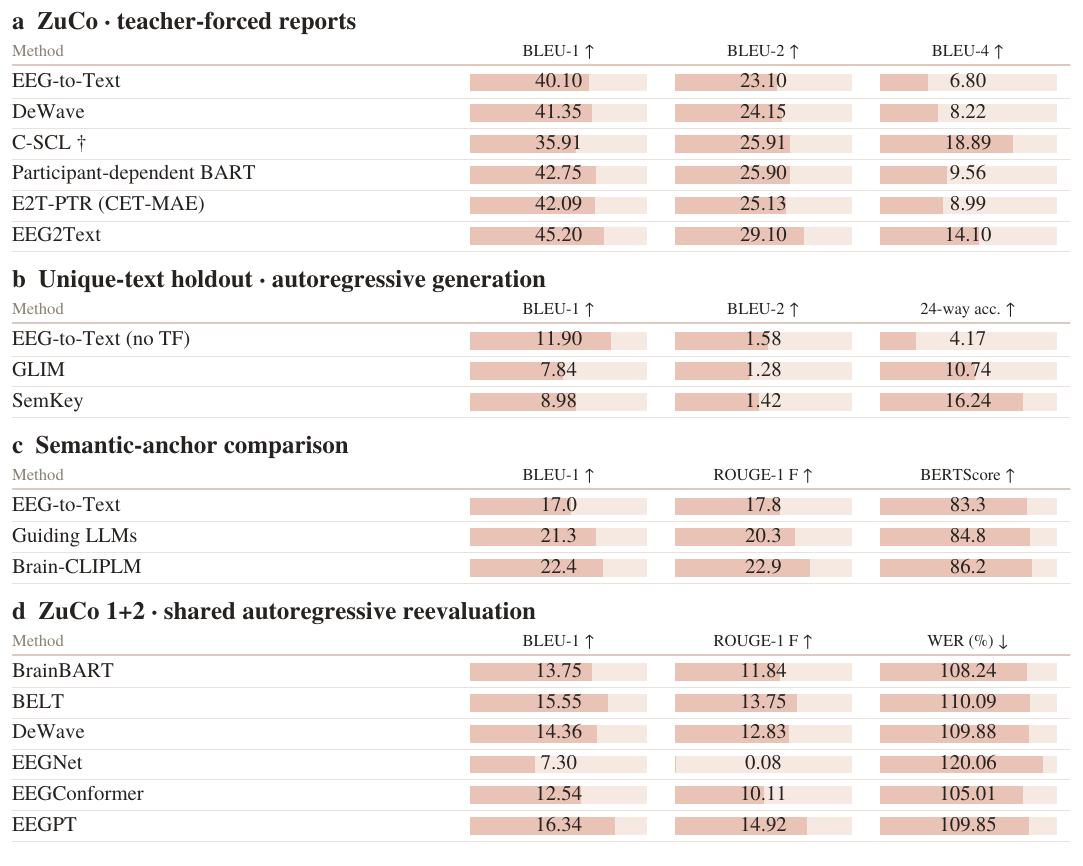}
\caption{\textbf{Reading EEG-to-text comparisons distinguish supplied prefixes from autonomous generation.} (a) Teacher-forced reports: EEG-to-Text \citep{wang2022s0780}, DeWave \citep{duan2023s0782}, C-SCL \citep{feng2023s0781}, participant-dependent BART \citep{amrani2024s0783}, E2T-PTR (CET-MAE) \citep{wang2024s0785} and EEG2Text \citep{liu2024s0784}. Training pools differ; the dagger marks C-SCL's removal of consecutive repeated output words. (b) EEG-to-Text, GLIM \citep{liu2025s0801} and SemKey \citep{wang2026s0791} use SemKey's unique-text autoregressive benchmark. Identification compares generated text with its target and 23 distractors; chance is 4.17\%. (c) Brain-CLIPLM \citep{yang2026s0806} uses a task-specific sentence corpus for reconstruction; its semantic-anchor comparison quotes earlier EEG-to-Text and Guiding LLMs \citep{zheng2026guiding} values. (d) COFETT's autoregressive reruns \citep{zhang2026s0792} compare BrainBART/EEG-to-Text, BELT \citep{zhou2024belt}, DeWave, EEGNet \citep{lawhern2018eegnet}, EEGConformer \citep{song2023eegconformer} and EEGPT \citep{wang2024eegpt}. Scores are displayed as percentages; WER can exceed 100\%. Panels are separate evaluation settings.}
\label{fig:method_reading_controls}
\end{figure}

\paragraph{Shared benchmarks and successive improvements.} Repeated use of ZuCo makes several method families directly traceable through published comparisons (\cref{fig:method_reading_controls}a). E2T-PTR combines EEG representations pretrained by CET-MAE with BART for EEG-to-text decoding. In its 10{,}710-training-sample comparison, BLEU-4 rises from 6.80 for EEG-to-Text to 8.22 for DeWave and 8.99 for E2T-PTR (CET-MAE) \citep{wang2024s0785}. Broader representation learning and decoder adaptation extend this lineage: the reported multi-view EEG2Text result is 14.10, and BELT-2 with a T5 decoder reports 17.95 \citep{liu2024s0784,zhou2024s0787}. The earlier teacher-forced results are shown separately in \cref{fig:method_reading_controls}a; SEE and BELT-2 retain their original reporting conditions in \cref{tab:reading}. Their differing training pools, test-prefix policies and output processing determine the strength of cross-paper comparisons. For example, C-SCL removes consecutive repeated generated words before scoring, whereas the original EEG-to-Text pipeline retains them \citep{feng2023s0781,wang2024s0785}.

\paragraph{Autonomous generation and neural contribution.} Supplying correct previous tokens during testing changes autonomous generation into prediction conditional on a correct prefix. A ZuCo evaluation compared BART-, Pegasus- and T5-based pipelines with real and noise inputs, both with and without test-time teacher forcing \citep{jo2025noise}. Comparable EEG/noise performance in the tested configurations shows why text overlap alone cannot establish a neural contribution. Separately, sentence overlap in inspected cross-participant splits leaves linguistic familiarity available even when the participant changes \citep{congchi2025s0600}. These are distinct shortcuts: removing correct prefixes does not remove repeated content, and holding out participants does not necessarily hold out text.

\paragraph{From token overlap to semantic recovery.} GLIM combines language modelling with contrastive alignment and multiple textual variants to learn representations that support autoregressive generation \citep{liu2025s0801}. SemKey adds explicit semantic attributes and evaluates this progression against EEG-to-Text under a unique-text split \citep{wang2026s0791}. In that common comparison, original-reference BLEU-1 is 11.90, 7.84 and 8.98 for EEG-to-Text, GLIM and SemKey, respectively; generated-text 24-way identification is 4.17\%, 10.74\% and 16.24\% (\cref{fig:method_reading_controls}b). The strongest n-gram baseline is therefore at chance on the semantic identification task. Higher overlap with a reference and better discrimination of its meaning are distinct forms of progress, and both are needed to interpret these methods.

A complementary line explicitly recovers semantic anchors before reconstructing a sentence. Guiding LLMs aligns EEG and language representations; Brain-CLIPLM then uses ordered keyword anchors, language-model reasoning and a task-specific sentence corpus \citep{zheng2026guiding,yang2026s0806}. Brain-CLIPLM's published comparison reports BLEU-1 of 17.0, 21.3 and 22.4 for EEG-to-Text, Guiding LLMs and Brain-CLIPLM, alongside ROUGE-1 F of 17.8, 20.3 and 22.9 (\cref{fig:method_reading_controls}c). The progression motivates intermediate semantic targets, while the corpus-assisted reconstruction condition remains central to interpreting what the neural signal supplies.

COFETT additionally re-evaluates reading decoders without correct test prefixes. On its ZuCo 1+2 comparison, BrainBART, BELT and DeWave produce BLEU-1 values of 13.75, 15.55 and 14.36, with WER above 108\% in all three cases; EEGNet, EEGConformer and EEGPT adaptations extend the same comparison to distinct neural encoders (\cref{fig:method_reading_controls}d) \citep{zhang2026s0792}. Discrete diffusion offers another output model: DELTA raises BLEU-1 from 13.69 for its autoregressive BART baseline to 21.93 under the same residual-vector-quantisation interface, while WER remains 110.03\% \citep{jeon2025s0798}. Improvements in local overlap have thus outpaced reliable sentence reconstruction in these autonomous tests.

A related but different task is COFETT's cued recall after reading: its central comparison measures Pearson correlation between EEG-derived and text embeddings across sessions with repeated sentences and noise controls \citep{zhang2026s0792}. This tests cross-session representation alignment on repeated content; it does not establish autonomous generation of novel sentences. The distinction connects reading benchmarks to internal recall without treating them as the same behavioural task.

\begin{table*}[!htbp]
\caption{Reading EEG-to-text methods, grouped by the interface between neural representations and language. Results retain the stated input and inference conditions; shared benchmark comparisons appear in \cref{fig:method_reading_controls}. BLEU, ROUGE and BERTScore values are on a 0--100 scale. TF denotes test-time teacher forcing (correct previous tokens). Classification, EEG-embedding retrieval and semantic identification of generated text are distinct evaluation tasks.}
\label{tab:reading}
\vskip 0.1in
\centering\scriptsize
\begin{tabular}{@{}>{\raggedright\arraybackslash}p{2.0cm} >{\raggedright\arraybackslash}p{2.1cm} >{\raggedright\arraybackslash}p{3.35cm} >{\raggedright\arraybackslash}p{3.95cm} >{\raggedright\arraybackslash}p{2.5cm}@{}}
\toprule
Study & Input / task & Intermediate / mechanism & Learning / inference conditions & Selected result \\
\midrule
\multicolumn{5}{@{}l}{\textit{Sequence decoders and output-model adaptation}}\\
\citep{wang2022s0780} & ZuCo EEG / text & Transformer EEG encoder $\rightarrow$ pretrained BART & Fixation-aligned word features; original evaluation uses TF & BLEU-1 40.1 \\
\citep{xi2023s0790} & ZuCo EEG / text & UniCoRN: snapshot and sequence reconstruction $\rightarrow$ text decoder & Two-stage representation pretraining; architecture also evaluated separately on listening fMRI & BLEU-1 57.68; BLEU-4 37.04 \\
\citep{amrani2024s0783} & ZuCo 1+2 EEG / text & Participant-dependent encoder $\rightarrow$ BART; optional GPT-4 refinement & Learned participant embeddings; sentence-based split; reported scores before GPT-4 refinement & BLEU-1 42.75; BERTScore-F 53.86 \\
\citep{lu2025s0794} & ChineseEEG / text & Character-level EEG encoder with MiniLM & Autoregressive character decoding; known sentence start and fixed-length input & BLEU-1 6.38 \\
\addlinespace[3pt]
\multicolumn{5}{@{}l}{\textit{Contrastive and masked representation learning}}\\
\citep{feng2023s0781} & ZuCo EEG / text & C-SCL: curriculum contrastive alignment & BrainBART-Large variant; semantic curriculum across participants & BLEU-1 35.91; WER 68.48\% \\
\citep{liu2024s0784} & ZuCo EEG / text & EEG2Text: regional and global transformers $\rightarrow$ BART & EEG pretraining on ZuCo and Image-EEG; v1 multi-view result & BLEU-1 45.2; BLEU-4 14.1 \\
\citep{wang2024s0785} & ZuCo EEG / text & E2T-PTR (CET-MAE): pretrained EEG encoder $\rightarrow$ BART & Paired-modality reconstruction and alignment; 10{,}710 training samples; TF & BLEU-1 42.09; BLEU-4 8.99 \\
\citep{zhou2024s0788} & ZuCo EEG / word classification; retrieval & ETER: masked contrastive encoder; keyword-based sentence retrieval & Word encoder predicts a 100-word vocabulary; training-free sentence search & Top-20 word accuracy 55.15\% \\
\addlinespace[3pt]
\multicolumn{5}{@{}l}{\textit{Discrete, token-level and semantic alignment}}\\
\citep{duan2023s0782} & ZuCo EEG / text & DeWave: vector-quantised EEG codebook $\rightarrow$ BART & Reconstruction pretraining; fixation-aligned features or raw waves; both evaluated with TF & BLEU-1 41.35 / 20.51, respectively \\
\citep{tao2024s0786} & ZuCo EEG / text & SEE: cross-modal codebook and semantic matching & Text-encoder similarities provide soft targets; joint semantic-matching and generation losses & BLEU-4 7.7; ROUGE-1-F 31.1 \\
\citep{zhou2024s0787} & ZuCo EEG / text & BELT-2: byte-pair alignment and Q-Conformer $\rightarrow$ LLM & Multi-task representation learning; T5 decoder with learned prefix tuning (source Table~1) & BLEU-1 52.38; BLEU-4 17.95 \\
\addlinespace[3pt]
\multicolumn{5}{@{}l}{\textit{Autoregressive semantic guidance and anchor-based reconstruction}}\\
\citep{liu2025s0801} & ZuCo EEG / text & GLIM: contrastive alignment, language modelling and multiple text variants & Autoregressive; original-reference scoring and generated-text identification \citep{wang2026s0791} & BLEU-1 7.84; 24-way accuracy 10.74\% \\
\citep{wang2026s0791} & ZuCo EEG / text & SemKey: semantic-attribute heads and EEG key--value embeddings & Autoregressive; unique-text split; generated-text semantic identification (v3) & BLEU-1 8.98; 24-way accuracy 16.24\% \\
\citep{zheng2026guiding} & ZuCo EEG / text & Guiding LLMs: EEG--language semantic alignment & Published benchmark values quoted by Brain-CLIPLM \citep{yang2026s0806} & BLEU-1 21.3; ROUGE-1 F 20.3 \\
\citep{yang2026s0806} & ZuCo EEG / sentence reconstruction & Brain-CLIPLM: ordered keyword anchors $\rightarrow$ LLM reasoning and retrieval & Semantic-anchor recovery; reconstruction supported by a task-specific sentence corpus & BLEU-1 22.4; ROUGE-1 F 22.9 \\
\addlinespace[3pt]
\bottomrule
\end{tabular}
\end{table*}

\paragraph{Testing the learned representation.} A useful comparison holds the generator, segmentation information and inference protocol fixed, then tests whether neural inputs improve decoding of held-out text over matched uninformative inputs. Participant or session holdout tests whether that advantage survives recording changes. Together with controls for visual, ocular and other task-related signals, these comparisons provide stronger evidence that the representation carries transferable linguistic information.

\subsubsection{Semantic targets and generated text}

\paragraph{Cued text generation.} Cue-conditioned generation combines a neural representation with content supplied by the experiment. An EEG masked-autoencoder/BERT system reported masked-token accuracy of 48.9\% in training and 27.3\% in validation \citep{sparsh2025s0626}, illustrating the gap between learning a token representation and predicting held-out content. MindSpeech uses HD-fNIRS from word-cloud-cued internal sentence production to prompt-tune Llama-2 for text-prefix-conditioned continuation, reporting BLEU-1 and BERT-precision improvements in three of four participants \citep{suyi2024s0450}. Cue words and the language prior already constrain plausible sentences. Prompt-only and matched shuffled-signal comparisons are therefore needed to measure the additional content supplied by the neural recording.

\paragraph{Semantic reconstruction.} fMRI-based language decoding with a language-model prior reconstructed the gist of continuous heard stories in three participants and transferred to internally narrated stories and silent videos \citep{tang2023s0356}; the target is semantic content rather than verbatim wording. In the imagined-speech experiment, the decoder generated text for five one-minute stories; evaluation compared these sequences with separately recorded retellings using semantic similarity and five-story identification. This demonstrates generative transfer under that protocol, while leaving unrestricted self-generated communication unestablished. Autoregressive generation reconstructed ten minutes of language stimulus \citep{ye2025s0549}, and language-free movies have been used for cross-participant alignment \citep{tang2025s0604}. These extensions test whether a shared semantic representation can support transfer when the source and target stimuli differ.

Semantic reconstruction describes the target relationship between input and output, rather than a medium alongside text and audio. An embedding-retrieval model and a text generator can both recover semantic information but produce different outputs. Prompt-conditioned HD-fNIRS generation similarly requires separating neural evidence from cue and language-model constraints \citep{suyi2024s0450}.

Semantic models also differ in how brain-derived information conditions language generation. PredFT introduces a predictive representation from selected brain regions alongside a primary fMRI language-decoding network \citep{yin2026predft}. Multimodal alignment models instead route among text-, image- and audio-related semantic features \citep{changrui2026s0685}. These alternatives extend semantic conditioning beyond a single fixed embedding. Their evaluation still depends on the task and dataset; use of a shared architecture across modalities is not itself transfer of a fitted decoder between them.

\begin{figure}[!htbp]
\centering
\includegraphics[width=\textwidth]{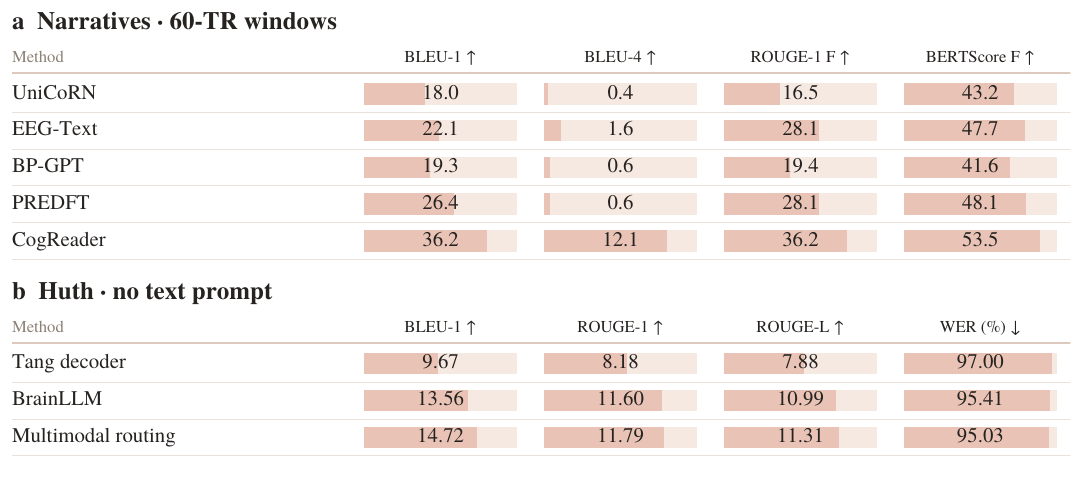}
\caption{\textbf{fMRI language decoding has complementary Narratives and Huth benchmarks.} (a) CogReader's 60-TR comparison \citep{lu2025cogreader} includes UniCoRN \citep{xi2023s0790}, an fMRI-adapted EEG-Text encoder \citep{wang2022s0780}, BP-GPT \citep{chen2025bpgpt}, PREDFT \citep{yin2026predft} and CogReader \citep{lu2025cogreader}; train, validation and test stories differ. (b) Multimodal routing \citep{changrui2026s0685} is compared with Tang's decoder \citep{tang2023s0356} and BrainLLM \citep{ye2025s0549} without a supplied text prompt. The two datasets and decoding protocols remain separate. All metrics use a 0--100 scale; 20/40-TR and text-prompt conditions appear in \cref{tab:fmri_window_values,tab:fmri_prompt_values}.}
\label{fig:method_fmri_benchmark}
\end{figure}

\paragraph{Comparing fMRI language decoders.} The Narratives benchmark connects these representational choices to reconstruction over longer context windows. CogReader compares UniCoRN, an fMRI-adapted EEG-to-Text model, BP-GPT, PredFT and its own incremental-and-wrap-up architecture on story-disjoint data \citep{lu2025cogreader}. Across 20, 40 and 60 repetition-time (TR) windows, CogReader's BLEU-1 rises from 25.4 to 31.2 and 36.2, whereas PredFT rises from 24.3 to 25.9 and 26.4; UniCoRN decreases from 22.9 to 18.0. The 60-TR comparison appears in \cref{fig:method_fmri_benchmark}a. CogReader's BERTScore F1 also increases, from 46.3 to 53.5 on the source's 0--100 scale. These contrasting window-length trends are consistent with differences in context use, although the comparison does not isolate which architectural component causes them. A second comparison on Huth recordings tests Tang's decoder, BrainLLM and multimodal routing without a supplied text prompt: BLEU-1 is 9.67, 13.56 and 14.72, respectively, while WER decreases from 97.00\% to 95.41\% and 95.03\% \citep{changrui2026s0685}. The multimodal route scores higher on these text metrics, although exact word recovery remains much weaker than the fluency of generated sentences might suggest (\cref{fig:method_fmri_benchmark}b). Narratives and Huth form separate benchmark lineages; their absolute scores depend on different recordings and decoding windows.

\subsection{Speech synthesis}

\paragraph{Choosing what the neural decoder predicts.} Speech synthesis separates prediction from rendering: the neural model estimates a representation and a synthesiser turns it into sound. Unit selection can reuse the participant's voice, as demonstrated from motor, premotor and inferior frontal ECoG \citep{herff2019s0796}. Articulatory-kinematic targets instead impose a production-related structure, enabling ECoG sentence synthesis and transfer to mouthed production in a five-participant study \citep{anumanchipalli2019s0127}; its listener tests used a limited candidate word pool. Interpretable speech parameters paired with a differentiable synthesiser were studied in a 48-participant ECoG cohort \citep{chen2024s0381}, although each condition need not contain all 48 participants. Direct reconstruction from sensorimotor ECoG during overt speech provides another production-based route, evaluated with complementary acoustic and perceptual measures \citep{berezutskaya2023s0324}. These alternatives change the supervision needed and the aspects of voice accessible to the decoder.

\paragraph{What successful rendering establishes.} Heard-speech reconstruction has a measured acoustic reference: auditory-cortex ECoG recovered spectrotemporal structure of words and sentences with average correlation $r=0.28$ in 15 participants \citep{pasley2012s0021}. Predicting vocoder parameters from iEEG supported 75\% digit recognition in a five-participant, ten-alternative listening test \citep{akbari2019s0130}. The latter adds perceptual evidence to acoustic similarity, but still leaves open whether listeners can recover sentences without supplied alternatives. Evaluation must follow the information the output is intended to convey.

Interactive control is a further test, even within a small output space. A single-participant sEEG system controlled two synthesised vowels after Overt training, with 85\% first-round success \citep{tankus2025s0495}. This tests whether a user can act through audible feedback; it does not resolve the larger representational problem of composing intelligible sentences. Output diversity and closed-loop controllability should therefore be varied and measured separately.

\paragraph{Inner synthesis and perceptual tests.} The voice route of \cref{fig:routes} is also applied to Inner conditions. Real-time synthesis transferred from overt training to whispered and Inner speech in one sEEG participant, but the authors state the audio is not yet intelligible, and the $r=0.62$ often quoted is overt cross-validation \citep{angrick2021s0221}. NeuroTalk reported inner-speech character error rate (CER) of 68.3\% with a mean opinion score (MOS) of 2.78 on twelve words and CER 83.1\% on the single held-out word \citep{youngeun2023s0372}. An EEG synthesis study \citep{xiong2025s0617} reports 91.2\% listener accuracy for ``intelligible utterances'' from EEG, but the test is a four-alternative choice among known Mandarin disyllables with overt bridging and dynamic-time-warping (DTW) reference alignment, and the train/test independence is not fully documented.

The synthesis route predicts acoustic features or articulatory parameters and renders audio. Overt training offers paired signals, whereas Mouthed and Inner transfer must cope with missing or altered acoustic supervision. Some ten-participant ECoG conditions yielded recognisable online Overt or Mouthed synthesis without intelligible Inner synthesis \citep{meng2023s0312}. Clinical streaming systems extend the objective to timely audible feedback \citep{littlejohn2025s0496,wairagkar2025s0504,wairagkar2026s0653}. Generated audio, perceptual intelligibility and an interactive speaking experience therefore require separate evidence.

\paragraph{Non-invasive overt decoding.} Non-invasive overt results are smaller in scale and must be checked for orofacial EMG, head motion, eye movement and acoustic cross-talk. Early EEG-to-acoustics regression with recurrent networks in four participants predicted acoustic features, evaluated by mel-cepstral distortion and cepstral reconstruction error \citep{gautam2020s0177}. These acoustic errors do not measure whether a listener can recover the words. MEG-to-mel synthesis with a Squeezeformer and BigVGAN vocoder reported a spectral correlation of $r=0.954\pm0.014$ for five repeated phrases in four participants \citep{jinuk2024s0420}; this is a within-participant, closed-phrase correlation, not word accuracy.

\paragraph{Perceived-auditory reconstruction.} Listening supplies a measured acoustic target for non-invasive synthesis. FESDE connects EEG and speech latent modules to generate the listened waveform directly, providing an alternative to a separately predicted spectrogram \citep{jihwan2024s0481}. Together with the production-based systems above, this illustrates how waveform rendering can be learned through different intermediate representations.

\begin{figure}[!htbp]
\centering
\includegraphics[width=\textwidth]{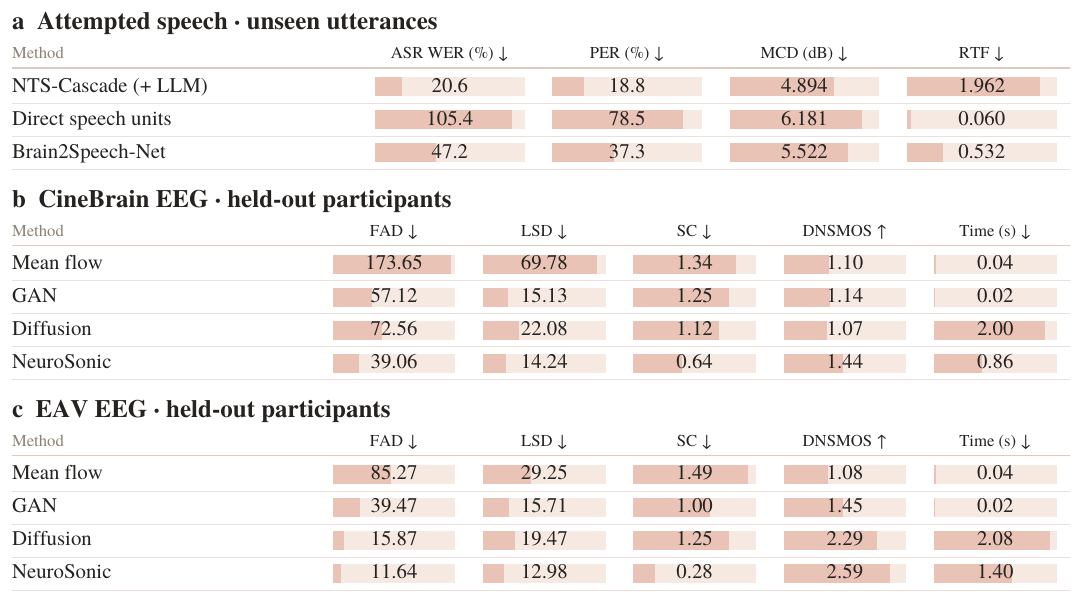}
\caption{\textbf{Synthesis benchmarks compare lexical fidelity, acoustic quality and computation.} (a) Brain2Speech-Net \citep{shreeram2026s0651} reimplements an LLM-rescored text cascade \citep{willett2024lessons} and direct speech units \citep{metzger2023s0299} with a common neural encoder on 880 unseen attempted-speech utterances. ASR WER and phoneme error rate (PER) score generated audio; mel-cepstral distortion (MCD) uses synthetic VITS references. RTF is computation time divided by audio duration. (b,c) NeuroSonic's held-out-participant comparisons \citep{gao2026neurosonic} adapt MeanFlow \citep{geng2025meanflow}, HiFi-GAN \citep{kong2020hifigan} and Mo\^usai diffusion \citep{schneider2023mousai} to EEG. CineBrain includes environmental sound; EAV contains conversation. FAD is Fr\'echet audio distance, LSD log-spectral distance and SC spectral convergence. DNSMOS predicts overall audio quality; it is not a listener intelligibility score. Seconds report the source's inference time, not clinical feedback latency. Values are source means; each dataset forms its own comparison.}
\label{fig:method_synthesis_benchmark}
\end{figure}

\paragraph{Comparing synthesis routes on shared recordings.} Brain2Speech-Net compares text-mediated synthesis, direct speech units and a phoneme-bottleneck generator on a common attempted-speech corpus \citep{shreeram2026s0651}. On 880 unseen utterances, the LLM-rescored text cascade attains ASR WER of 20.6\%, compared with 105.4\% for direct speech units and 47.2\% for Brain2Speech-Net (\cref{fig:method_synthesis_benchmark}). Their reported real-time factors are 1.962, 0.060 and 0.532, respectively. Text mediation preserves lexical content most accurately in this comparison, while direct units render fastest; Brain2Speech-Net occupies an intermediate position. These are implementations with a shared neural content encoder, evaluated against synthetic reference speech, rather than the original systems' clinical outcomes. ASR WER measures machine recognition of the rendered output, and real-time factor measures rendering cost relative to audio duration; listener intelligibility and interactive onset latency remain separate outcomes.
For non-invasive recordings, NeuroSonic compares EEG-conditioned adaptations of mean-flow, GAN and diffusion generators on held-out participants in CineBrain and EAV \citep{gao2026neurosonic}. Its EAV result reduces Fr\'echet audio distance from 15.87 for the diffusion baseline to 11.64 and increases predicted overall audio quality from 2.29 to 2.59; on CineBrain, the corresponding values change from 72.56 to 39.06 and from 1.07 to 1.44. CineBrain includes environmental sound, whereas EAV contains conversation. These comparisons extend acoustic generation beyond clinical speech corpora, but neither audio-distribution similarity nor a predicted quality score measures how accurately a listener can recover linguistic content.

\subsection{Facial and body animation}

Facial animation adds a visual channel to neural communication. An ECoG neuroprosthesis combines decoded text, personalised speech and an avatar \citep{metzger2023s0299}. Its real-time avatar is driven by decoded audio, while direct neural prediction of facial motion was evaluated offline. These two designs place visual rendering at different points in the pipeline: one inherits speech timing from the synthesiser, and the other estimates movement from neural features. Visual evaluation includes sentence identification and lip/jaw trajectories, complementing the text and acoustic outcomes.

Parallel speech and gesture decoding extends expressive control beyond the face. Each of two participants used a 253-channel ECoG array for concurrent phrase-to-text and gesture decoding, within a three-participant movement-mapping study \citep{brosler2026s0760}. In Bravo-6, who attempted speech with vocalisation and imagined the gestures, 200 simultaneous real-time copy trials yielded phrase accuracy of 70\% and gesture accuracy of 66\%, using ten phrases and ten gestures plus rest. Decoded gestures triggered preset full-body avatar animations. Training on both isolated and simultaneous attempts improved performance across contexts, showing that combining output channels also requires learning their concurrent neural activity. The contribution is coordinated verbal and nonverbal control within this repertoire, with phrase selection and gesture classification measured separately.

\subsection{Pretraining, transfer and personal calibration}

\paragraph{Three roles for pretrained models.} Pretraining can supply a neural encoder, an acoustic or linguistic alignment target, or an output generator. These roles address different shortages in paired neural-language data \citep{mai2025reviewr35,wang2025reviewr34}. Contrastive EEG pretraining in BENDR and masked intracranial representation learning in BrainBERT illustrate reusable encoders \citep{kostas2021bendr,wang2023brainbert}. Speech-specific work applies self-supervised learning to overt ECoG, heterogeneous MEG data and long-context word retrieval \citep{yuan2024ecogpretraining,dulhan2024s0480,jayalath2026megxl}. Speech and language models provide the other two roles: a structured target to which neural features can align, and a prior for rendering those features as an output.

\SSPretrainingMiniFigure

\begin{figure}[!htbp]
\centering
\includegraphics[width=\textwidth]{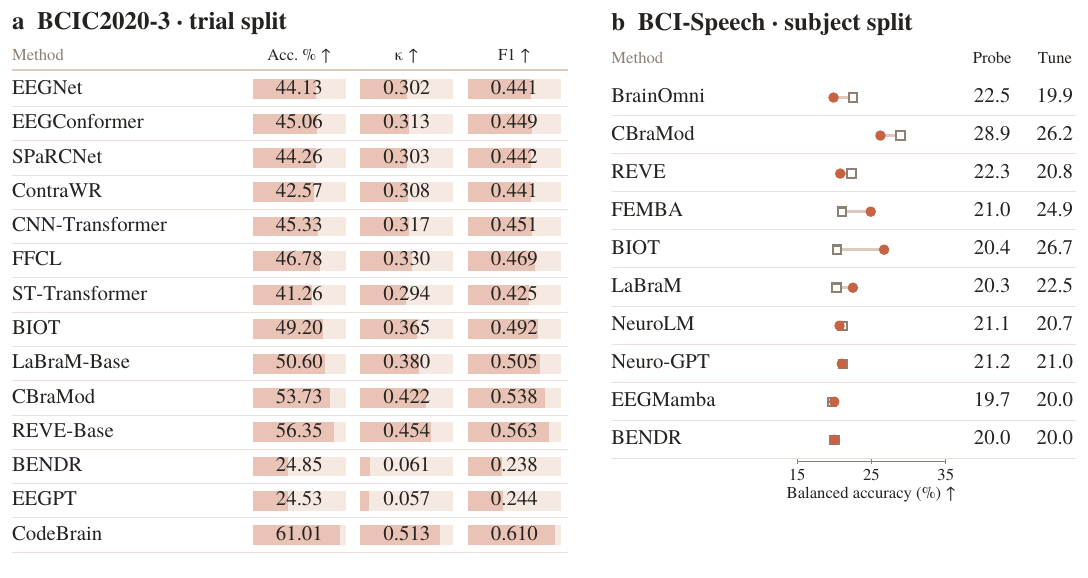}
\caption{\textbf{Pretrained encoders differ in task adaptation and transfer across participants.} (a) Official BCIC2020-3 trial-split results combine REVE's comparison \citep{elouahidi2025reve} with reported BENDR, EEGPT and CodeBrain values \citep{ma2026codebrain}; downstream tuning remains paper-specific. (b) OmniEEG-Bench \citep{lu2026omnieegbench} uses subject-disjoint BCI-Speech data: squares are frozen probes and circles full fine-tuning. The two panels use separate datasets. Method references: EEGNet \citep{lawhern2018eegnet}, EEGConformer \citep{song2023eegconformer}, SPaRCNet \citep{jing2023sparcnet}, ContraWR \citep{yang2023contrawr}, CNN-Transformer \citep{peh2022cnntransformer}, FFCL \citep{li2022ffcl}, ST-Transformer \citep{song2021sttransformer}, BIOT \citep{yang2023biot}, LaBraM \citep{jiang2024labram}, CBraMod \citep{wang2025cbramod}, REVE \citep{elouahidi2025reve}, BENDR \citep{kostas2021bendr}, EEGPT \citep{wang2024eegpt}, CodeBrain \citep{ma2026codebrain}, BrainOmni \citep{xiao2025brainomni}, FEMBA \citep{tegon2025femba}, NeuroLM \citep{jiang2025neurolm}, Neuro-GPT \citep{cui2024neurogpt} and EEGMamba \citep{gui2024eegmamba}. Accuracy is balanced; F1 is weighted; $\kappa$ is Cohen's agreement. Source SDs appear in \cref{tab:pretraining_trial_values,tab:pretraining_transfer_values}.}
\label{fig:method_pretraining_benchmark}
\end{figure}

\paragraph{Shared benchmarks for pretrained neural encoders.} Imagined-speech classification supplies a language-related downstream test for general EEG models. REVE's BCIC2020-3 comparison brings together eleven methods, from task-trained EEGNet and Conformer models to BIOT, LaBraM and CBraMod \citep{elouahidi2025reve}. Five-class balanced accuracy increases from 44.13\% for EEGNet to 49.20\% for BIOT, 50.60\% for LaBraM, 53.73\% for CBraMod and 56.35\% for REVE (\cref{fig:method_pretraining_benchmark}a). Earlier scores are quoted from prior reports under the trial split, and each complete downstream procedure includes its own model and tuning choices. CodeBrain extends this trial-split comparison to 61.01\% balanced accuracy, a Cohen's $\kappa$ of 0.513 and weighted F1 of 0.610; its BENDR and EEGPT baselines remain near 25\% accuracy under the same reported split \citep{ma2026codebrain}. The spread across encoders is therefore larger than the incremental gains among recent leaders. Model architecture, source data and the downstream tuning recipe jointly determine the quality of adaptation.

Transfer to new participants changes this picture. OmniEEG-Bench evaluates ten foundation models on the separate five-class BCI-Speech dataset using a subject-disjoint split and common adaptation procedures \citep{lu2026omnieegbench}. Frozen linear probes and full fine-tuning produce different rankings (\cref{fig:method_pretraining_benchmark}b): CBraMod scores 28.9\% and 26.2\%, BIOT 20.4\% and 26.7\%, and REVE 22.3\% and 20.8\%, respectively, against 20\% chance. Fine-tuning helps some representations but does not consistently improve all models on unseen participants. A separate BCIC2020-T3 comparison also finds lower scores when a participant is excluded from training: its five methods span 25.7--30.2\% within participant and 18.9--20.2\% in leave-one-participant-out evaluation \citep{owais2026s0687}. That study evaluates official validation trials and, in leave-one-participant-out folds, stops training on the held-out participant's accuracy (\cref{tab:leakage_routes}); its absolute values should therefore be kept separate from REVE's test split, and its cross-participant scores are, if anything, optimistic. Trial-split performance and cross-participant transfer answer complementary questions: how well an encoder can fit a target task, and how much of its learned structure survives a change of user.

Transfer can bridge tasks, participants or recording sessions; these are different learning problems. Overt-trained synthesis transferred to Mouthed production in ECoG \citep{anumanchipalli2019s0127}. Overt or Perceived--Auditory training also improved Inner word classification in a ten-participant low-density ECoG study; four participants exceeded chance on Inner only after such transfer \citep{de2026s0773}. Silent-reading reconstruction using Overt reference alignment remains a different Perceived--Visual test \citep{martin2014s0032}. Transfer is useful to the extent that the target task retains the learned representation without requiring an unavailable reference.

Transfer outcomes depend on the source and target task. Articulated EEG did not improve speaker-independent imagined-vowel classification in one study \citep{chris2024s0485}, whereas other multi-condition EEG work investigates how overt, whispered and imagined modes can be combined \citep{alonsovzquez2025s0546}. Such comparisons identify when shared production structure is useful and when the change in task outweighs it. They also motivate reporting learning curves over target-task labels, rather than only the final accuracy of a pretrained model.

Intracortical Inner decoding used attempted-speech pretraining and up to an hour of Inner data. Cued online decoding reached a 50-word vocabulary in three participants and 125{,}000 words in two, with large-vocabulary WERs of 26\% and 54\%; the full study enrolled four participants \citep{kunz2025s0568}. Articulatory and Auditory instructions, prompted sentences and the partial free-recall or counting results remain distinct conditions. The large-vocabulary result thus concerns instructed linguistic production; spontaneous content and unrestricted message choice pose additional tests.

Across participants, shared latent dynamics and position-aware models address differences in the populations sampled by micro-ECoG, surface and depth electrodes \citep{spalding2026s0759,singh2025s0629,chen2025s0630}. Calibrated multi-user intracortical models reuse phoneme structure while fitting a new user's projection \citep{fogg2026s0639}. Across sessions, ALIGN uses semi-supervised adversarial alignment to reduce session-specific variation while retaining phoneme information \citep{zhang2026align}. Source-task transfer, new-user fitting and longitudinal adaptation thus solve different forms of mismatch; the relevant comparison keeps the target-data budget and output task explicit.

Additional indexed directions are documented in \cref{sec:coverage}; the main comparisons here retain their task and source boundaries.

\paragraph{What can be shared, and what remains personal?} These studies share a design question: which learned structure can be reused, and which mapping must be adapted to the target recordings? The reusable component may be neural temporal structure, acoustic or linguistic targets, or decoder dynamics. Adaptation may use labelled target-task examples, a fitted user projection or session-domain alignment, depending on the method \citep{de2026s0773,kunz2025s0568,fogg2026s0639,zhang2026align}. Their value is measured by performance at an explicit target-data budget.

Evaluation connects these design choices to their outcomes: which learned representation or prior improves performance, which intended distinctions are preserved, and under what conditions the gain remains useful.

\section{Evaluation}
\label{sec:eval}

Evaluation has expanded with the outputs that neural decoders can produce. Early classification and retrieval studies measured whether a recording distinguished specified alternatives. Sentence decoders adopted transcription and language-generation metrics; speech synthesis added listener judgements; online neuroprostheses added speed, feedback and sustained operation. Established BCI frameworks had already separated decoder performance, interface performance and the benefit to the user \citep{thompson2013aac,thompson2014performance}. That distinction remains useful as language outputs become less constrained.

Shared benchmarks now make some algorithmic comparisons possible. Brain-to-Text '24 fixed a private attempted-speech test set, allowing different decoding pipelines to be compared by word error rate \citep{willett2024lessons}. The PNPL series introduced standardised listening-MEG tasks and data conditions, progressing from speech detection and phoneme classification to word classification and adaptation with limited target-user data \citep{gilad2025s0618,mantegna2026pnpl}. In parallel, independent evaluations of EEG-to-text have examined how inference settings and language priors affect the interpretation of text scores \citep{jo2025noise,zhang2026s0792}. The development is therefore both quantitative and methodological: better outputs are being accompanied by clearer definitions of what was predicted and what information was available.

\begin{figure*}[!htbp]
\centering
\includegraphics[width=\textwidth]{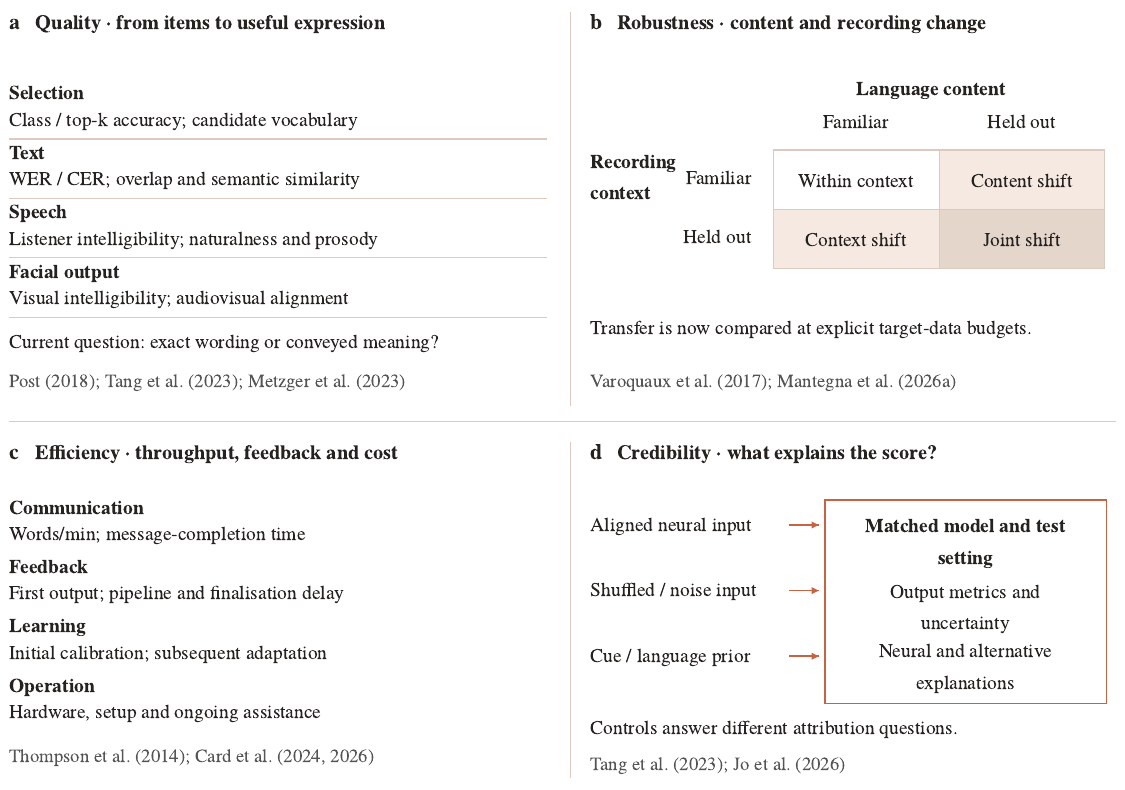}
\caption{\textbf{Evaluation practices have broadened from decoding scores to transfer, communication costs and source attribution.} Each panel connects a measurement question to methods used in the literature and an active point of discussion \citep{thompson2014performance,post2018clarity,willett2024lessons,jo2025noise,mantegna2026pnpl}. The four dimensions organise the review; they do not define a new composite score or a universal evaluation protocol. WER and CER denote word and character error rates.}
\label{fig:evaluation_framework}

\end{figure*}

Four dimensions organise the following synthesis: \emph{Quality}, \emph{Robustness}, \emph{Efficiency} and \emph{Credibility} (\cref{fig:evaluation_framework}). They summarise related measurement traditions rather than a settled universal standard. Published results can be compared where their task, data and scoring conditions match; broader clinical progress can be described across studies while retaining differences in participants and use settings. Method-specific benchmark results accompany their decoding routes in \cref{sec:methods}; the following discussion traces the measurement standards and their interpretation. Reported timing examples retain their original definitions and use settings.

\subsection{Quality}
\paragraph{From item recognition to sequences.} Classification and retrieval remain useful when the scientific question concerns a specific linguistic distinction or an interface offers a defined set of alternatives. Accuracy, classwise scores and confusion matrices describe recoverable contrasts; top-$k$ accuracy or rank describes retrieval among supplied candidates. Candidate size and class frequency matter because a system that favours common words can achieve high aggregate accuracy while rarely recovering less frequent content. Balanced measures and explicit vocabularies make that behaviour visible. Finite-sample significance additionally depends on the number and dependence of observations, not only the nominal chance probability \citep{combrisson2015chance}. The PNPL 2026 design retains explicit word vocabularies and discusses vocabulary coverage alongside performance, reflecting the difficulty of comparing retrieval tasks whose candidate sets differ \citep{mantegna2026pnpl}.

Sequence outputs require accounting for insertions and deletions as well as wrong choices. Phoneme, character and word error rates measure edit operations against a reference; WER is $(S+D+I)/N$, where $S$, $D$ and $I$ count substitutions, deletions and insertions, and $N$ is the number of reference words. Clinical text studies and shared attempted-speech benchmarks use WER to compare decoded messages with prompted transcripts \citep{willett2023s0300,metzger2023s0299,willett2024lessons}. This supports interpretable within-protocol comparison while retaining an important distinction between the neural-unit decoder and the language-assisted final transcript.

\paragraph{From exact wording to conveyed meaning.} BLEU and ROUGE entered neural text generation through machine translation and summarisation, while contextual similarity measures such as BERTScore allow agreement beyond exact lexical overlap \citep{papineni2002bleu,zhang2020bertscore}. These metrics answer different questions: BLEU-1 measures unigram overlap, not word accuracy, and semantic similarity may reward a paraphrase despite changes in a name, relation or negation. Tokenisation and implementation also change the resulting score \citep{post2018clarity}. For semantic reconstruction, information understood by a recipient supplies a complementary endpoint. Independent readers answered questions about a heard story after reading its fMRI-decoded reconstruction \citep{tang2023s0356}. This links generation to comprehension while leaving verbatim transcription as a separate objective.

The current debate concerns both the metric and the evidence that produced it. Jo and colleagues compared EEG-to-text models under the same decoding settings with real and noise inputs, showing that overlap scores can remain similar in the tested configurations \citep{jo2025noise}. COFETT instead emphasises sentence-level embedding alignment across repeated reading/recall conditions and recording sessions \citep{zhang2026s0792}. These studies broaden the available tests, but embedding alignment, repeated-content recognition and autonomous generation of new sentences remain different endpoints. For text intended as exact communication, edit errors and meaning-sensitive assessment therefore provide complementary information; for a semantic target, recipient understanding directly addresses the purpose of the representation.

\paragraph{From an acoustic match to intelligible expression.} Speech synthesis has moved from correlation with recorded acoustic features towards testing what listeners can understand and how they perceive the voice. Objective measures such as STOI were developed for particular distortions of audible speech \citep{taal2011stoi}; subjective assessment separately measures intelligibility and perceived quality \citep{itu1994p85}. Neural speech studies use both constrained identification and less constrained transcription. Ten-alternative digit recognition and four-alternative disyllable identification establish recoverable distinctions within those inventories \citep{akbari2019s0130,xiong2025s0617}; open transcription asks whether a listener can recover an utterance without being offered the answer set. Naturalness, prosody and voice identity describe further gains that a transcription score alone would miss. Avatar neuroprostheses add sentence identification and facial-trajectory measurements for visual expression \citep{metzger2023s0299}. Consequently, a useful assessment pairs content recovery with the expressive property that the system aims to preserve.

\paragraph{What makes a benchmark comparison interpretable.} A shared test set supports a method comparison when the target, scoring implementation and permitted inference information are fixed. Training-data budgets, external language resources and candidate vocabularies remain part of that comparison. The route-specific results in \cref{fig:method_attempted_benchmark,fig:method_meg_retrieval,fig:method_reading_controls,fig:method_meg_generation} illustrate these choices in Methods. Across routes, the reusable standard is the declared evaluation contract: what is held out, what information is supplied and which output property is scored. It does not require pooling different tasks into one ranking.

\subsection{Robustness}
\paragraph{Generalisation follows the available datasets.} Repeated-item recordings made within-person decoding and held-out repetitions early practical evaluation settings. Larger natural-language corpora subsequently enabled tests of new sentences or stories, while pooled recordings enabled session and participant transfer. These splits change different parts of the prediction problem: new content challenges linguistic generalisation, a new day changes the recording, and a new participant changes anatomy, coverage and the learned neural mapping. Cross-validation methodology formalises the corresponding separation between training, model selection and testing \citep{varoquaux2017decoders}. The settings are complementary, rather than successive grades on one robustness scale.

This distinction explains several apparently conflicting results. A decoder can recover another recording of a familiar sentence yet perform poorly on unfamiliar text. Repeated-sentence ECoG results, reading split analyses and repeated-cue EEG studies show why the unit held out matters \citep{makin2020s0165,congchi2025s0600,jiang2026s0672}. In listening-MEG generation, story holdout changes the task from recovering familiar material to composing text for new linguistic content (\cref{sec:methods}). Common data and fixed holdouts make these differences explicit and allow subsequent studies to improve the same prediction setting.

\SSHoldoutMiniFigure

\paragraph{From personal fitting to measured transfer.} Shared representations across distributed intracranial recordings demonstrate that data from other people can improve a target decoder \citep{singh2025s0629}. Multi-user models, shared MEG encoders and session-domain alignment extend this approach through different reusable components \citep{fogg2026s0639,jayalath2026megxl,zhang2026align}. The important comparison is performance at a stated amount of target data: zero-shot prediction asks what transfers without fitting, whereas calibrated prediction asks how efficiently prior learning supports a new recording. High accuracy after extensive personal fitting answers a different question from effective adaptation with a few minutes of data.

Benchmark design is beginning to make this cost explicit. The PNPL 2025 tasks separated standard and extended data conditions; PNPL 2026 introduced deep within-person learning and a broad cross-person track with progressively reduced target-user training budgets \citep{gilad2025s0618,mantegna2026pnpl}. At the review date, the latter describes an ongoing competition, not a completed leaderboard. Its design illustrates a methodological shift from reporting a single transfer score to comparing performance across available calibration data. Continued-use studies should also report calibration effort, update type and performance after recording changes; operating mechanisms are discussed in \cref{sec:practical} \citep{card2024s0386,card2026s0729}.

\subsection{Efficiency}
\label{sec:efficiency}
Efficiency concerns four related costs: producing a message, receiving feedback, learning the user's neural mapping and maintaining the running system. Classical BCI work formalised the distinction between the rate of neural selections and the communication enabled by an interface \citep{thompson2013aac,thompson2014performance}. Selection-based information-transfer rates assume a defined alternative set and error model; they do not transfer unchanged to variable-length language generation. Speech neuroprostheses therefore commonly report words per minute alongside decoding errors, while calibration and processing measurements describe different demands on the user and equipment.

\paragraph{Communication rate and message fidelity.} Representative task, rate and error results are consolidated in \cref{tab:attempted}. Words per minute, correctly decoded throughput and word error rate describe different aspects of communication; rate must be read alongside vocabulary and error burden. Rapid output may still require substantial correction, while a slower stream with fewer errors may support more usable communication.

\paragraph{From initial learning to maintenance.} Calibration reporting has moved beyond total training-corpus size towards the time needed to begin and continue using a decoder. In the 2023 intracortical study, daily training-data collection averaged 41 minutes, while preparation including breaks and model training averaged 140 minutes \citep{willett2023s0300}. A later participant achieved 0.44\% WER on a 50-word copy task after 30 minutes of first-day copy-task data, then 9.8\% WER with a 125,000-word vocabulary after an additional 1.4 hours on day two \citep{card2024s0386}. The increase in error accompanied a much larger vocabulary, illustrating why calibration time must be reported with the capability it enables. The 2026 home-use follow-up combined background updating with optional daily prompted sentences \citep{card2026s0729}. Together with transfer benchmarks that vary target-data budgets, these studies make adaptation cost an outcome in its own right rather than an unreported preliminary step.

\paragraph{Vocabulary coverage and transmitted information.} Recent work extends information-based evaluation from accuracy within a decoder's vocabulary to the fraction of a user's language that vocabulary can express. Open-vocabulary mutual information (OVMI) weights information conveyed among supported words by their coverage of an explicit reference language distribution \citep{jayalath2026ovmi}. It therefore distinguishes a highly accurate small-vocabulary interface from a broader but less accurate decoder, and makes the intended communication domain part of the evaluation. The proposal complements shared benchmarks: a common reference distribution supports a common lexical-information scale, whereas a benchmark controls the neural task and test data. OVMI estimates from scalar accuracy or WER still depend on assumptions about errors and the reference distribution; they do not make attempted-speech and listening experiments clinically equivalent. This direction broadens the discussion from correctly decoded words to the range of messages a system could support.

\paragraph{From completed utterances to streaming feedback.} Streaming synthesis reduces the amount of output that must be accumulated before feedback. Littlejohn and colleagues generated audio in 80 ms increments; Wairagkar and colleagues generated 10 ms frames \citep{littlejohn2025s0496,wairagkar2025s0504}. The latter also measured acquisition through speech-sample synthesis at less than 10 ms using an RTX A6000, with audio-driver playback measured separately. Frame duration specifies how often output advances, while acquisition-to-synthesis time measures the processing that produces it. A GO-cue-to-audio measurement additionally includes the participant's response timing. These distinct endpoints explain how feedback becomes more continuous without equating the output step with the total time from intention to audible speech.

\paragraph{Incremental feedback and final output.} Text systems can display a provisional sequence before completing a more expensive sentence-level pass. The long-term home system estimates incremental feedback at 160--240 ms and reports median finalisation processing of 2,657 ms for its RNN route and 1,791 ms for its Transformer route \citep{card2026s0729}. These medians concern the reported utterances rather than a matched-length test of decoder speed. The measurements locate two opportunities for improvement: quick feedback supports ongoing interaction, while faster finalisation shortens the wait for a revised final sentence. Neither includes every action required to repair an incorrect message, so communication utility remains broader than either component delay.

\paragraph{Computational cost and the remaining comparison gap.} Some recent papers report hardware-specific component times. An offline causal voice model evaluated on earlier T15 recordings reports mean neural-model inference times of 1.47 ms and 2.11 ms for continuous and tokenised output heads on an RTX 5090, with an additional reported 1.2 ms for vocoding \citep{wairagkar2026s0653}. These source measurements describe computational feasibility, not a new online clinical trial. Comparable end-to-end energy, memory and maintenance measurements are still too sparse among the selected reports for a common cost ranking. The emerging pattern is more specific: as text accuracy and raw output rate improve, feedback, finalisation, correction and calibration become increasingly consequential parts of the remaining communication time. For other tasks, especially transfer to new users, collecting enough target data remains a major cost.

\subsection{Credibility}

Credibility concerns the inference supported by an experiment. Prediction conditions determine what information the model could use; controls distinguish possible sources of its performance; statistical analysis describes variation and uncertainty. Reproducible protocols and independent replication connect these questions across studies.

\subsubsection{Information available during testing}
Offline, pseudo-online and live closed-loop evaluation expose a system to different information and interaction conditions \citep{ko2025s0558,card2026s0729}. Published protocols differ in their use of correct prefixes, reference-dependent alignment, known candidates, stimulus boundaries, future samples and repeated observations. These inputs change what the prediction measures. Test-time teacher forcing is one case of privileged information, particularly relevant to autoregressive text; it is not a universal evaluation criterion for classifiers or synthesis systems \citep{jo2025noise}. Known boundaries or reference-dependent alignment can simplify decoding even without teacher forcing (\cref{tab:claimledger}).

\SSAutoregressionMiniFigure

\subsubsection{Neural contribution and controls}

\paragraph{Behaviour.} The first control concerns what the participant actually did. Mouthing retains articulatory activity, and preparation windows remain contiguous with forthcoming speech \citep{willett2023s0300,anumanchipalli2019s0127,dash2020s0150,dash2024s0456}. Reading versus Inner task contrasts and comparisons involving loss of residual articulation show why removing sound does not hold the behavioural source constant \citep{csaky2025s0624,jude2026s0677}. In an around-ear EEG--EMG system, peripheral activity can be a useful communication input while limiting a neural-only interpretation \citep{inoue2026s0725}. The appropriate question is not whether all movement should be excluded from a useful device, but whether the reported contribution matches the claimed source. Channel-specific comparisons separate the benefit of fused inputs, while peripheral recordings address residual movement in internally produced speech.

\paragraph{Priors.} A prior can improve communication while the neural input remains necessary. Attempted-speech systems report phoneme error separately from the \wer obtained after language-model decoding \citep{willett2023s0300,metzger2023s0299}. In fMRI semantic reconstruction, replacing brain-based likelihoods with random scores caused decoding to fall to chance, whereas resetting linguistic context produced a transient loss followed by recovery \citep{tang2023s0356}. That ablation demonstrates ongoing neural constraint in a system that also uses a strong language prior. Prompt-conditioned generation adds cue information that needs its own comparison \citep{suyi2024s0450}. Conversely, the ZuCo audit compared BART, Pegasus and T5 using real and noise inputs, with and without test-time teacher forcing \citep{jo2025noise}. Comparable EEG/noise results limit the neural-decoding interpretation of those Perceived--Visual pipelines; they do not establish a universal impossibility result for Inner, Auditory or later untested generators. The useful question is which information the tested decoder actually requires.

A control is informative only when its preserved and removed information match the alternative explanation. Prior-only or cue-only decoding asks how much can be predicted without a neural measurement. Shuffling the association between recordings and targets tests trial-specific information only if nuisance structure and the permitted timing information are preserved. Replacing inputs with noise tests dependence on the measured values, but an arbitrary replacement can also create an unfamiliar input distribution; failure on that control alone is weak evidence of neural specificity. Conversely, success with signal-blind input exposes an available shortcut, as in the MEG retrieval audit \citep{zhang2026s0793}. Peripheral-channel comparisons then ask a different question: which measured source supplies the usable information. No single control resolves all of these explanations, and artifact cleaning cannot replace this analysis.

\subsubsection{Statistical support and replication}

Theoretical chance is a null expectation, not a finite-sample threshold for reliable decoding \citep{combrisson2015chance}. Permutation tests and uncertainty estimates address sampling variation, with the inferential unit determined by repeated items, sessions and participants. At the group level, the fraction of individually significant participants differs from an estimate of population prevalence; the latter requires its own inferential framework \citep{allefeld2016prevalence}. This distinction matters when many trials are available from only a few people.

\paragraph{Variation across participants.} The Inner decoding results in \cref{sec:imagined} show why participant-level distributions matter more than a best score. Repeated clinical reports also require participant links, because additional follow-up does not add independent recruitment (\cref{tab:attempted}).

\paragraph{Replication in Inner EEG decoding.} A replication audit \citep{tates2026s0682} re-implemented published inner-speech pipelines on multiple public and in-house EEG datasets and found a mean fraction of 36.1\% statistically significant participants across the tested speech-imagery datasets, below the corresponding motor-imagery fraction, with replicated accuracies below the original reports. This is an average of dataset-level fractions, rather than a pooled participant proportion. This does not bound every Inner method. It does show why the distribution of reliable decoding across the tested participants is a different outcome from the best observed accuracy. Paired cross-mode and paradigm-design datasets can help distinguish failure to transfer a model from changes in task execution or engagement \citep{he2025s0633,ma2025s0486,aguilerarodrguez2025s0502}.

These studies expose two levels of reproducibility. Reusing ZuCo or a clinical benchmark tests whether an algorithmic result can be obtained on the same recordings; recruiting additional participants tests whether it extends to another population sample. Repeated reports from T15 or BRAVO3 add longitudinal or functional evidence without adding independent recruitment (\cref{tab:attempted}). Participant-level analysis and correction for multiple comparisons also affect conclusions in auditory EEG benchmarking \citep{li2026s0726}. The field therefore has established tools for assessing individual decoding and increasingly explicit protocols for algorithm comparison, while population-level reliability remains less well characterised.

\paragraph{From evidence to use.} Output metrics, shared benchmarks, source controls and longitudinal tests answer complementary questions. Their combined development makes some method comparisons increasingly reproducible while exposing remaining differences in task and use conditions. \Cref{tab:checklist} collects the details needed to interpret a result. The next section follows the corresponding system mechanisms: initiation, feedback, correction, adaptation and continued access.

\section{Practical Use}
\label{sec:practical}

\begin{figure*}[!htbp]
\centering
\includegraphics[width=\textwidth]{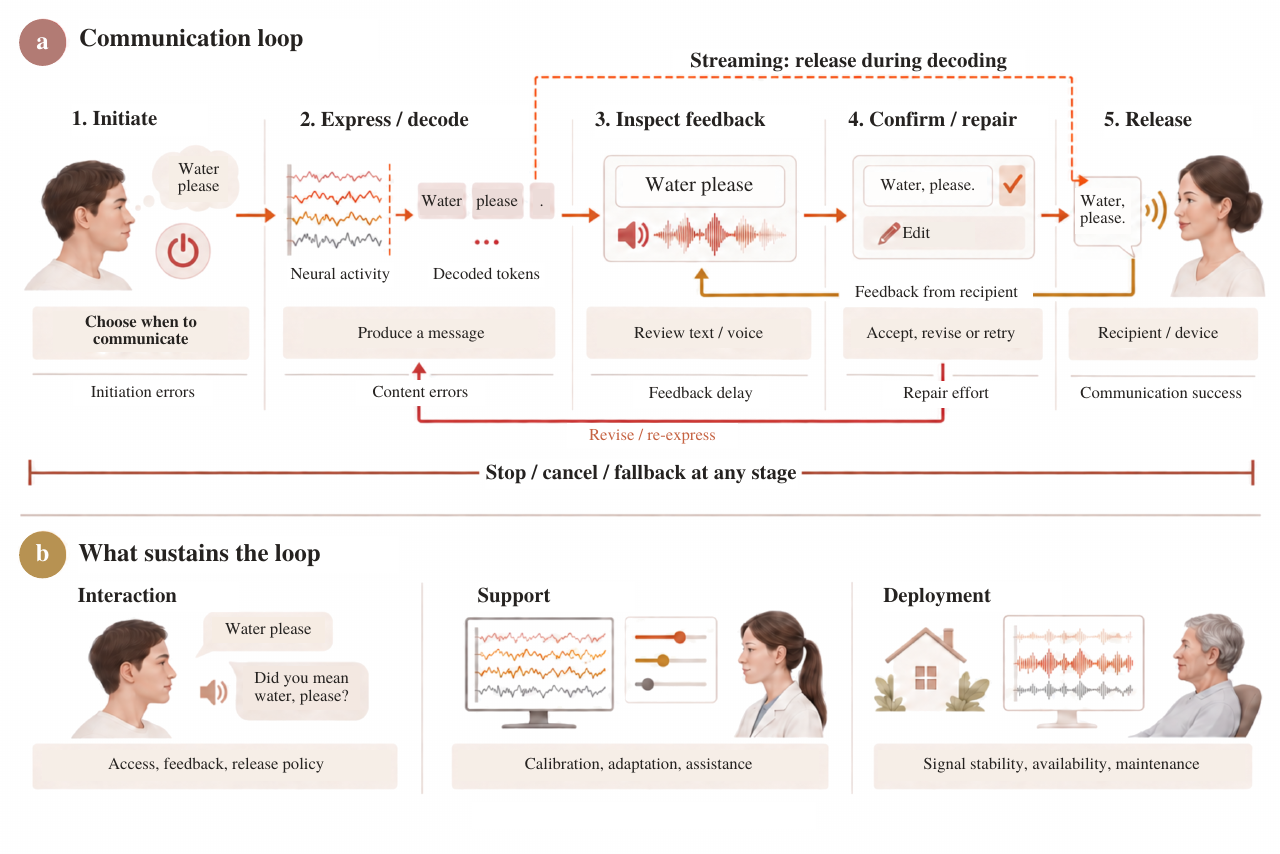}
\caption{\textbf{Communication combines message production, feedback, release and repair under user control.} (a) The solid forward path illustrates confirmation before release; the dashed branch illustrates provisional streaming to the recipient during decoding. Feedback and revision provide return paths, while initiation errors, content errors, delay and repair effort locate different sources of communication cost. (b) Interaction design, calibration and assistance, and sustained recording access support the loop. The people and messages are illustrative. Interfaces may combine these functions differently, including their release policy and available stopping or fallback controls.}
\label{fig:communication_loop}

\end{figure*}

A practical language interface lets a person initiate, express and complete a communication goal. Decoding contributes the content; feedback, correction and adaptation help turn that content into an exchange. User-centred BCI studies consequently assess effectiveness, workload and satisfaction together, while prospective-user research establishes communication priorities before engineering choices are fixed \citep{kubler2014usercentered,huggins2011preferences,branco2021preferences}. These priorities connect three parts of the interface: \emph{Interaction} governs expression and user control, \emph{Support} supplies context and decoder updates, and \emph{Deployment} sustains recording access and assistance. \Cref{fig:communication_loop} shows how they contribute to the same communication loop.

\subsection{Interaction}

Interaction determines when decoded content becomes a message and how the user responds to an error. The distinction between executory control over releasing speech and guidance control over its expression helps separate initiation or cancellation from linguistic correction \citep{vanstuijvenberg2024reviewr64}. Different interfaces distribute these decisions between automatic detection, explicit user actions and continuous feedback. A confirmation step can prevent an unwanted utterance, but it also consumes time and can interrupt a conversational turn.

\subsubsection{Speech onset, stopping and unintended activation}

Speech-state classification, onset detection and idle-state modelling allow a decoder to operate between utterances as well as during them \citep{de2025s0611,ko2025s0558,padfield2025s0494}. In Card's clinical text system, conversation mode detects speech automatically and finalises an utterance after six seconds without detected speech or an eye-tracker button action \citep{card2024s0386}. Automatic segmentation reduces the need to issue a separate command for every turn; the button provides a direct way to finish early. A later Inner-speech system demonstrates real-time keyword gating in one participant, adding a deliberate internally produced action before message output \citep{kunz2025s0568}.

These mechanisms address different use conditions. Automatic detection supports natural turn-taking, whereas a deliberate gate can help separate communication from other verbal activity. Long-term home use also exposes states uncommon in cued testing: occasional false detections produced nonsensical output when the participant was nominally asleep \citep{card2026s0729}. The available activation action, sensitivity to idle behaviour and means of stopping output therefore jointly shape dependable access, especially when fatigue or another control channel changes.

\subsubsection{Feedback, confirmation and error correction}

Feedback has developed from audible formant control to multimodal and streaming speech interfaces \citep{guenther2009s0004,brumberg2018s0081,littlejohn2025s0496,wairagkar2025s0504}. Sentence-level text can be reviewed before release; streaming audio guides ongoing adjustment, but an error already spoken may require the user to clarify the message to the recipient.

Auxiliary controls make repair part of routine use: the home-use system supports text correction through gaze or a neural cursor \citep{card2026s0729}. Feedback can also change the user's control strategy: in a five-day binary syllable-imagery study, continuous feedback supported learning, with substantial individual differences and a discontinuous-feedback comparison \citep{bhadra2025s0575}. This constrained task provides evidence for user learning alongside decoder learning, without equating that gain with conversational proficiency.

\subsubsection{Communication benefit and user experience}

Useful communication includes choice of language, voice and expression. Bilingual decoding extends language choice, while personalised speech, facial animation and expressive voice restore additional aspects of identity and delivery \citep{silva2024s0375,metzger2023s0299,wairagkar2025s0504}. Loudness or mode control and parallel gestures add further dimensions \citep{srinivasan2026s0698,brosler2026s0760}. Their benefit depends on the person's goals and the effort required to control them; attention, fatigue and engagement consequently matter alongside output quality \citep{aguilerarodrguez2025s0502}.

Long-term conversational reports capture both decoding and the user's contribution to a successful message. In the home-use system, sentence ratings include immediately correct, user-corrected and mostly correct output, whereas cued-copy testing provides a known reference transcript \citep{card2026s0729}. The former describes achieved communication with the interface; the latter more directly isolates word recovery. Effectiveness, workload and satisfaction complement these outcomes \citep{kubler2014usercentered}. The Communicative Participation Item Bank supplies a patient-reported construct for speaking situations, whose applicability to a particular neural communication aid still requires appropriate validation \citep{baylor2013cpib}. Together, these measures connect the quality and timing endpoints in \cref{sec:eval} to the experience of completing an exchange.

\subsection{Support}

Support mechanisms work at different points in learning and use. Context helps interpret the current neural evidence; personal calibration fits a new recording; continued adaptation maintains the mapping as signals and user behaviour change. Their practical effects differ accordingly, from fewer corrections within a sentence to fewer interruptions between sessions.

\subsubsection{Inference-time assistance}

Language models help convert uncertain phoneme evidence into words, while a preceding question narrows the likely answer \citep{willett2023s0300,moses2019s0123}. Semantic beam search uses contextual representations to guide sentence generation, and the home-use text system rescores candidate sequences at finalisation \citep{tang2023s0356,card2026s0729}. These mechanisms place linguistic constraints around the neural prediction. They extend an established principle in BCI spelling, where probabilistic language context improves the conversion of noisy selections into text \citep{speier2015particle}.

The resulting fluency is useful when contextual expectations agree with the intended message. Names, negation and unusual expressions create a different demand: the interface needs to retain a low-probability choice that matters to the user. Direct editing and confirmation offer ways to resolve that conflict. The relevant practical gain is the reduction in total correction effort while preserving intended distinctions; the matched model and control comparisons in \cref{sec:eval} address how much of the improvement comes from neural evidence and inference support.

\subsubsection{Long-term stability and recalibration}

The calibration times and paired text performance are reported in \cref{sec:efficiency}. Continued use adds background recalibration in the home system, while cued Inner decoding combines attempted-speech pretraining with participant-specific updates \citep{card2026s0729,kunz2025s0568}. Update labels differ: prompted sentences provide known targets; T16's Inner protocol incorporates each cued sentence after decoding \citep{kunz2025s0568}; Card's system uses sentences confirmed correct via eye tracking for decoder updates \citep{card2024s0386}. Confirmation can reduce separate calibration but requires reliable validation, and each update carries costs in time, labels and assistance \citep{swanson2026reviewr46}.

Related control BCIs illustrate two further approaches. Aligning low-dimensional neural activity recovered cursor control after recording changes in nonhuman primates; sustained user--decoder adaptation enabled stable human ECoG control \citep{degenhart2020stabilization,silversmith2021plugplay}. These are algorithmic stabilisation and coadaptation precedents, respectively. Their relevance to language is the prospect of maintaining a useful representation with fewer interruptions, while preserving the linguistic distinctions and temporal structure that communication requires.

\subsection{Deployment}

Deployment joins the decoder to the equipment and people that keep it available. Recording access, preparation, maintenance and recovery determine when a successful laboratory capability becomes a dependable communication option.

\subsubsection{Setup, portability and maintenance}

The home-use report makes this arrangement concrete: one participant's wired, multi-computer system required about 20 minutes of daily caregiver setup, then supported speech, cursor control and correction without a researcher present \citep{card2026s0729}. Setup assistance and self-directed message production describe complementary parts of the same achievement.

Longitudinal selection/control systems do not decode language content, but connect access to health, support and recording conditions \citep{vansteensel2024longevity}. Wolpaw's home-use study tracked system availability while providing user and caregiver training and technical support \citep{wolpaw2018home}. Caregiver contributions include preparation, maintenance and troubleshooting \citep{wohns2024reviewr47}. Smaller sensors and automatic calibration reduce particular tasks; their contribution to autonomy depends on which tasks the user or caregiver still performs.

\subsubsection{Privacy and user control}

Privacy controls operate at several stages of an interface. Cooperation affected successful semantic reconstruction in the fMRI study; keyword gating in the Inner study supplied a deliberate activation action; a privacy mode in the home-use system disabled data logging \citep{tang2023s0356,kunz2025s0568,card2026s0729}. Activation, message release and data retention are therefore separate design decisions with different technical mechanisms. Disabling storage, for example, changes the retained record rather than the decoder's output policy.

Access to the input also matters. Stimulus-based EEG probing demonstrated information leakage under a defined adversary and observation protocol \citep{martinovic2012sidechannel}, while developer interviews identify responsibilities around autonomy and maintenance \citep{vanstuijvenberg2024developers}. These findings connect control of recordings to control of inferred content and released messages. A usable language interface combines such controls with feedback and repair, so that the user can express a message and manage its consequences throughout an exchange.

\section{Limitations}
\label{sec:limits}\label{sec:open}

The remaining challenges concern self-generated expression, feasible signal access, transfer and sustained communication (\cref{tab:discriminating_tests}).

\begin{table*}[!htbp]
\caption{Remaining limitations and the approaches that provide a basis for addressing them. Each direction concerns a capability needed across learning, decoding and use.}
\label{tab:discriminating_tests}
\centering\small
\begin{tabular}{@{}p{2.4cm}p{4.1cm}p{4.4cm}p{4.4cm}@{}}
\toprule
Challenge & Source of difficulty & Existing progress & Remaining capability \\
\midrule
Self-generated expression & Intended content has no externally supplied transcript or reliable acoustic reference. & Cued Inner decoding and listening-to-imagery transfer \citep{kunz2025s0568,tang2023s0356}. & Flexible expression of user-chosen content with a dependable account of message fidelity. \\
Signal access and supervision & Coverage, task and labels differ between large corpora and intended users. & Pooled perception data and shared intracranial representations \citep{alexandre2023s0322,dascoli2025s0623,singh2025s0629}. & Source representations that reduce target-task learning costs under feasible recording conditions. \\
Transfer and stability & New content, people and sessions change different parts of the learned mapping. & Calibrated multi-user models, session alignment and continued adaptation \citep{fogg2026s0639,zhang2026align,card2026s0729}. & Predictable performance across users with manageable personal fitting and recovery. \\
Sustained communication & Expression, correction and support compete for the user's time and effort. & Streaming voice, home operation and user-centred assessment \citep{littlejohn2025s0496,card2026s0729,kubler2014usercentered}. & Improved participation and complete exchanges with affordable maintenance and assistance. \\
\bottomrule
\end{tabular}
\end{table*}

\subsection{Task coverage and self-generated expression}

Perceived-language tasks offer reusable recordings and precise targets, while clinical attempted-speech systems demonstrate voluntary communication under a different acquisition and supervision setting \citep{tang2023s0356,card2026s0729}. Inner research connects these settings through constrained classification, feedback-driven learning and overt-bridged synthesis \citep{bhadra2025s0575,youngeun2023s0372}. Intracortical work further combines attempted-speech pretraining with instructed Inner decoding \citep{kunz2025s0568}. The open question is which parts of these learned representations remain useful when a person chooses new content without an external stimulus or audible reference.

The neural state matters as much as the word label. Motor attempt, internal articulation and auditory imagery can share content-related structure while differing in intention and execution. Results from users without residual articulation, together with paired-task and peripheral-monitoring studies, make these differences accessible to investigation \citep{jude2026s0677,nieto2022s0292,ma2025s0486,inoue2026s0725}. Their value is to identify transferable linguistic structure while separating it from cue processing and correlated movement. Linking controlled tasks to user-chosen messages also changes the reference: a known prompt provides exact labels, whereas confirmation and independent accounts of intended meaning become central to judging self-generated expression. Better learning from these complementary forms of supervision is a route towards less constrained communication.

\subsection{Signal access and supervision}

Large pooled perception corpora and long within-person recordings have made representation learning possible at a scale that is difficult to reproduce for clinical attempted or Inner speech \citep{alexandre2023s0322,dascoli2025s0623,miran2025s0576}. General EEG and intracranial pretraining provide reusable encoders, while speech-specific latent alignment and multi-user models bring source information closer to a target decoder \citep{kostas2021bendr,wang2023brainbert,singh2025s0629,fogg2026s0639}. The limitation is the mismatch between available supervision and intended use. More source data help when they preserve distinctions the target task needs; differences in task, sensor coverage or labels can reduce that benefit. Target-data learning curves consequently offer a more informative account of reuse than source hours alone.

Recording advances can change this balance by improving access to relevant populations or making acquisition easier. Around-ear EEG--EMG resources, distributed intracranial recordings and emerging magnetic or optical configurations broaden the available design choices \citep{inoue2026s0725,spalding2026s0759,xu2026s0638,tripathy2025s0586}. Their demonstrated targets currently differ: OPM-MEG includes overt-vowel classification, HD-DOT includes language mapping and audiovisual-segment retrieval, and human fUSI language work primarily provides task maps (\cref{sec:signals}). These achievements establish different starting points. Extending a new configuration towards communication depends on obtaining sufficient content-specific information under the target behaviour, together with a practicable supply of training labels.

\subsection{Transfer across content, people and time}

Shared representations and session alignment have begun to reduce the dependence on a separately trained decoder for every recording \citep{singh2025s0629,zhang2026align}. Their benefits address different changes: new content challenges linguistic composition, a new participant changes anatomy and sampled populations, and a later session changes the recording within a person. Calibrated multi-user models and continued adaptation provide evidence that reuse and personal fitting can work together \citep{fogg2026s0639,card2026s0729}. The unresolved issue is how reliably the gains persist across users and how much target supervision each change requires.

Common datasets support repeated method comparisons, while longitudinal and independent-cohort studies extend the range of conditions represented. Split analyses and controlled replications have shown that familiar stimuli and output priors can account for part of apparent transfer in the tested configurations \citep{congchi2025s0600,jo2025noise,tates2026s0682,li2026s0726}. These findings identify specific obstacles to learning robust neural-to-language mappings. Assessing performance across content novelty, recording shift and calibration budgets can reveal whether a model reuses linguistic structure or remains tied to its original setting. Variation between participants, including unsuccessful users, remains consequential for estimating how broadly a reported capability applies \citep{allefeld2016prevalence}.

\subsection{Sustained communication and operating burden}

Streaming synthesis, expressive control, facial animation and prolonged home use have moved neural language interfaces closer to everyday exchange \citep{littlejohn2025s0496,wairagkar2025s0504,metzger2023s0299,card2026s0729}. These developments improve different parts of communication: immediate feedback supports ongoing production, additional output dimensions carry expression, and background adaptation maintains access. Each also changes the work required of the user or support network. A richer output may need additional control, and continued learning depends on labels and recovery when the mapping changes.

User-centred and longitudinal BCI studies already connect effectiveness to workload, availability and assistance \citep{kubler2014usercentered,wolpaw2018home,vansteensel2024longevity}. For direct speech interfaces, the remaining challenge is to sustain gains over complete exchanges and changing needs, including correction and unavailable time. Comparisons with a person's existing communication aid can identify whether an advance reduces effort, improves participation or transfers work to setup and maintenance. This connects the technical directions above to clinical translation: useful representations and efficient models matter through the communication they keep available \citep{stavisky2025reviewr07,dohle2025reviewr45,swanson2026reviewr46}.

\subsection{Limits of this survey}
\label{sec:surveylimits}

The procedures described in \cref{sec:method} also bound the conclusions of this survey. Screening was rule-assisted, and three AI research agents supported the September thematic supplement; neither step provides dual independent human screening. Reading depth varies across sources. The 21 September registry records \nfulltext{} full-text checks among \nkey{} structured extractions; the other extractions are tagged as abstract, targeted-field or inherited checks, and \ncandidate{} entries retain bibliographic and available label information. The October paradigm revision adds separately documented checks of original methods, results and figures, so the registry totals describe the September snapshot rather than all subsequent reading. Abstract-derived descriptions remain author-reported claims. Review comparisons assess coverage rather than provide a full methodological appraisal, and the quantitative comparisons compile reported results without independent replication.

Coverage is also selective. Scopus, Web of Science, IEEE Xplore and ACM DL were not searched natively, citation tracking started from seed reviews and key papers, and English-language search terms can miss work without English titles or abstracts. Much of the recent evidence remains provisional: about one in six cited references is a preprint, two-thirds of these preprints date from 2025--2026, and findings taken from them may change with later versions or peer review. Within these bounds, the survey offers a critical synthesis of the cited evidence rather than an exhaustive account of the field.

\section{Beyond}
\label{sec:beyond}

\begin{figure*}[!htbp]
\centering
\includegraphics[width=\textwidth]{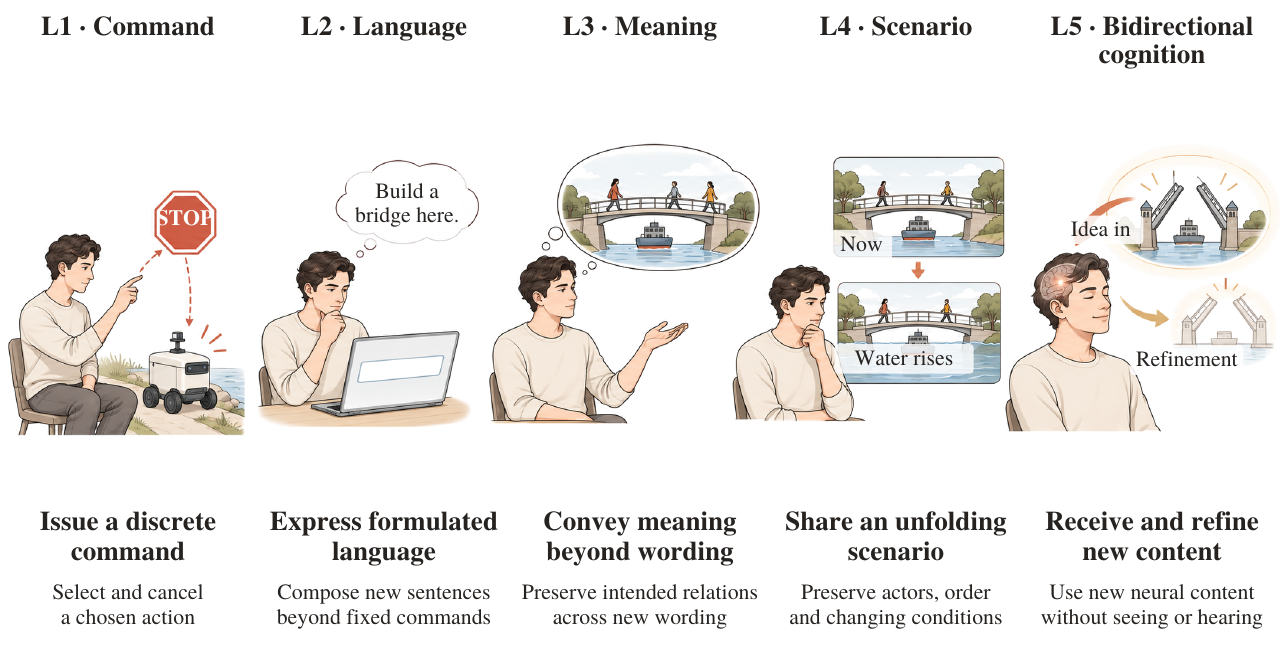}
\caption{\textbf{The proposed progression expands what a person can convey, then adds direct neural receipt of meaningful content.} The river-crossing story distinguishes commands (L1), formulated language (L2), intention independent of wording (L3), actors and relations in an unfolding scenario (L4), and direct neural receipt and refinement of a concept (L5). The upper labels identify the added function; the smaller lower labels state tests for that transition. L4 conveys the user's scenario; L5 requires incoming content to inform understanding without audiovisual presentation. These are proposed research targets, not maturity grades or a ranking of existing systems. L3 remains a frontier and L4--L5 are prospective. The vignettes illustrate hypothetical functions rather than measured results.}
\label{fig:beyondlevels}
\end{figure*}

Future interfaces could expand both what a person communicates and how information returns to them. We propose five levels organised by the \emph{depth of cognitive interfacing} (\cref{fig:beyondlevels}). L1--L4 progress from a selected action to formulated language, intended meaning and an unfolding scenario. L5, a speculative long-term outlook, would add a neural return channel through which meaningful content could enter the person's cognitive process. These are prospective functional targets, not an established standard or a ranking of current systems. A river-crossing story illustrates the added capability at each transition; throughout, the user retains authority to initiate, reject and revise the exchange.

Published evidence meets these transition tests unevenly; locating it marks the current frontier rather than grading systems. Discrete selection (L1) is best established: an EEG-based speller was used independently at home, mainly for communication, by people with ALS for up to 18 months \citep{wolpaw2018home}, and clicks decoded from implanted subdural electrodes supported one person's independent home communication for seven years \citep{vansteensel2024longevity}; in both, the neural signal encodes a selection, and any message is assembled through successive selections rather than decoded as language. Formulated language (L2) has been demonstrated for attempted speech in people with paralysis: large-vocabulary decoding of prompted sentences provided the foundations \citep{willett2023s0300,metzger2023s0299}, and self-directed communication during long-term independent home use by one participant meets the L2 test \citep{card2026s0729}. Cued online decoding of inner speech is an early step towards L2 through inner rather than attempted speech; uninstructed inner speech was decoded only in part, and complete, intelligible sentences could not be recovered during free-form thinking \citep{kunz2025s0568}. Semantic reconstruction of heard stories, imagined narratives and silent videos provides relevant foundations for L3 \citep{tang2023s0356}; voluntary expression of an intended meaning independent of wording remains a prospective target. L4 and L5 formulate prospective tests, drawing on demonstrations of event representations and stimulation-mediated signalling.

\paragraph{L1: Command interface---issue a discrete instruction.}
A person intends one action from a finite set, and the interface selects the corresponding command. In the story, the user mentally selects ``move'' or ``stop'' to position a delivery robot beside a river. Establishing L1 requires evidence of intentional selection, including the ability to withhold or cancel an action; task-correlated activity alone is insufficient.

\paragraph{L2: Language interface---express formulated words.}
The person composes a sentence internally---``Build a bridge here''---and the interface recovers it as text or speech without requiring an audible utterance. The transition from L1 is composition beyond a fixed command inventory. Large-vocabulary speech neuroprostheses provide relevant foundations \citep{willett2023s0300,metzger2023s0299}. To establish the transition, the user should convey new combinations of words beyond learned command associations. The relevant advance is composing a new intended message, beyond selecting or repeating a familiar command.

\paragraph{L3: Meaning interface---convey an intention before choosing wording.}
The person intends that villagers should cross the river while boats remain able to pass, without needing to formulate a particular sentence. The interface helps express that meaning. The transition from L2 is abstraction from wording: different expressions can preserve the same intended distinctions. Semantic reconstruction and the perspective of moving from speech to language BCIs motivate this frontier \citep{tang2023s0356,li2026reviewr54}; neither establishes voluntary meaning-level communication on its own.

The defining capability is preservation of intended relations across wording: allowing a boat to pass and blocking its passage share the same topic but express opposite requirements. An interface at this level would preserve that distinction and the user's revisions without requiring a fixed verbal formulation. Paraphrase accuracy alone cannot establish this transition, because decoding covertly formulated words would also yield accurate paraphrases. Descriptions decoded from fMRI responses to silent videos show that a decoder trained on heard stories can transfer to largely nonverbal input \citep{tang2023s0356}; participants were not instructed to narrate, but covert verbalisation was not manipulated.

One test would show a wordless river scene, let the person privately decide whether the boat should pass or be stopped, and decode the chosen goal after the scene is removed, with the decision confirmed afterwards by a picture choice whose response mapping is revealed only then. Because both goals are chosen after the same scene, outputs would be scored on that minimal pair: any wording that expresses the chosen relation counts as correct, whereas a fluent sentence stating the opposite counts as an error. The test trials request no sentence and pair the decision with either articulatory suppression or a nonverbal secondary task of matched difficulty \citep{nedergaard2023verbal}. Separate sentence-composition trials would validate a wording-level decoder and check that the interference disrupts verbal formulation under the same recording conditions. Evidence would favour wording-independent recovery if meaning-level accuracy remained comparable under the matched tasks despite verified disruption of verbal formulation, and transferred to untrained goals and relations. Transfer beyond the trained choices distinguishes this test from a learned pass-or-stop classifier. Because an unchanged result under verbal interference cannot exclude covert language \citep{nedergaard2023verbal}, such evidence would make wording-independent decoding more plausible without proving it.

\paragraph{L4: Scenario interface---share an unfolding situation.}
The person imagines villagers walking across a bridge, a cargo boat approaching below, and rising water reducing the space available for passage. The interface externalises the scenario's actors, goals, relationships and changes. The transition from L3 is contextualisation: an intention is situated within a structured, evolving situation that supports predictions about what may happen. A scenario can be conveyed through language, diagrams or other representations. Multimodal brain-encoding/decoding surveys broaden the representations available for investigation \citep{oota2024reviewr67}, without establishing voluntary scenario communication.

Shared event-specific neural patterns during perception and verbal recall provide a basis for investigating scenario structure \citep{chen2017sharedmemories}. They suggest that an account of an event can retain organisation across experience and recollection. Turning such representations into voluntary transmission of a new scenario remains a prospective interface objective.

The defining capability is relational and temporal structure: who acts, in what order and under which changing conditions. Changing the arrival time of the boat would change the consequences conveyed by the interface. This can be expressed in language or a diagram; a realistic image or three-dimensional object alone need not contain that scenario structure. The additional evidence concerns new combinations of actors, relations and temporal changes beyond rehearsed narratives; an output description alone does not identify the cognitive representation that produced it.

\paragraph{L5: Bidirectional cognitive interface---receive and refine new content.}
L5 is a speculative long-term outlook that adds delivery of meaning into the brain to the outgoing communication of L1--L4. Hypothetically, through a neural return channel, the AI proposes an opening bridge; without seeing or hearing it, the person considers the idea and returns a refinement such as waiting areas on both banks. The transition from L4 would be reciprocity: content would enter the person's cognition as well as leave it, and the user could assess, reject and revise what arrives.

Demonstrated direct input to the brain is narrower. Invasive brain stimulation does not yet consistently elicit speech-like percepts \citep{hong2024reviewr55}. Brain-to-brain studies have delivered EEG-decoded bits as the presence or absence of phosphenes induced by transcranial magnetic stimulation \citep{grau2014b2b,jiang2019brainnet}, and receivers have combined two senders' bits into a game decision \citep{jiang2019brainnet}. Information was transmitted through a prearranged binary code.

Supporting the hypothesis of concept-level receipt would require a recipient to use neurally delivered content, absent from audiovisual channels, in a new judgement and return a meaningful revision that sham stimulation does not elicit, with transfer to untrained content beyond a memorised stimulation--response code.

The direction is increasing depth of exchange under the user's control: selecting an action, composing language, conveying meaning, sharing a scenario and receiving meaningful neural input. Utility remains separate from level. A reliable command interface may serve a person better than an unreliable scenario interface; the purpose of the progression is to define what additional capacity and evidence each new form of exchange requires.

\section{Conclusion}
\label{sec:conclusion}

Across Articulated, Inner and Perceived tasks, recoverable content depends on the participant's operation, the neural populations sampled and the measurement timescale. Phonetic and lexical, acoustic and articulatory, and contextual-semantic routes support distinct outputs. Evaluation should match the target, separate neural evidence from language priors and pretraining, and connect prediction quality to feedback and sustained use; shared benchmarks make particular learning and inference choices comparable.

The next stage is to make these capabilities more transferable and useful for self-chosen expression. Learning from broader neural and language experience can reduce calibration, but interfaces must preserve the distinctions a user intends and make errors easy to recognise and repair. The five-level outlook extends this direction from commands and formulated language towards meaning, scenarios and a prospective neural return channel. Richer exchange will require evidence of the added capability at each transition, together with the user's ability to initiate, understand and control the communication.

\section*{Impact Statement}
Speech neuroprostheses can restore communication functions for people who have lost speech. Advances in inner-speech decoding also raise questions about mental privacy, consent and the interpretation of generated content. Distinguishing cued from self-generated messages, and neural evidence from language-model assistance, is necessary for accurate public understanding of these capabilities. User authority over activation, message release and data retention should remain explicit as interfaces become more capable. Fluent output alone should not be interpreted as unrestricted access to thought.

\paragraph{Illustration provenance.} Figures~\ref{fig:teaser}, \ref{fig:residual} and \ref{fig:beyondlevels} include conceptual artwork generated with OpenAI's image-generation tool in Codex and independently typeset labels. Anatomy, device schematics and scientific meaning were reviewed against the cited sources. These illustrations contain no participant photographs or measured data.

\bibliography{refs}
\bibliographystyle{icml2026}

\appendix
\crefalias{section}{appendix}
\crefalias{subsection}{appendix}
\crefname{appendix}{Appendix}{Appendices}
\Crefname{appendix}{Appendix}{Appendices}
\raggedbottom
\makeatletter
\setlength{\@fptop}{0pt}
\setlength{\@dblfptop}{0pt}
\makeatother
\section{Glossary}
\label{app:glossary}
\begingroup\small
\setlength{\tabcolsep}{4pt}
\begin{longtable}{@{}p{3.4cm}p{11.6cm}@{}}
\caption{Terms and abbreviations as used in this survey.}
\label{tab:glossary}\\
\toprule
Term & Meaning here \\
\midrule\endfirsthead
\toprule

Term & Meaning here \\
\midrule\endhead
\bottomrule\endfoot

Paradigm & Articulated, Inner or Perceived, assigned from the actual condition (\cref{tab:paradigms}) \\
Task descriptor & content source, elicitation, repetition, cue modality, preparation interval and peripheral monitoring; separate from paradigm and subtype \\
Source terminology & covert and imagined are mapped to Inner only when instructions meet its definition; silent alone does not determine a paradigm \\
Articulated & actual articulation: Overt, Mouthed or Limited; Limited requires residual articulation during impaired attempts \\
Output type & what the decoder produced: candidate selection, text decoding, speech synthesis and facial/body animation (\cref{tab:outputs}) \\
Evidence scope & whether a study decoded content, detected speech state, mapped language, analysed representations, released data or evaluated methods \\
Validation conditions & test units, controls, uncertainty and verification provenance, recorded separately from output type \\
Use evidence & condition-specific participants, online evaluation and duration of actual system use \\
Bridging & training on Overt or clinical attempted speech and transferring to a specified Mouthed or Inner condition \\
Route & text, voice or meaning pipeline (\cref{fig:routes}) \\
iEEG & ECoG and sEEG collectively; \muecog = micro-ECoG \\
EXG / EMG\_aux & peripheral electrodes (EOG, EMG) recorded alongside EEG; systems that use them are flagged \\
OPM-MEG & MEG with optically pumped magnetometers \\
HD-DOT & high-density diffuse optical tomography, a dense fNIRS variant \\
fUSI, tFUS & functional ultrasound imaging (recording; in scope) vs.\ transcranial focused ultrasound stimulation (out of scope) \\
ESM & electrical stimulation mapping \\
PEEK & polyether ether ketone, the polymer of the acoustically transparent cranial window \\
\wer, CER, PER & word, character and phoneme error rate \\
MOS & mean opinion score from listeners \\
DTW & dynamic time warping, used to align synthesised and reference audio \\
RNN-T & recurrent-network transducer, a streaming sequence decoder \\
LM & language model; LLM = large language model \\
WFST & weighted finite-state transducer; combines token, lexicon and language-model constraints in decoding \\
SEM & standard error of the mean \\
LOSO & leave-one-subject-out evaluation \\

\end{longtable}
\endgroup

\section{Supplementary Signal Mechanisms}
\label{app:modalities}

The acquisition examples in \cref{tab:signal_resolution_comparison} distinguish temporal and spatial sampling from physiological response and functional localisation. The complementary map below explains why faster sampling alone cannot equate their information content. Electrical and magnetic measurements and vascular responses have different physiological origins; decoder windows add another temporal constraint. Recording access also determines which tissue can be sampled. These dependencies support choosing an acquisition configuration for the intended task and operating setting, rather than inferring decoding capability from a nominal resolution value (\cref{sec:signals}).

\begin{figure}[!htbp]
\centering
\includegraphics[width=\textwidth]{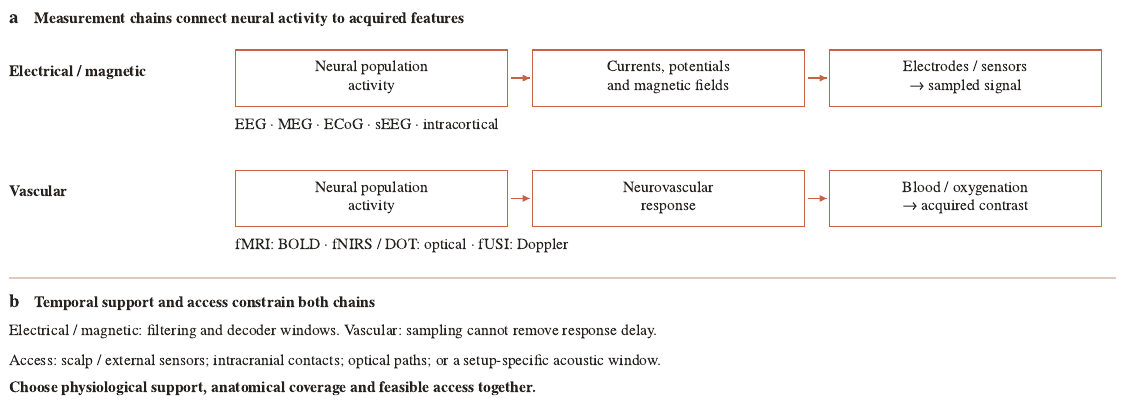}
\caption{The usable neural measurement depends on its physiological origin, temporal support and recording access. Electrical and magnetic recordings and vascular measurements observe different consequences of neural activity. Acquisition speed does not remove vascular response delay, while filtering and decoder windows also constrain electrical or magnetic information. The access map is schematic, not a ranking of spatial resolution, latency or decoding accuracy. fUSI depends on the acoustic window and preparation; its placement does not imply established word- or sentence-level speech decoding.}
\label{fig:signal_mechanisms}

\end{figure}

\begin{table}[!htbp]
\caption{Selected human-relevant fUSI reports examined here. The listed studies do not demonstrate word- or sentence-level speech-content decoding; adjacent feasibility and motor tasks are kept distinct.}
\label{tab:fus}
\vskip 0.1in
\centering\small
\begin{tabular}{@{}p{4.0cm}p{4.4cm}p{6.5cm}@{}}
\toprule
Study & Access, task & Reported evidence scope \\
\midrule
\citep{soloukey2019s0116} & intra-op; aloud and source-labelled covert repetition & language mapping; covert instructions require task-level interpretation \\
\citep{soloukey2023s0333} & intra-op; language task & mapping, fUS--fMRI--ESM overlap \\
\citep{imbault2017s0073} & intra-op; motor task & feasibility of intra-operative fUS \\
\citep{rabut2024s0437} & cranial window; motor and guitar & non-linguistic decoding \\
\citep{lin2026s0707} & cranial window; effectors, fingers & single-trial motor decoding \\
\citep{soloukey2025s0584} & polymer (PEEK) window; walking, imagined lip licking & mapping; motor imagery \\
\citep{verhoef2025s0583} & window/defect; finger task & 4-D imaging feasibility \\
\citep{vienneau2026s0769} & intact skull; breath hold & vascular reactivity \\
\citep{chen2025s0548} & skull defect; clicks & non-speech auditory mapping \\
\bottomrule
\end{tabular}
\end{table}

\section{Search Strategy Summary}
\label{app:search}
\begingroup\small
\setlength{\tabcolsep}{4pt}
\begin{longtable}{@{}p{3.2cm}p{7.2cm}p{4.6cm}@{}}
\caption{Search sources, procedures and coverage dates. Detailed queries and version links are available in the accompanying search materials.}
\label{tab:search}\\
\toprule
Step & Search procedure & Coverage / interpretation \\
\midrule\endfirsthead
\toprule

Step & Search procedure & Coverage / interpretation \\
\midrule\endhead
\bottomrule\endfoot

Broad database search & Europe PMC: 12 query groups; OpenAlex: 4 query groups; full pagination without a lower year limit & First publication through 15 September 2026 \\
Keyword coverage & Signal synonyms combined with speech, phoneme, vowel, word or language and decoding, recognition, classification, synthesis or reconstruction & Separate unspoken, attempted, watched and functional-ultrasound queries \\
Citation tracking & Forward citations of 16 reviews and key papers; backward references of surveys and fUSI reports; Crossref metadata completion & Seed-based follow-up; not an estimate of field-wide recall \\
Sanity check & 44 earlier-list studies linked by \texttt{Old\_survey\_ID}: 43 with query or citation-tracking provenance and one ISCA-archive addition & Curated seed set; not a recall estimate \\
Author and reading supplements & Author tracing added three records; reading-route searches added 36 stable IDs & 16 September; original publication cut-off retained \\
Recent-review search & Europe PMC, Crossref, publisher and repository discovery, and reference tracking for reviews from 2024 onwards & 18 September; separate review inventory (\cref{app:recentreviews}) \\
Thematic follow-up & Subsection-specific primary-source searches and recent comparative-result tables & 19--21 September; eligible reports/resources added to the updated registry \\
Paradigm follow-up & Targeted primary-source searches for the neural-processing accounts in \cref{sec:paradigms} & 10 October; cited sources not added to the registry \\
Version consolidation & DOI, normalised title and explicit preprint--publication links; aliases retain source provenance & \nrecords{} retained IDs; \nidx{} canonical report/resource entries \\
Screening and extraction & Condition-level task, signal, target and evaluation fields; cohort links recorded separately & \nkey{} entries with selected structured extraction; depth varies by claim \\
Companion resources & Dataset releases linked to corresponding articles; follow-up studies retained when they report new evidence & Entries are not independent participant cohorts \\

\end{longtable}
\endgroup

The accompanying \href{https://github.com/Uploading-Inc/Neural-to-Text-Research/tree/main/silent_speech/paper/evidence/survey_rebuild_20260919}{thematic-search records} retain the subsection queries and source checks; the \href{https://github.com/Uploading-Inc/Neural-to-Text-Research/tree/main/silent_speech/paper}{manuscript repository} provides the registry-derived counts and figure scripts. Example query structure:
\begingroup\small
\begin{verbatim}
(speech OR phonem* OR vowel* OR word* OR language OR verbal)
AND (decod* OR recogn* OR classif* OR synthes* OR reconstruct* OR neuroprosthe*)
AND (EEG OR electroencephalog* [or other signal synonyms])

Silent paradigms:  ("imagined speech" OR "inner speech" OR "covert speech"
                    OR "internal speech" OR "speech imagery" OR "silent speech")
Ultrasound:        "functional ultrasound" OR fUSI
                   OR "functional ultrasonography"
\end{verbatim}
\endgroup

\paragraph{Rule-assisted screening.} The screening log of the accompanying study workbook records one triage category for each of the 20{,}335 merged records, but not the rule definitions. Title rules prioritised 616 records indicating language-content decoding, speech-state detection or speech and language datasets, and 158 records were curated retentions under the main-text scope, including the 44 earlier-list studies; most of these 774 entries remain title-level candidates. The other records were assigned as follows: neither prioritised nor flagged (10{,}761); other ultrasound-search or fUSI-string hits, including acronym, homonym and substring noise (3{,}209); non-target outputs or tasks, such as attention, emotion, speaker identity, handwriting, clinical state, conventional speech recognition or peripheral devices (2{,}152); reviews and comments, kept as search seeds or background (2{,}036); repository copies, data versions and theses (1{,}165); peer-review reports, corrections and data copies (148); peripheral EMG or camera lip-reading input (86); and speech-processing models without brain recordings (4). Over 400 records labelled as fUSI-string hits have neither an ultrasound title term nor an ultrasound-related recorded query. These labels are triage outcomes, rather than reasons verified by reading each record. Of the entries added by the three supplements, 73 had been among the records not retained: all three author-traced studies, 23 of the 33 canonical reading-route additions and 47 source-led additions, some of which reflect that supplement's broader resource scope. Of these 73, 54 had been neither prioritised nor flagged, and 27 are labelled content-decoding reports.

\section{Recent Reviews and Their Scope}
\label{app:recentreviews}
The targeted review supplement searched 2024--2026 publications through 18 September 2026. The first part of \cref{tab:recentreviews} lists 30 directly relevant reviews; the second lists 16 contextual reviews or perspectives used in the synthesis. These are reports with different scopes, not independent replications of the studies they cite. Characterisations identify a reported focus rather than score completeness or methodological quality. A denotes abstract-level scope checking; P denotes targeted article-section checking, not complete full-text appraisal; T denotes title/metadata only. Source descriptions are retained without importing their task terminology into this survey's operational definitions.

\small
\setlength{\tabcolsep}{4pt}
\begin{longtable}{@{}p{3.3cm}p{1.55cm}p{9.45cm}p{1.0cm}@{}}
\caption{Recent reviews informing the survey. Preprints, short reviews, proceedings and online/issue-year differences are explicitly identified.}\label{tab:recentreviews}\\
\toprule
Reference & Date & Scope and publication qualification & Check \\
\midrule
\endfirsthead
\toprule
Reference & Date & Scope and publication qualification & Check \\
\midrule
\endhead
\bottomrule
\endfoot
\multicolumn{4}{@{}l}{\textbf{Directly relevant reviews}}\\
\midrule

\citep{silva2024neuroprosthesisreview} & 2024 & Speech neuroprostheses: cortical representations, output forms and clinical performance.  & A \\
\citep{rahman2024reviewr02} & 2024 & EEG speech imagery: neural basis, signal processing and learning methods.  & A \\
\citep{sensors2024review} & 2024 & EEG imagined-speech classification: paradigms, features, classifiers and datasets.  & A \\
\citep{zhang2025reviewr04} & 2025 & EEG speech imagery: datasets, input representations, deep models and evaluation. 2024 online; 2025 issue. & A \\
\citep{jiye2024s0442} & 2024 & Imagined-speech reconstruction across neural sources and methodological routes.  & A \\
\citep{murad2024review} & 2025 & EEG-to-text: acquisition, signal processing and text-generation pipelines. 2024 preprint/online; 2025 issue. & A \\
\citep{stavisky2025reviewr07} & 2025 & Clinical need, discrete and continuous speech decoding, closed-loop systems and metrics.  & A \\
\citep{qiu2025reviewr08} & 2025 & Language decoding from signal interpretation to intelligent communication.  & A \\
\citep{wang2025linguisticreview} & 2025 & Linguistic decoding across perception and production, signals and deep-learning methods.  & A \\
\citep{peerj2025review} & 2025 & EEG speech imagery: encoding paradigms, neural connectivity and decoding algorithms.  & P \\
\citep{tates2025reviewr11} & 2025 & Speech-imagery pipelines across modalities; task instructions, transfer rates and online use.  & A \\
\citep{jin2025reviewr12} & 2025 & EEG imagined speech: theory, data, features and learning methods.  & A \\
\citep{almufareh2025reviewr13} & 2025 & Inner-speech signals, models, calibration and ethical or translational constraints.  & A \\
\citep{chowdhury2025reviewr14} & 2025 & Inner-speech recognition: datasets, preprocessing and machine-learning approaches.  & P \\
\citep{jhilal2025reviewr15} & 2025 & Implantable speech decoders: patient selection, recording targets and training. 2025 online; 2026 issue. & A \\
\citep{doda2025reviewr16} & 2025 & Speech BCI models, communication rehabilitation and ethical challenges.  & A \\
\citep{yu2025reviewr17} & 2025 & Deep-learning approaches to brain-to-text and brain-to-speech interaction.  & A \\
\citep{duhair2026reviewr18} & 2026 & EEG imagined-speech decoding methods and challenges.  & A \\
\citep{estrellaibarra2026reviewr19} & 2026 & Imagined-speech EEG pipelines with linguistic, neural and computational perspectives.  & A \\
\citep{sensors2026review} & 2026 & Task granularity, constrained output spaces and output pathways.  & P \\
\citep{belfrouh2026reviewr21} & 2026 & Brain-to-speech for paralysis: acquisition, AI methods and clinical evidence appraisal.  & P \\
\citep{sumerarpak2026reviewr22} & 2026 & Non-invasive inner speech and foundation models; direct evidence and transfer prospects. Mini review. & P \\
\citep{he2026reviewr23} & 2026 & Intracranial language BCI mechanisms, hardware, algorithms, evaluation and clinical pathways. Preprint, v2; not peer reviewed. & P \\
\citep{k2026s0678} & 2026 & EEG imagined-speech techniques, applications and challenges.  & A \\
\citep{sbaih2026reviewr25} & 2026 & Deep learning for speech classification and decoding; PRISMA-organised synthesis. 2026 availability; 2027 issue. & A \\
\citep{geetha2025reviewr26} & 2025 & Brain-to-text technologies and communication aids (title-level scope only). 2025 conference; indexed 2026. & T \\
\citep{gao2026reviewr27} & 2026 & Brain-to-text, speech and image interfaces; visual reconstruction extends beyond our scope. Conference proceedings. & A \\
\citep{shrividya2025reviewr28} & 2025 & Fluency in non-invasive imagined-speech decoding; short conference review. 2025 conference; indexed 2026. & A \\
\citep{yang2025reviewr29} & 2025 & Chinese-language review of EEG speech-imagery encoding paradigms and decoding algorithms.  & A \\
\citep{motaqi2026reviewr72} & 2026 & EEG, fNIRS and hybrid imagined-speech decoding.  & A \\
\midrule\multicolumn{4}{@{}l}{\textbf{Contextual reviews and perspectives}}\\\midrule
\citep{wang2025reviewr34} & 2025 & Non-invasive foundation models: representation, alignment and generation. Preprint; not peer reviewed. & A \\
\citep{mai2025reviewr35} & 2025 & Brain-conditional multimodal synthesis and its taxonomy. 2024 online; 2025 issue. & A \\
\citep{kuruppu2026reviewr38} & 2026 & EEG foundation models: progress, evaluation and open challenges.  & A \\
\citep{li2025reviewr41} & 2025 & Domain adaptation for neural signal decoding.  & A \\
\citep{khan2025reviewr43} & 2025 & Invasive BCI communication, including channels other than speech.  & P \\
\citep{dohle2025reviewr45} & 2025 & Implantable BCI translation and decoder, task and clinical outcome measures.  & P \\
\citep{swanson2026reviewr46} & 2026 & Recalibration frequency, duration, assistance and setting for long-term use.  & A \\
\citep{wohns2024reviewr47} & 2024 & Caregiver participation in implantable BCI research and operation.  & P \\
\citep{kreiner2024reviewr52} & 2024 & The form and prosody of inner speech.  & A \\
\citep{daho2025reviewr53} & 2025 & Inner-speech phenomenology, neural correlates and clinical implications.  & A \\
\citep{li2026reviewr54} & 2026 & A perspective on moving from speech to language BCIs. Perspective. & A \\
\citep{hong2024reviewr55} & 2024 & Stimulation-induced speech perception and prospective neural return channels.  & A \\
\citep{tang2026reviewr56} & 2026 & Silent-speech sensing across neural and peripheral modalities.  & A \\
\citep{vanstuijvenberg2024reviewr64} & 2024 & Executory and guidance control in AI-assisted speech BCIs.  & P \\
\citep{oota2024reviewr67} & 2024 & Language, visual and auditory brain encoding/decoding with deep neural networks. 2024 journal; 2023 preprint. & A \\
\citep{alexander2024reviewr73} & 2024 & Multidimensional inner speech and interdisciplinary clinical evidence.  & A \\
\end{longtable}
\normalsize
\setlength{\tabcolsep}{6pt}

\section{Additional Study Coverage}
\subsection{Coverage beyond the selected performance examples}
\label{sec:coverage}

The performance examples below are a selective synthesis, whereas the
accompanying study registry contains every workbook entry. This separation
matters because bibliographic coverage, task coding and result verification
are different stages. The companion coverage table retains original registry labels; these are provenance fields, not the three-group classification. Pending entries are retained with source links and stable IDs;
their presence establishes what remains to be checked, not a positive
decoding result. The following map brings additional directions already
recorded in the workbook into the scope of the narrative. Author--year citations identify the reports discussed here; the accompanying
registry records the extraction status of each entry. Descriptions of pending work are limited to its
indexed topic or existing task note, without adding unverified accuracies.


\paragraph{Articulated: Overt and Mouthed methods.}
The index extends from phoneme localisation and classification
\citep{blakely2008s0003,mugler2014s0033} and cortical-surface spoken-word
classification \citep{kellis2010s0010} to continuous phoneme decoding
\citep{moses2016s0062} and direct ECoG synthesis
\citep{herff2016s0064,angrick2019s0126}. More recent entries address sparse
depth electrodes \citep{petrosyan2022s0286}, surface/depth Transformer models
\citep{chen2025s0630}, syntactic roles during sentence production
\citep{morgan2025s0519}, EEG continuous versus discrete outputs
\citep{craik2025s0512}, speech envelopes \citep{ivucic2025s0609}, and neural--audio
matching \citep{wang2026s0757}. They motivate separate comparisons of
classification, waveform reconstruction and retrieval. Acoustic--linguistic
dual-pathway models \citep{li2026s0712} and evaluation-specific work
\citep{wu2025s0565} also belong in that comparison once their protocols are
checked. A reported architectural improvement is not automatically evidence
of transfer to a new speaker or an unseen sentence.

\paragraph{Limited and clinical attempted-speech boundaries.}
Limited membership requires documented residual articulation; the broader clinical comparison also retains source-labelled attempts outside that boundary. The index includes attempted-speech spelling \citep{metzger2022s0266}, bilingual
neuroprostheses \citep{silva2024s0375}, stable control without recalibration
\citep{luo2023s0365}, and newer reports concerning cross-session or
cross-participant generalisation \citep{fogg2026s0639} and pontine-stroke
dysarthria \citep{nasontomaszewski2026s0756}. These directions are relevant to usability,
language coverage and clinical heterogeneity even when they do not furnish
a directly comparable word error rate. Their task-level inclusion here does
not promote pending records to verified clinical milestones.

\paragraph{Cross-condition transfer and peripheral contributions.}
Original covert or imagined labels do not themselves establish an Inner strategy. The broader index contains EEG cortical-current vowel decoding
\citep{yoshimura2016s0055}, online internal-speech classification \citep{sereshkeh2017s0074}, and
fNIRS Overt/Inner comparisons \citep{kamavuako2018s0084}. Cross-mode studies
include overt-trained ECoG Transformers \citep{komeiji2024s0436}, nested
Overt/Mouthed/Inner activity \citep{soroush2023s0369}, and speech-event
detection transferred between modes \citep{de2024s0440}. The latter is
about event timing rather than word identity. Sentence-level wearable
EEG--EMG systems \citep{satterlee2025s0605} and large-model pretraining described
as active silent-speech BCI \citep{jinzhao2025s0596} require explicit inspection
of instructions, peripheral inputs and splits. The term ``silent'' in an
indexed title alone does not establish motionless neural decoding.

\paragraph{Inner: datasets and multimodal measurement.}
Beyond the selected intracortical and EEG/MEG results, the registry includes
bimodal EEG--fMRI datasets \citep{simistira2023s0308,simistira2025s0607}, hybrid
decoding \citep{wellington2024s0443}, inter-trial-coherence features
\citep{lopezbernal2024s0434}, and Arabic inner-speech data \citep{metwalli2025s0506}.
These entries broaden language and recording coverage. Dataset availability
does not itself validate open-vocabulary decoding, and features computed
across trials require checking that test information is not shared with
training. Possible shared cohorts and dataset versions remain explicit
verification tasks rather than assumed independent replications.

\paragraph{Inner: methods, control and imagery design.}
The index spans common spatial patterns \citep{dasalla2009s0007}, syllable-rhythm
features \citep{deng2010s0011}, Overt/Inner phonetic components
\citep{pei2011s0017}, deep metric learning \citep{lee2021s0198}, and
multiscale temporal convolution \citep{li2021s0199}. Rhythm or state
classification should not be pooled with content recognition. Online fNIRS
\citep{rezazadeh2019s0121}, ear-EEG device control \citep{kaongoen2022s0239}, and adaptive
LDA feedback \citep{wu2024s0384} add interaction settings that offline
classification alone misses. Tonal imagery \citep{guo2022s0252}, delayed
dynamical features \citep{carvalho2024s0410}, Chinese word models
\citep{liu2025s0487}, and Chisco \citep{zhang2024s0394} broaden the tasks.
Additional indexed questions concern articulation and coarticulation in
datasets \citep{moreira2025s0505}, Overt-to-Inner training
\citep{alonsovzquez2025s0546}, idle-state detection \citep{padfield2025s0494}, learning to
operate a BCI \citep{bhadra2025s0575}, imagery rate \citep{guo2025s0562}, and
word-specific difficulty \citep{turk2025s0635}. EEG-to-EEG translation and neural simulation represent different uses of generative models. EEG-Translator maps imagined-speech EEG towards the overt-speech domain and reports a 6.2\% improvement in class decoding accuracy over the original imagined signals \citep{lee2025s0534}. Rajagopal et al. instead use language-model-generated text to drive phoneme-linked EEG synthesis and neurometabolic simulation \citep{rajagopal2025s0559}. The former evaluates reconstructed signals for classification; the latter simulates electrical and metabolic patterns rather than recovering text from measured brain activity. A Mandarin vowel classifier reports 49.47\% four-class accuracy and 69.83\% mean binary accuracy across ten participants under its original ``intended speech'' label \citep{wang2024s0444}. Multimodal feedback in real-time synthesis \citep{brumberg2018s0081} adds another interaction question. Non-speech visual imagery comparisons
\citep{lee2020s0168} are controls outside Perceived--Visual speech.

\paragraph{Perceived: Auditory, Visual and Audiovisual conditions.}
Perceived--Auditory decoding also includes Chinese MEG \citep{jia2025s0578}, EEG
acoustics beyond the amplitude envelope \citep{macintyre2026s0679}, cross-subject
generalisability \citep{aoke2026s0668}, and phoneme-sequence auxiliary
prediction \citep{jihwan2025s0538}. Listening to a question and speaking an
answer \citep{moses2019s0123} is a mixed task: question context can constrain
the response. Perceived feedback in a production BCI
\citep{schippers2025s0619} likewise does not make the whole experiment a passive
listening study. Non-human-primate auditory recordings
\citep{heelan2019s0110} and audiovisual representational studies
\citep{li2026s0772} belong to different evidence strata. An Inner-to-Perceived--Auditory mapping proposal \citep{maryam2026s0774} requires checking both the
training paradigm and the tested target. The preprint and formal versions of
the lip-reading report \citep{bourguignon2020s0163} count as one report.

\paragraph{Functional ultrasound: adjacent evidence remains adjacent.}
The full registry retains auditory-hierarchy mapping
\citep{bimbard2018s0095}, activity propagation in behaving primates
\citep{dizeux2019s0117}, motor-plan decoding \citep{griggs2024s0412}, human cranial
windows \mbox{\citep{rabut2024s0437}}, four-dimensional imaging
\citep{verhoef2025s0583}, crow auditory responses \citep{liao2025s0543}, and
spatiotemporal-resolution work \citep{bian2026s0657}. These records inform
measurement feasibility and experimental design; they are not additional
human speech-content decoders. EEG neural-versus-nonneural contribution
analysis \citep{sato2024s0418} is similarly retained as methodological
evidence rather than a communication milestone.

\paragraph{Reading-route evidence and remaining protocol checks.}
GLIM \citep{liu2025s0801} and Brain-CLIPLM \citep{yang2026s0806} have selected result and protocol checks supporting their use in the main synthesis (\cref{tab:claimledger}); they are no longer merely unchecked bibliographic leads. The GLIM values in the main comparison come from SemKey's shared unique-text benchmark \citep{wang2026s0791}, whereas Brain-CLIPLM's comparison quotes earlier literature values and uses a task-specific sentence corpus. Multimodal alignment and routing \citep{changrui2026s0685} likewise contributes a selected fMRI comparison in Methods. These claim-level checks do not imply a complete audit of every experiment in those reports.

The reading supplement also retains diffusion, retrieval-augmented, hierarchical and privacy-related leads
\citep{lamprou2025s0797,jeon2025s0798,elgedawy2025s0799,masry2025s0800,murad2025s0802,murhekar2026s0803,collautti2026s0804,sharma2026s0805,samanta2026s0807,khushiyant2025s0809,lamprou2024s0810,mcguire2024s0811}
for further protocol assessment. Their inclusion alone does not add verified systems to the main comparison. Study-wide registry status and the depth of a selected claim check are distinct; version aliases do not add independent studies.

\subsection{Additional published benchmark conditions}
\label{sec:benchmark_supplement}
Complementary decoding conditions and uncertainty clarify how benchmark results depend on training data, temporal context and adaptation. Parentheses indicate the source-reported SD or SEM specified in each caption.
\begin{table}[!htbp]
\centering\small
\setlength{\tabcolsep}{4pt}
\caption{LibriBrain word retrieval: top-10 balanced accuracy (\%), with the standard error of the mean (SEM) over three seeds in parentheses. The candidate set contains 50 words. Source: Table 1 \citep{jayalath2026megxl}. Model references appear in \cref{fig:method_meg_retrieval}.}
\label{tab:meg_retrieval_values}
\begin{tabular}{@{}lrr@{}}
\toprule
Method & 13\% training data & All training data \\
\midrule
BioCodec & 19.9 (0.1) & 41.9 (1.0) \\
EEGPT & 20.3 (0.4) & 22.9 (0.3) \\
BIOT & 20.6 (0.3) & 45.6 (0.4) \\
BBL & 32.1 (1.2) & 49.9 (0.3) \\
BrainOmni & 29.7 (9.3) & 63.0 (0.1) \\
LaBraM & 40.3 (0.1) & 47.7 (0.3) \\
MEG-XL & 57.3 (0.4) & 63.0 (0.4) \\
\bottomrule
\end{tabular}
\end{table}
\begin{table}[!htbp]
\centering\small
\setlength{\tabcolsep}{4pt}
\caption{LibriBrain sentence generation and matched noise inputs. Scores are displayed as percentages; WER can exceed 100\%. Parentheses show source-reported SD. The paper describes seven-seed training for the proposed model but does not enumerate Table 3 repeat counts for each baseline; differences are descriptive. All methods receive sentence boundaries; the d\textquotesingle Ascoli implementation also receives word timing. Source: Table 3 \citep{landau2026semanticbottleneck}.}
\label{tab:meg_noise_values}
\begin{tabular}{@{}lrrrr@{}}
\toprule
Method & WER $\downarrow$ & BLEU-1 $\uparrow$ & ROUGE-1 $\uparrow$ & BERTScore $\uparrow$ \\
\midrule
Brain2Semantics2Text & 192.5 (10.0) & 10.0 (0.8) & 13.2 (0.3) & 83.0 (0.1) \\
Brain2Semantics2Text (noise) & 264.8 (41.5) & 8.6 (0.9) & 11.3 (0.4) & 81.9 (0.6) \\
BrainECHO & 101.8 (3.0) & 6.1 (0.8) & 9.1 (0.7) & 82.8 (0.2) \\
BrainECHO (noise) & 101.2 (0.5) & 5.6 (0.6) & 8.5 (0.8) & 82.5 (0.2) \\
d'Ascoli word-aligned & 87.1 (0.2) & 19.0 (0.4) & 17.2 (0.4) & 82.0 (0.1) \\
d'Ascoli word-aligned (noise) & 99.4 (0.3) & 7.2 (2.4) & 5.7 (2.1) & 79.4 (0.4) \\
\bottomrule
\end{tabular}
\end{table}
\begin{table}[!htbp]
\centering\small
\setlength{\tabcolsep}{4pt}
\caption{Narratives text reconstruction across 20, 40 and 60 repetition-time (TR) windows. All metrics use the source 0--100 scale; higher is better. Source: Table 1 \citep{lu2025cogreader}. Model references appear in \cref{fig:method_fmri_benchmark}.}
\label{tab:fmri_window_values}
\begin{tabular}{@{}lrrrrr@{}}
\toprule
Method & TR & BLEU-1 & BLEU-4 & ROUGE-1 F & BERTScore F \\
\midrule
UniCoRN & 20 & 22.9 & 0.0 & 20.3 & 43.9 \\
EEG-Text & 20 & 24.6 & 1.9 & 21.9 & 44.6 \\
BP-GPT & 20 & 21.6 & 1.7 & 21.6 & 44.1 \\
PREDFT & 20 & 24.3 & 0.1 & 20.1 & 45.9 \\
CogReader & 20 & 25.4 & 2.6 & 23.4 & 46.3 \\
UniCoRN & 40 & 19.1 & 0.1 & 17.8 & 43.8 \\
EEG-Text & 40 & 20.1 & 1.3 & 24.4 & 45.4 \\
BP-GPT & 40 & 19.9 & 1.5 & 21.1 & 42.6 \\
PREDFT & 40 & 25.9 & 0.4 & 21.1 & 46.3 \\
CogReader & 40 & 31.2 & 8.2 & 29.6 & 50.0 \\
UniCoRN & 60 & 18.0 & 0.4 & 16.5 & 43.2 \\
EEG-Text & 60 & 22.1 & 1.6 & 28.1 & 47.7 \\
BP-GPT & 60 & 19.3 & 0.6 & 19.4 & 41.6 \\
PREDFT & 60 & 26.4 & 0.6 & 28.1 & 48.1 \\
CogReader & 60 & 36.2 & 12.1 & 36.2 & 53.5 \\
\bottomrule
\end{tabular}
\end{table}
\begin{table}[!htbp]
\centering\small
\setlength{\tabcolsep}{4pt}
\caption{Huth reconstruction with a supplied text prompt, kept separate from the no-prompt main comparison. Source: Table 2 \citep{changrui2026s0685}. Model references appear in \cref{fig:method_fmri_benchmark}.}
\label{tab:fmri_prompt_values}
\begin{tabular}{@{}lrrrr@{}}
\toprule
Method & BLEU-1 $\uparrow$ & ROUGE-1 $\uparrow$ & ROUGE-L $\uparrow$ & WER (\%) $\downarrow$ \\
\midrule
Tang decoder & 14.95 & 13.39 & 13.26 & 93.34 \\
BrainLLM & 16.91 & 15.81 & 15.03 & 92.16 \\
Multimodal routing & 17.35 & 15.89 & 15.28 & 91.35 \\
\bottomrule
\end{tabular}
\end{table}
\begin{table}[!htbp]
\centering\small
\setlength{\tabcolsep}{4pt}
\caption{BCIC2020-3 imagined-speech classification: means and source SDs. The first eleven rows use source Table 11 \citep{elouahidi2025reve}; the last three use source Table 2 \citep{ma2026codebrain}. All use the official trial split, with paper-specific downstream tuning. All metrics are higher-is-better. Method references appear in \cref{fig:method_pretraining_benchmark}.}
\label{tab:pretraining_trial_values}
\begin{tabular}{@{}lrrr@{}}
\toprule
Method & Balanced accuracy (\%) & Cohen\textquotesingle s $\kappa$ & Weighted F1 \\
\midrule
EEGNet & 44.13 (0.96) & 0.302 (0.012) & 0.441 (0.010) \\
EEGConformer & 45.06 (1.33) & 0.313 (0.018) & 0.449 (0.015) \\
SPaRCNet & 44.26 (1.56) & 0.303 (0.023) & 0.442 (0.011) \\
ContraWR & 42.57 (1.62) & 0.308 (0.022) & 0.441 (0.018) \\
CNN-Transformer & 45.33 (0.92) & 0.317 (0.012) & 0.451 (0.013) \\
FFCL & 46.78 (1.97) & 0.330 (0.036) & 0.469 (0.021) \\
ST-Transformer & 41.26 (1.22) & 0.294 (0.016) & 0.425 (0.014) \\
BIOT & 49.20 (0.86) & 0.365 (0.018) & 0.492 (0.008) \\
LaBraM-Base & 50.60 (1.55) & 0.380 (0.024) & 0.505 (0.021) \\
CBraMod & 53.73 (1.08) & 0.422 (0.016) & 0.538 (0.010) \\
REVE-Base & 56.35 (1.23) & 0.454 (0.015) & 0.563 (0.012) \\
BENDR & 24.85 (0.75) & 0.061 (0.009) & 0.238 (0.017) \\
EEGPT & 24.53 (1.31) & 0.057 (0.016) & 0.244 (0.011) \\
CodeBrain & 61.01 (0.52) & 0.513 (0.006) & 0.610 (0.005) \\
\bottomrule
\end{tabular}
\end{table}
\begin{table}[!htbp]
\centering\small
\setlength{\tabcolsep}{4pt}
\caption{BCI-Speech subject-disjoint balanced accuracy (\%); parentheses show SD across three runs. Source: Supplementary Table 5 \citep{lu2026omnieegbench}. Frozen probing and full fine-tuning use different adaptation procedures. Model references appear in \cref{fig:method_pretraining_benchmark}.}
\label{tab:pretraining_transfer_values}
\begin{tabular}{@{}lrr@{}}
\toprule
Method & Linear probe & Full fine-tuning \\
\midrule
BrainOmni & 22.5 (1.0) & 19.9 (0.7) \\
CBraMod & 28.9 (3.3) & 26.2 (1.0) \\
REVE & 22.3 (0.9) & 20.8 (1.2) \\
FEMBA & 21.0 (1.3) & 24.9 (0.3) \\
BIOT & 20.4 (1.4) & 26.7 (2.5) \\
LaBraM & 20.3 (1.1) & 22.5 (0.3) \\
NeuroLM & 21.1 (0.4) & 20.7 (0.5) \\
Neuro-GPT & 21.2 (0.8) & 21.0 (2.2) \\
EEGMamba & 19.7 (1.2) & 20.0 (0.8) \\
BENDR & 20.0 (0.0) & 20.0 (0.0) \\
\bottomrule
\end{tabular}
\end{table}

\section{Selected Conditions and Source Checks}
\label{app:claimchecks}
The condition overview and consolidated source-check table provide the detailed protocols behind the main comparisons. Shared participant identifiers, including T15 across Inner and long-term-use reports, indicate repeated evidence rather than additional independent recruitment. Source-verification labels describe the inspected material, not experimental quality.
\begingroup\small
\setlength{\tabcolsep}{4pt}
\begin{longtable}{@{}>{\raggedright\arraybackslash}p{2.2cm} p{3.2cm} p{4.4cm} p{4.8cm}@{}}
\caption{Selected reported outputs, validation boundaries and use evidence. Rows are illustrative, not exhaustive maxima. N is attached to a named study or condition; study totals, test-condition participants and independent clinical cohorts are not interchangeable. Source provenance for selected conclusions is in \cref{tab:claimledger}.}
\label{tab:state}\\
\toprule
Group / condition & Reported outputs / examples & Validation boundary & Participant and use boundary \\
\midrule\endfirsthead
\toprule

Group / condition & Reported outputs / examples & Validation boundary & Participant and use boundary \\
\midrule\endhead
\bottomrule\endfoot

Articulated: Overt & text and sentence synthesis (ECoG, sEEG) & 3\% \wer in a closed-set repetition condition \citep{makin2020s0165}; listener choice differs from free transcription & Makin study N=4; Chen study N=48 \citep{chen2024s0381}; different studies and conditions \\
Articulated: Mouthed & transfer of Overt-trained synthesis to unvoiced articulation & actual movement remains; Mouthed and Inner results differ \citep{anumanchipalli2019s0127,meng2023s0312} & source-specific participant and listener-test conditions; no motionless-decoding inference \\
Articulated: Limited & clinical decoding with residual articulation & residual execution must be documented; speech attempts cannot be assigned from a diagnostic label alone \citep{willett2023s0300,angrick2024s0460} & report residual movement and voicing for each evaluated condition \\
Inner & cued text, word classification, acoustic and semantic generation & strategy, content source, bridging and model priors require separate controls \citep{kunz2025s0568,youngeun2023s0372,suyi2024s0450,sparsh2025s0626} & Kunz study N=4; large-vocabulary condition N=2; unspecified imagery strategy is not Mixed \\
Perceived: Auditory & retrieval, generated text and semantic gist & candidate sets, repeated stimuli and unseen stories define different tests \citep{alexandre2023s0322,yang2024mad,tang2023s0356} & pooled retrieval datasets and small generation cohorts do not establish self-generated communication \\
Perceived: Visual / Audiovisual & Reading text, lip-reading categories and audiovisual representations & Reading protocols are detailed in \cref{tab:reading}; Lip-reading and Audiovisual use different targets \citep{karthik2024s0391,li2026s0772} & N=64 fMRI and N=14 iEEG belong to the lip-reading study, not a common audiovisual-generation cohort \\
\midrule
Attempted boundary & clinical attempted-speech results without established group membership & actual motor intention does not establish residual articulation \citep{kunz2025s0568,jude2026s0677} & retained for clinical comparison; not pooled into Limited or reassigned to Inner \\

\end{longtable}
\endgroup

\begingroup\small
\setlength{\tabcolsep}{3pt}
\begin{longtable}{@{}>{\raggedright\arraybackslash}p{2.2cm}p{3.2cm}p{4.9cm}p{4.9cm}@{}}
\caption{Selected result conditions and inference boundaries. F denotes inspected full-text sections, not a complete study-wide audit. Dataset size and the retained sample for a particular result are distinguished; an unreported field remains unreported. Version-specific locators identify the evidence actually checked. These targeted checks do not upgrade the registry's study-wide verification status.}
\label{tab:claimledger}\\
\toprule
Study / check & Condition & Source locator / permitted statement & Inference boundary \\
\midrule\endfirsthead
\toprule
Study / check & Condition & Source locator / permitted statement & Inference boundary \\
\midrule\endhead
\bottomrule\endfoot
\citep{willett2023s0300}\newline F & Clinical attempted speech (source task); intracortical; N=1; evaluation sentences & Results: Real-time speech decoding; Table 1. T12: large-vocabulary online text in a prompted task. & No aggregate claim about home use or other participants. \\
\citep{kunz2025s0568}\newline F & Cued inner speech; intracortical; N=2 in large-vocabulary condition; 4 in whole study; non-overlapping evaluation sentences; online retraining & Results: Real-time decoding of inner speech; Figure 3D; STAR Methods: Online inner speech decoder (Figure 3); Vocabulary and Sentence Selection. T15/T16: 125,000-word inner-speech decoding, WER 26-54\%; study-wide N=4 must not replace condition N=2. & Online adaptation and instructed strategies limit generalisation; no spontaneous private-thought claim. \\
\citep{card2026s0729}\newline F & Clinical attempted speech (source task); intracortical; N=1; longitudinal use; prompted-copy assessment & Results: Overview and Speech decoding; Methods: Continuous fine-tuning, Speech decoding latency, BCI system software. T15: over 3,800 h home BCI use including idle time; 99.2\% word accuracy belongs to prompted copy. & Caregiver setup remains; the 92\% conversational rating includes user-corrected sentences and is not uncorrected decoder accuracy. \\
\citep{yang2024neuspeech}\newline F & Perceived--Auditory speech; MEG; dataset N=27; retained result N not separately reported; sentence-recording pairs 8:1:1 & arXiv v3, Sections 3.1--3.2, 3.5.1 and Appendix A; Table 4: Gwilliams BLEU-1 60.30 without test-time teacher forcing. & Recording-pair holdout does not establish unseen-text generation. Table 1's 57.68 is a separate entry; v3 does not explain the difference. \\
\citep{yang2024mad}\newline F & Perceived--Auditory speech; MEG; dataset N=27; held-out story: cable spool fort & arXiv v2 (26 December 2025), Sections 4.1--4.3, Tables 1--2. Without test-time teacher forcing, BLEU-1 is 6.86 vs noise 3.87 and random sentence 5.86. & CER is 89.82 vs random-sentence CER 87.30; superiority is metric-specific. NeuGPT quotes an earlier MAD BLEU-1 of 6.94. \\
\citep{yang2024neugpt}\newline F & Perceived--Auditory speech; MEG-MASC; dataset N=27; held-out story: cable spool fort & arXiv v1, Sections 2.3.4, 3.1, 3.4.2; Table 3 reports BLEU-1 12.92 and CER 99.8. & Test-time prefix policy is not stated in the inspected methods. Table 3 has a random-sentence baseline, not a neural-input noise control; Table 2 WER 10.35 is audio ASR. \\
\citep{jo2025noise}\newline F & Reading audit; word-aligned ZuCo EEG; retained participant N not separately reported for Table 1; unique sentences; 80:10:10 split & Version of record, Methods: ZuCo datasets and EEG-to-text decoding; Figure 1; Tables 1 and 5. Tested BART/Pegasus/T5 implementations showed similar EEG/noise performance. & Table 5 gives sample counts, not the retained participant denominator. The audit does not directly evaluate every successor, COFETT, or MEG decoder. \\
\citep{liu2025s0801}\newline F & Reading; word-aligned ZuCo EEG; unique-text split; autoregressive generation & arXiv v1, Section 4.1 and Appendix A.4 explicitly describe generation without teacher forcing and text-disjoint partitions. Original Table 1 separates original-reference and multiple-variant scores. & The main survey's BLEU-1 7.84 and identification 10.74\% are SemKey's shared-benchmark results \citep{wang2026s0791}, not GLIM's original Table 1. \\
\citep{yang2026s0806}\newline F & Reading; fixation-aligned ZuCo EEG; fixed keyword bank; task-specific sentence corpus; descriptive retrieval N=30 & Version of record, Sections 3.1--3.4 and 4.3--4.4; Tables 2--4. Retrieval summaries use participant-level SD; matched-material inferential analyses use N=12. & Table 3 quotes earlier literature baselines rather than rerunning every method. The retrieval pool is the full task sentence set; this is corpus-assisted reconstruction, not unrestricted unseen-text generation. \\
\citep{zhang2026s0792}\newline F & COFETT corpus: cued recall after reading; 128-channel EEG; N=2 completing the full experiment; session holdout with repeated sentences & ACL version, Sections 3.1--3.2, 4.1--4.2 and Appendix A; Tables 1--2 separate ZuCo reruns from COFETT's embedding-correlation endpoint. & The Methods generation comparison uses the ZuCo reruns, not the N=2 COFETT corpus. Correlation under repeated-content session holdout does not establish generation of novel sentences. \\
\end{longtable}
\endgroup

\begin{table}[!htbp]
\caption{Ten reporting questions connect a measured result to its intended interpretation. Answering them makes task, output, generalisation and operating conditions explicit.}
\label{tab:checklist}
\vskip 0.1in
\centering\scriptsize
\begin{tabular}{@{}p{5.1cm}p{10.1cm}@{}}
\toprule
Field & Question answered \\
\midrule
Paradigm/subtype, actual articulation, cue type & Which linguistic process and information source are being decoded? \\
Output type and class / candidate / vocabulary count & What output space is available, and what counts as success? \\
Train / validation / test unit & Does the test introduce new observations, new content or both? \\
Cross-participant, cross-session, cross-day & How well does performance transfer across recording conditions? \\
LM-only, shuffled-signal, cue-removed, noise-matched controls & What does aligned neural input add beyond cues and model priors? \\
EMG, eye, head and acoustic cross-talk (EEG/MEG); instrument cross-talk (invasive) & How much information is carried by neural and peripheral channels? \\
All participants, successful subset, best individual & How are outcomes distributed across the evaluated participants? \\
Truly online, pseudo-online, offline replay & Does evaluation include live initiation, causal feedback and user adaptation? \\
Update step, algorithm latency, haemodynamic delay, end-to-end latency & When does useful feedback arrive, and how long does a corrected exchange take? \\
Independent cohort, repeated cohort, public-data reuse & Which results add new participants, and which reuse earlier recordings? \\
\bottomrule
\end{tabular}
\end{table}

\section{Dataset and Resource Inventory}
\label{app:datasets}
Neural-language resources span controlled production and imagery, attempted communication, listening, reading and naturalistic interaction. The 110 resource/version entries below distinguish recruited, released and analysed cohorts where these differ. Continuous recording hours include within-run pauses; task-window hours describe only the stated epochs or stimulus intervals. Trials, runs and linguistic items provide complementary measures of scale. Derived files, simultaneous modalities and successive releases can share recordings or participants and are not independent additions. Access descriptions were checked on 20--21 September 2026; public annotations or model responses do not imply public raw neural signals.
\begingroup\fontsize{8}{8.5}\selectfont\setlength{\tabcolsep}{4pt}\renewcommand{\arraystretch}{1.05}
\begin{longtable}{@{}>{\raggedright\arraybackslash}p{3.1cm}>{\raggedright\arraybackslash}p{1.9cm}>{\raggedright\arraybackslash}p{2.45cm}>{\raggedright\arraybackslash}p{2.35cm}>{\raggedright\arraybackslash}p{5.3cm}@{}}
\caption{Neural-language data resources and versions. Each amount retains its denominator: participants, continuous recordings, task epochs, trials or stimulus items. Approximate and protocol-derived durations are identified explicitly. Resource citations identify the report or deposit; the access and scope column distinguishes raw signals, derived releases and shared acquisitions.}\label{tab:dataset_inventory}\label{tab:datasets}\\
\toprule
Resource and citation & Signal; participants & Task & Recording amount & Access, scope and relationships \\
\midrule\endfirsthead
\multicolumn{5}{l}{\small Dataset and resource inventory (continued)}\\
\toprule
Resource and citation & Signal; participants & Task & Recording amount & Access, scope and relationships \\
\midrule\endhead
\bottomrule\endfoot
KARA ONE\par \citep{dataset_zhao2015_6d3ab} & EEG + face/\allowbreak audio\par N=14 & Imagined and overt prompts & Nominal 5-s imagery trials & \textbf{Academic/\allowbreak nonprofit use}\par Typically 132 trials; MM08 has 131, while P02 provides 165 retained feature entries. \\
UNL-CONICET /\allowbreak  Coretto\par \citep{germn2017s0075} & EEG\par N=15 & Imagined/\allowbreak overt vowels and commands & 9,314 epochs; 10.35 h & \textbf{Author Drive; sign-in}\par Four-second task epochs; excludes visit/\allowbreak setup time. \\
ASU /\allowbreak  Nguyen\par \citep{nguyen2018s0091} & EEG\par N=15 & Cued speech imagery & 6,600 planned trials & \textbf{Author release; MOABB mirror}\par Condition subsets contain 8/\allowbreak 6/\allowbreak 6/\allowbreak 6 overlapping participants. \\
Broderick\par \citep{dataset_broderick2018_8c331} & EEG\par N=19 & Narrative listening & $\approx$ 19 h listening & \textbf{Public; CC0}\par Forward-speech subset; other Dryad conditions excluded. \\
DTU auditory attention\par \citep{dataset_fuglsang2018_459ce} & EEG\par N=18 released; 19 recorded & Selective listening to competing Danish speech & Approximately 18 h continuous EEG; 1,080 trials & \textbf{Public raw + processed EEG}\par The 50-s task segments alone total 15 h. \\
ZuCo 1.0\par \citep{hollenstein2018s0778} & EEG + eye tracking\par N=12 & Natural/\allowbreak task-specific reading & 22.16 h; 273 raw blocks & \textbf{Public; CC BY}\par Summed original EEG samples; excludes setup time. \\
ASU online adaptive MI/\allowbreak SI\par \citep{dataset_nguyen2019_f707f} & EEG\par N=8 & Speech and motor imagery with adaptive online feedback & 56 runs; 2,240 trials & \textbf{Public release; CC0}\par Half the four classes are speech imagery. Each trial includes 2 s preparation and 8 s imagery; overlapping analysis windows are not extra trials. \\
Brennan--Hale Alice EEG\par \citep{dataset_brennan2019_7d11b} & EEG\par N=49 released; 52 recruited; 33 analyzed & Listening to Alice in Wonderland & 10.13 h story-aligned support; 49 recordings & \textbf{Public raw + preprocessing metadata}\par The 42 preprocessed datasets and 33-case analysis are subsets of the 49-case release. \\
FEIS\par \citep{dataset_wellington2019_2f194} & EEG\par N=23 & Phoneme cue, imagery and overt speech & 21.28 h retained EEG & \textbf{Public; ODC-BY}\par 21 English plus two Chinese participants; phase duplicates excluded. \\
KUL auditory attention\par \citep{dataset_das2019_e9ccf} & EEG\par N=16 & Selective listening to two Dutch speakers & Approximately 19.2 h; 20 trials per participant & \textbf{Public processed EEG; raw by request}\par Protocol-based duration includes repeated excerpts. \\
ZuCo 2.0\par \citep{hollenstein2019s0779} & EEG + eye tracking\par N=18 & Natural/\allowbreak task-specific reading & 21.05 h; 252 raw blocks & \textbf{Public; CC BY}\par One of 19 recruits excluded; fourteen blocks per released participant. \\
ERP CORE: N400 language task\par \citep{dataset_kappenman2021_25e4b} & EEG\par N=40 released; 39 N400 analysis & Visual word-pair relatedness judgments & 40 N400 task recordings & \textbf{Public release and processing code}\par The other five ERP tasks are outside the language subset; excluded participants remain in the public files. \\
Russian silent commands\par \citep{vorontsova2021s0224} & EEG\par N=268 in Methods demographic table & Silent and aloud Russian command reading & 5--14 sessions per participant; 10 trials per session & \textbf{Study corpus; no raw repository identified}\par Eight commands plus pseudoword control; abstract lists 270 participants. \\
BCI Competition 2020 Track 3\par \citep{dataset_jeong2022_15320} & EEG\par N=15 released & Five-class imagined command speech & 6,000 epochs; 3.33 h of imagery windows & \textbf{Public MATLAB datasets; OSF}\par Each 3.1-s epoch contains a 2-s imagery task. \\
Thinking Out Loud /\allowbreak  Nieto\par \citep{nieto2022s0292} & EEG + EOG/\allowbreak EMG\par N=10 & Overt/\allowbreak inner speech and visualization & 5,640 trials; 30 recordings & \textbf{Public OpenNeuro}\par Visualization is a separate control condition. \\
DAIS\par \citep{bo2023s0315} & EEG\par N=20 & Reading, covert and overt speech & 24,370 epochs; 13.54 h & \textbf{DANS; request required}\par Two-second epochs including rest; validation uses 15 participants. \\
Padfield motor/\allowbreak speech imagery\par \citep{dataset_padfield2023_e6c37} & EEG\par N=5 & Four imagined direction words; separate motor imagery and idle conditions & 800 speech trials; 0.89 h & \textbf{Public release; CC BY 4.0}\par Speech trials last 4 s. Motor imagery, idle trials and intertrial fixation are excluded from the speech amount. \\
ArEEG\_Chars\par \citep{verified_d23_release,hazem2024s0389} & EEG\par N=30 reported & Arabic character imagery & 929 CSVs; 2.80 h raw file coverage & \textbf{Public; CC BY}\par Raw trial-file coverage; inconsistent clock labels; 31 characters. \\
ArEEG\_Words\par \citep{verified_d24_release,hazem2024s0390} & EEG\par N=22 reported & Arabic word imagery & 359 CSVs; 1.08 h raw file coverage & \textbf{Public; CC BY}\par Raw trial-file coverage; includes distinct repeated word recordings. \\
Chisco\par \citep{zhang2024s0394} & EEG\par N=3 & Sentence reading and imagery & 20,043 trials; $>$45 h & \textbf{Public; CC0}\par Three released participants selected from eleven screened. \\
EEGIS\par \citep{dataset_lara2024_ea7de} & EEG\par N=10 & Eight imagined Spanish words and rest & 4,044 one-second windows & \textbf{Public release; CC BY 4.0}\par Windows overlap by 48 of 128 samples; raw and five filtered bands describe the same recordings. \\
IIST BCI 1: Common Malayalam words\par \citep{dataset_nair2024_489f5} & EEG\par N=1 & 26 Malayalam words: overt and imagined; overt English equivalents & 780 word trials & \textbf{IEEE DataPort subscription}\par Variable-length, manually bounded trials; visit duration is not summed EEG time. \\
IIST BCI 2: Common Marathi words\par \citep{dataset_tayade2024_48328} & EEG\par N=1 & 100 Marathi words, overt and subvocal; overt English equivalents & 3,000 word trials & \textbf{IEEE DataPort subscription}\par Variable-length, manually bounded trials; visit duration is not summed EEG time. \\
IIST BCI 3: 100 Malayalam words\par \citep{dataset_s2024_7e576} & EEG\par N=1 & 100 Malayalam words, vocal/\allowbreak subvocal; vocal English translations & 3,000 word trials & \textbf{IEEE DataPort subscription}\par Variable-length, manually bounded trials; visit duration is not summed EEG time. \\
IIST BCI 4: 100 Telugu words\par \citep{dataset_likhitha2024_97664} & EEG\par N=1 & 100 Telugu words and English equivalents, vocal and subvocal & 4,000 word trials & \textbf{IEEE DataPort subscription}\par Four language/\allowbreak production conditions; manually bounded, variable-length word records. \\
IIST BCI 5: Malayalam vowels/\allowbreak consonants\par \citep{dataset_sunil2024_b5cda} & EEG\par N=1 & Malayalam letters, vocal and subvocal & 1,120 letter trials & \textbf{IEEE DataPort subscription}\par Variable-length, manually bounded trials; visit duration is not summed EEG time. \\
IIST BCI 6: Common Odia words\par \citep{dataset_sahoo2024_bbbfa} & EEG\par N=1 & 100 Odia words and English translations, each vocal and subvocal & 4,000 word trials & \textbf{IEEE DataPort subscription}\par Variable-length, manually bounded trials; visit duration is not summed EEG time. \\
IIST BCI 7: Human space missions\par \citep{dataset_menon2024_bbbfa} & EEG (8 channels)\par N=1 & Vocal and subvocal English/\allowbreak Malayalam words & 20,000 reported trials; 100 words per language & \textbf{IEEE DataPort subscription}\par 50 trials per word and mode. Subvocal is described as silent thought; no fixed trial duration or total hours are reported. \\
IIST BCI 8: Common Telugu words\par \citep{dataset_boddapu2024_6abc8} & EEG\par N=2 & 100 Telugu words and English translations, vocal and subvocal in both languages & 8,000 word trials & \textbf{IEEE DataPort subscription}\par Variable-length, manually bounded trials; visit duration is not summed EEG time. \\
Imperial ear-EEG AAD\par \citep{dataset_thornton2024_69b95} & In-ear EEG\par N=18 & Auditory attention in competing-speech scenes & Approximately 12 h; 288 trials & \textbf{Public EEG}\par Duration uses the reported mean of 150 s per trial. \\
MSEEG\par \citep{dataset_mseeg2024_bfb7a} & EEG\par N=10 & Imagined, intended and spoken syllables & 16.98 h retained EEG & \textbf{Public; CC BY}\par Continuous files contain preprocessing; mode archives overlap. \\
Peking University ear/\allowbreak scalp EEG\par \citep{dataset_haolin2024_24e83} & Scalp + around-ear EEG\par N=16 & Spatial auditory attention to Mandarin speech & Approximately 10.67 h per simultaneous montage & \textbf{Public EEG; original audio by request}\par Peking University dataset; scalp and ear signals share acquisition time. \\
Proposed EEG Vowel Dataset\par \citep{iqbal2024s0388} & EEG\par N=5 recorded & Five-vowel imagery & 2,125 one-minute blocks; 35.42 h of imagery & \textbf{Data available on reasonable request}\par One-second feature windows are subdivisions of the same recordings. \\
SparrKULee\par \citep{dataset_accou2024_14921} & EEG\par N=85 & Clean/\allowbreak noisy story listening & 168 h & \textbf{Public + restricted; CC BY-NC}\par Continuous Dutch speech; final 2024 release. \\
UBA natural dialogue EEG\par \citep{dataset_gonzalez2024_08e83} & EEG hyperscanning\par N=20 released (10 dyads); 18 analyzed & Unscripted cooperative Spanish dialogue & 249 paired trials; 10.69 participant-h of EEG & \textbf{Public artifact-corrected EEG, audio and annotations}\par 5.35 pair-h; 512 Hz trial files with inter-trial gaps removed. Release includes the excluded dyad. \\
Aguilera paradigm comparison\par \citep{aguilerarodrguez2025s0502} & EEG\par N=15 & Paced and gamified speech imagery & 3,600 imagery instances; nominal 1.4 h & \textbf{Public; CC BY}\par Same participants; three paced instances are nested within each long trial. \\
ArEEG (five commands)\par \citep{metwalli2025s0506} & EEG\par N=12 & Arabic command imagery & 4,650 trials; 186 sessions & \textbf{Public OpenNeuro}\par One participant completes 21 sessions; others complete 15. \\
CIRE\par \citep{dataset_he2025_5e3d5} & EEG\par N=38 released; 40 recruited & Mandarin sentences with contrasting prosody & 51.26 h; 188 raw blocks & \textbf{Public raw + derivatives}\par Two of 190 planned blocks absent; 5-s analysis epochs are a different duration basis. \\
Chinese spoken-word production\par \citep{dataset_tan2025_a1857} & EEG\par N=87: 27 /\allowbreak  23 /\allowbreak  19 /\allowbreak  18 & Overt picture naming with priming/\allowbreak interference & 87 raw recordings across four experiments & \textbf{Public raw + derivatives}\par Independent cohorts; raw, epoched and averaged data reuse the same recordings. \\
ChineseEEG\par \citep{verified_d29_release,dataset_mou2024_ee5c5} & EEG + eye tracking\par N=10 released; 15 recruited & Silent reading of two Chinese books & 121.74 h; 245 raw EEG runs & \textbf{Public raw + derivatives}\par Five of 250 planned runs absent from the current release. \\
ChineseEEG-2\par \citep{dataset_chen2025_0be34} & EEG\par N=12: 4 reading, 8 listening & Reading aloud; passive listening & 32.4 h: 10.8 reading + 21.6 listening & \textbf{Public EEG + embeddings}\par Disjoint task cohorts; original participant audio withheld. \\
Cueless imagined speech\par \citep{dataset_derakhshesh2025_e9a98} & EEG\par N=11; 55 sessions & Self-selected imagined Persian commands & 14.82 h; 55 raw FIF recordings & \textbf{Public raw + derivatives}\par A timing cue is present, but it does not specify the imagined word. \\
ETS /\allowbreak  Xiong\par \citep{xiong2025s0617} & EEG\par N=16 recruited; 13 analyzed & Four-word Chinese reading and imagery & 1,560 EEG segments; 120 per retained participant & \textbf{Author inquiry; no linked raw repository}\par Reading and imagery segments are paired within trials. \\
Ear-SAAD\par \citep{dataset_geirnaert2025_70c28} & Scalp + around-ear + in-ear EEG\par N=15 & Auditory attention with concurrent ear/\allowbreak scalp recording & 15 h per simultaneous montage; 90 trials & \textbf{Public raw BIDS + processed data}\par Three montages observe the same 15 participant-hours. \\
IIST BCI 10: Telugu vowels/\allowbreak consonants\par \citep{dataset_boddapu2025_cf78f} & EEG\par N=3 & Telugu vowels and consonants, vocal and subvocal & 52 character classes; two modes & \textbf{IEEE DataPort subscription}\par Vowels and consonants recorded on separate days. \\
IIST BCI 11: Malayalam letters\par \citep{m2025s0555} & EEG\par N=6 & Malayalam character vocalization and imagery & 6,120 recordings; 51 characters & \textbf{IEEE DataPort subscription}\par Ten trials per character and mode per participant; raw and processed files overlap. \\
IIST BCI 12: Telugu dialect commands\par \citep{dataset_vivek2025_44908} & EEG\par N=2 & 87 Telugu command variants in six classes, spoken and imagined without lip/\allowbreak tongue movement & 2,610 word trials & \textbf{IEEE DataPort subscription}\par Variable-length, manually bounded trials; visit duration is not summed EEG time. \\
IIST BCI 9: Malayalam dialect commands\par \citep{dataset_k2025_75c87} & EEG\par N=10 & Malayalam navigation commands, vocal and subvocal; 15 dialect sets, seven commands per set & 2,100 word trials & \textbf{IEEE DataPort subscription}\par Variable-length, manually bounded trials; visit duration is not summed EEG time. \\
Matin/\allowbreak Faraji imagined speech\par \citep{dataset_matin2025_a030a} & EEG\par N=9 & Imagined words with visual, auditory, combined or self-generated cues & 9 subject collections; 2.5-s epochs & \textbf{Public release; CC BY 4.0}\par Continuous EDF, processed epochs and event logs represent the same sessions; processed copies are not extra recordings. \\
Moreira articulation corpus\par \citep{moreira2025s0505} & EEG (TMS)\par N=8 + 16 released & Phoneme/\allowbreak word discrimination and repetition & 3,742 + 10,888 released trials & \textbf{Public OpenNeuro}\par 2019/\allowbreak 2021 cohorts; single-phoneme listening and repetition share trials. \\
Native Arabic Silent Speech /\allowbreak  SSDC\par \citep{dataset_chowdhury2025_cdc0f} & EEG\par N=10 recorded & Inner and overt Arabic commands & 8,994 study-wide trials; 6.25 h of speech-task windows & \textbf{Registered teams; signed agreement; registration closed}\par Challenge shares inner speech only; overt speech was not shared. Separate public tutorial sample. \\
SS-EEG\par \citep{dataset_zhou2025_6cc0f} & EEG\par N=12 & Silent articulation & 72,000 reported trials & \textbf{Author EULA; teaser available}\par Attempted articulatory movement without sound. \\
3M-CPSEED\par \citep{ma2025s0486} & EEG\par N=20 & Overt, silently articulated and imagined Pinyin & 29.39 h; 80 raw recordings & \textbf{Public; release CC0, paper CC BY 4.0}\par Hours count original EDF recordings only; processed files are also released. \\
Arabic--Hindi L2 word reading\par \citep{aldhaheri2025learningeeg,aldhaheri2026multifeature,dataset_aldhaheri2026_18f57} & EEG\par N=20 released IDs; 20 analyzed in 2026 report & Arabic/\allowbreak Hindi word learning with visually cued vocalization & 296 EDF files; 4.60 h after overlap removal & \textbf{Public EDF, samplewise and feature CSVs; CC BY 4.0}\par Release description reports 300 files; 12 s of repeated coverage removed. \\
COFETT /\allowbreak  Chisco-2.0\par \citep{zhang2026s0792} & EEG\par N=7 screened; 2 released & Chinese sentence reading and cued recall & 64 runs; 62.12 h continuous EEG; 17,056 trials & \textbf{Public raw and processed data; OpenNeuro}\par Reading, recall and rest phases belong to each trial. \\
Directional words: Russian/\allowbreak Spanish\par \citep{dataset_kostulin2026_d81df} & EEG; EMG subset\par N=22: 12 Russian, 10 Spanish & Overt and covert directional words & 24.39 h; 22 raw EDF recordings & \textbf{Public raw + derivatives}\par Four Russian participants have series markers for multiple utterances. \\
EMA--EEG speech production\par \citep{dataset_friedrichs2026_5d926} & EEG + EMA + audio\par N=29; 25 complete tri-modal cases & Habitual and rapid syllable production & 8,700 DDK trials; 300 per participant & \textbf{SWISSUbase; CC BY 4.0}\par The reported 17 h covers EMA/\allowbreak audio; additional reading and anatomy tasks do not imply EEG. \\
Japanese speech\par \citep{motoshige2026s0636} & EEG + EMG\par N=3 & Overt, listening and covert speech & 1,020 h continuous & \textbf{Public; CC0}\par Overt event support is 457.5 h; shared listening/\allowbreak covert runs counted once. \\
KUL-AAD-NFB\par \citep{dataset_rotaru2026_bcee4} & EEG\par N=11 released; 19 recorded; 12 released sessions & Auditory attention with neurofeedback/\allowbreak calibration & 15.52 h; 122 raw BDF files & \textbf{Public raw + metadata}\par One participant contributes two sessions; total includes calibration. \\
Mandarin multi-condition /\allowbreak  Wang\par \citep{wang2026s0671} & EEG\par N=1 recorded & Overt, noisy overt, silent articulation and imagined Chinese speech & 720 vowel and 480 word trials reported & \textbf{Raw data on request}\par Four task conditions; publication hours do not reconcile with trial timing. \\
TESSCCo\par \citep{mario2026s0637} & EEG\par N=24 released (21 native, 3 non-native Spanish) & Overt and silent articulation of bilingual TV commands & 7,936 epochs; 11.02 h of cue-plus-production segments & \textbf{Public raw and processed data; CC BY 4.0}\par 45 sessions; silent articulation includes mouth movement. \\
UGR-MINDVOICE\par \citep{dataset_valescortina2026_462d9} & EEG + audio\par N=15; 22 sessions & Overt/\allowbreak covert Spanish syllables, words and pseudowords & 38.04 h; 22 raw EDF recordings & \textbf{Public raw + derivatives}\par Overt and covert trials alternate within sessions; repeated visits are not new participants. \\
Armeni\par \citep{dataset_armeni2022_d4df1} & MEG\par N=3 & Audiobook listening & $\approx$ 30 h & \textbf{Donders data-use agreement}\par Ten stories over ten days per participant. \\
MEG-MASC\par \citep{gwilliams2023s0341} & MEG\par N=27 & Story and word-list listening & $\approx$ 49 h; 49 sessions & \textbf{Public; CC0}\par Five participants have one session; 22 have two. \\
LibriBrain\par \citep{miran2025s0576} & MEG\par N=1 & Audiobook listening & 52.32 h listening; 93 public sessions & \textbf{Public; CC BY-NC}\par Original release; included in LibriBrain100. \\
Chen vocoded audiobook MEG\par \citep{dataset_chen2026_vocoding,dataset_chen2023_c59ce} & MEG\par N=24 & Audiobook listening at six intelligibility levels & 175 runs; 22.30 h & \textbf{Public release; ANC data-use agreement}\par The 2026 release accompanies a 2023 article. Run coverage differs across participants; recorded time includes within-run pauses. \\
LPP MEG listening\par \citep{dataset_corentinbel2026_dc8f1} & MEG\par N=58 & Listening to Le Petit Prince & 94.81 h; 521 raw runs & \textbf{Public raw + annotations}\par One participant has eight rather than nine released runs. \\
LPP MEG reading\par \citep{dataset_corentinbel2026_06a08} & MEG\par N=50 & Rapid serial visual reading of Le Petit Prince & 63.99 h; 450 raw runs & \textbf{Public raw + annotations}\par Separate reading collection; no assumed cohort overlap with the listening release. \\
LibriBrain100\par \citep{francesco2026s0727} & MEG\par N=33 & Audiobook/\allowbreak sentence listening & 104.2 h & \textbf{Public; CC BY-NC}\par 80.5 h deep subject plus 23.7 h broad cohort; includes LibriBrain. \\
MegNIST\par \citep{dataset_kwon2026_8150b} & MEG\par N=1 released & Covert digit production & 12 runs; 12,000 trials; 5.20 h continuous MEG & \textbf{Public BIDS data; CC BY-NC 4.0}\par Digit presentation windows total 2.5 h. \\
Riegel audiovisual matrix sentences\par \citep{dataset_riegel2026_59f2f} & MEG\par N=32 & Audiovisual and visual-only sentence perception & 13,359 sentence segments & \textbf{Public release; CC BY 4.0}\par Two acquisition sessions per participant; archive omits 81 planned trials listed in the README. Audio-only calibration is separate. \\
StudyForrest audio movie 7T\par \citep{dataset_hanke2014_63327} & fMRI\par N=20 & Forrest Gump audio-description listening & 160 runs; 39.99 h & \textbf{Public release}\par 7 T, partial-brain coverage. Includes all acquired story-run volumes; audiovisual follow-up reuses 15 participants. \\
Wehbe Harry Potter\par \citep{dataset_wehbe2014_2a2b8} & fMRI\par N=9 recruited; 8 retained & Word-paced Harry Potter reading & 4 runs per participant; about 6 h retained & \textbf{Public author-hosted data}\par About 45 min per participant; excluded subject is not added. \\
Huth semantic maps\par \citep{dataset_huth2016_c0291} & fMRI\par N=7 & Natural narrative listening & 84 story runs & \textbf{Public maps; tutorial subset}\par Study has 12 story runs per person. Map viewer and tutorial subset do not represent the entire raw cohort. \\
StudyForrest audiovisual 3T\par \citep{dataset_hanke2016_afa97} & fMRI\par N=15 & Audiovisual Forrest Gump viewing with eye tracking & 120 runs; 29.99 h & \textbf{Public release; PDDL}\par New 3 T recordings from 15 participants in the original 20-person audio cohort; shared recruitment, different acquisitions. \\
Narrative Brain Dataset\par \citep{dataset_lopopolo2018_d54c6} & fMRI\par N=24 & Three Dutch stories and reversed versions & 16.02 h; 65,540 released volumes in 144 runs & \textbf{Public processed fMRI + annotations}\par First four volumes per run removed; this is processed support, not raw acquisition. \\
Pereira linguistic meaning\par \citep{dataset_pereira2018_751e7} & fMRI\par N=20 recruited; 16 retained & Concept presentation and sentence reading & 180 concepts; 384- and 243-sentence experiments & \textbf{Author activation-data release: MIT/\allowbreak Dropbox; materials: OSF}\par Sentence experiments contain overlapping participant subsets. \\
Naturalistic Neuroimaging Database\par \citep{dataset_aliko2020_400fd} & fMRI\par N=86 & Ten full-length audiovisual films & 264 raw BOLD runs; 161.20 h & \textbf{Public release}\par One film per participant. Raw acquisition support is distinct from the 160.59 h of participant-weighted movie exposure. \\
Narratives\par \citep{dataset_nastase2021_ee80a} & fMRI\par N=345 & Narrative listening & 153.96 h story-aligned BOLD & \textbf{Public; CC0}\par 369,496 story TRs; 4.6 h unique stimuli. \\
LeBel\par \citep{dataset_lebel2023_e4238} & fMRI\par N=8 & Narrative listening & 81 h BOLD & \textbf{Public OpenNeuro}\par Published eight-person release includes three extended participants. \\
Tang semantic reconstruction\par \citep{tang2023s0356} & fMRI\par N=7 studied; 3 in public decoding release & Heard and imagined narratives, movies and attention & 3 imagined-story runs, 14 min each; other decoding runs & \textbf{Public data and code; resistance data on request}\par Earlier story-training data overlap the UT Austin release. \\
Tuckute drive/\allowbreak suppress language\par \citep{dataset_tuckute2024_91d70} & fMRI\par N=14 & Reading sentences selected to drive or suppress language responses & 274 main-task runs & \textbf{Public derived data}\par Five training, five new evaluation and four additional blocked-design participants. Public responses are processed sentence-level data. \\
LPPC multilingual fMRI\par \citep{verified_d46_release,dataset_li2022_48d17} & fMRI\par N=112: 49 English, 35 Chinese, 28 French & Listening to Le Petit Prince & 181.93 h; 1,011 acquisition runs & \textbf{Public raw + derivatives}\par Three simultaneous echoes count once; three short interrupted runs are included. \\
Fedorenko sentence meaning\par \citep{dataset_fedorenko2016_69ce2} & ECoG\par N=6 original participants; 5 in benchmark derivative & Sentence and control-string reading & 52 sentences; 416 word responses at 97 electrodes & \textbf{Public processed Brain-Score benchmark}\par Benchmark derivative differs from the full original cohort and conditions. \\
SingleWord\allowbreak Production\allowbreak Dutch-iBIDS\par \citep{verwoert2022s0248} & sEEG + audio\par N=10 patients & Overt production of 100 Dutch words & 0.83 h task support; 1,000 trials & \textbf{Public iBIDS + aligned audio}\par Task support uses 2-s word display + 1-s fixation per trial. \\
T12 speechBCI /\allowbreak  B2T-24\par \citep{willett2023s0300} & Intracortical\par N=1 released (T12) & Attempted sentence production & 12,100 sentence trials reported; 10,850 used for training & \textbf{Public neural features; Dryad}\par Competition splits and subsequent corrections share the same acquisition. \\
Brain Treebank\par \citep{dataset_wang2024_f407f} & sEEG\par N=10 & Audiovisual movie viewing & 43 h & \textbf{Public; CC BY}\par Clinical monitoring cohort; includes visual context. \\
Jamali single-cell semantics\par \citep{dataset_jamali2024_002cf} & Single units\par N=13 & Listening to sentences, word lists and narrative controls & 21 sessions; 287 units & \textbf{Public derived data; CC BY 4.0}\par Ten microarray participants and three Neuropixels participants; word-aligned responses do not specify continuous recording hours. \\
T15 rapidly calibrating BCI\par \citep{card2024s0386} & Intracortical\par N=1 participant (T15) & Attempted text communication and conversation & 1,718 copy-evaluation and 22,126 conversation records & \textbf{Public evaluation records; Dryad}\par 2024 evaluation release; the 2025 neural expansion uses the same cohort. \\
Kunz inner-speech motor cortex\par \citep{kunz2025s0568} & Intracortical\par N=4 studied; 2 in large-vocabulary inner-speech condition & Attempted, cued inner and other controlled speech tasks & 22 participant-dates; 11 task archives & \textbf{Public neural features and code}\par Condition-specific cohorts overlap T12 and T15 resources. \\
T15 /\allowbreak  Brain-to-Text 2025\par \citep{dataset_braintotext2025_915ad} & Intracortical\par N=1 released (T15) & Attempted speech to text & 45 sessions; 10,948 trials; 53.92 h of feature windows & \textbf{Public neural features; Dryad and competition}\par 512 features at 20 ms: threshold crossings and spike-band power; same T15 release family. \\
VocalMind\par \citep{he2025s0633} & sEEG\par N=1 & Overt, mimed and imagined speech & 67.85 min task epochs & \textbf{Public; CC BY}\par 110 retained contacts; three modes share one cohort. \\
Cai natural language production\par \citep{dataset_cai2026_6c6ce} & Single units; LFP\par N=8 & Natural conversational speech production & 14 sessions; 17.51 h & \textbf{Public derived data; CC BY 4.0}\par Continuous firing-rate time includes conversation pauses and listening. Private utterance transcripts are withheld; word timing and model embeddings are released. \\
MOUS\par \citep{dataset_schoffelen2019_0130a} & MEG + fMRI\par N=204: 102 auditory, 102 visual & Sentence and word-list comprehension & 240 planned MEG + 120 fMRI trials per participant & \textbf{Registration + Data Use Agreement}\par Three MEG cases are incomplete; separate rest recordings also have documented omissions. \\
Cooney bimodal speech\par \citep{cooney2022s0236} & EEG + fNIRS\par N=19 recorded; 15 performed imagined speech & Words and word pairs with multimodal cues & 55 original EEG sessions; 49 paired sessions analyzed & \textbf{Public code; raw release not provided by repository}\par Six fNIRS sessions excluded; further artifact rejection reduces trial counts. \\
Pippi audiovisual iEEG--fMRI\par \citep{dataset_berezutskaya2022_ca8d9} & ECoG/\allowbreak sEEG + fMRI\par N=63 unique: 51 iEEG, 30 fMRI, 18 shared & Watching speech/\allowbreak music film; iEEG rest & 6.00 h main iEEG film; 3.25 h fMRI film & \textbf{Public neural data; film by request}\par Three supplemental HD streams include simultaneous channels and a repeated acquisition. \\
SMN4Lang\par \citep{dataset_wang2022_8999d} & MEG + fMRI\par N=12 in both modalities & Listening to 60 Mandarin stories & 68.09 h MEG; 66.76 h task fMRI & \textbf{Public raw + annotations}\par Separate modality visits; rest scans excluded from the task amounts. \\
Bimodal Inner Speech\par \citep{simistira2023s0308} & EEG + fMRI\par N=4 released; 5 recruited & Cued inner speech in separate recording sessions & 1.51 h EEG; 5.03 h fMRI & \textbf{Public raw + derivatives}\par One participant excluded for poor EEG quality; modality durations differ. \\
BABA\par \citep{dataset_li2025_aa6af} & OPM-MEG + fMRI\par N=60: separate cohorts of 30 & Watching a dialogue-rich social-interaction film & 12.66 h MEG film; 12.67 h fMRI film & \textbf{Public raw + task material}\par Questions, replay and fMRI rest are additional tasks, excluded from film totals. \\
LPP Multi-talker\par \citep{dataset_zhengwuma2025_6c0b0} & EEG + 7 T fMRI\par N=26 in both modalities & Single/\allowbreak competing Mandarin audiobook speech & 19.98 h EEG; 17.56 h task fMRI & \textbf{Public raw + derivatives}\par One EEG file has an incomplete final sample frame; fMRI rest excluded. \\
MEG-SCANS\par \citep{verified_d50_release,habersetzer2025s0579} & MEG; pilot EEG\par N=24; 22 under final protocol & Audiobook listening and speech-in-noise sentences & 18.89 h task MEG; 118 runs & \textbf{Public raw + metadata}\par Chirps remain inside audiobook runs; 2.846 h empty-room noise excluded. \\
Simultaneous Inner Speech\par \citep{simistira2025s0607} & EEG + fMRI\par N=3 & Simultaneous cued English inner speech & 2.19 h EEG; 2.39 h fMRI & \textbf{Public corrected EEG + fMRI; raw EEG by request}\par Five EEG sessions versus six fMRI runs; dummy volumes and interruption affect duration. \\
BCCWJ-Brain\par \citep{dataset_yushisugimoto2026_de0f8,dataset_bccwj_ds007752,dataset_bccwj_ds007763} & fMRI /\allowbreak  EEG /\allowbreak  MEG\par N=36 /\allowbreak  41 /\allowbreak  35 released; separate cohorts & Japanese newspaper word-paced reading & 23.12 /\allowbreak  25.34 /\allowbreak  21.87 h of continuous recordings & \textbf{Public neural data; OpenNeuro}\par Licensed BCCWJ text is separate; modalities use different participants. fMRI events accompany 134 of 144 runs. \\
Hons hybrid phonemes\par \citep{hons2026s0713} & EEG + fNIRS\par N=27 recruited; 22 analyzed & Four perceived and imagined phonemes & 12,709 retained trials; 17.65 h of task windows per modality & \textbf{Public EEG/\allowbreak fNIRS recording files, code and results; OSF}\par Retained task windows; simultaneous modalities. Public EEG deposit includes 24 ID prefixes; 22 participants were analyzed. \\
Defenderfer vocoded/\allowbreak noisy speech\par \citep{dataset_defenderfer2019_c5063,defenderfer2021frontotemporal} & fNIRS\par N=39 recruited; 38 analyzed; 38 released IDs & Listen and repeat sentences in vocoding/\allowbreak noise conditions & 23.24 h continuous release; 6,421 study trials & \textbf{Public data; Mendeley}\par Healthy listeners; paper-derived listening-plus-repetition windows total 10.70 h. \\
Guo tonal monosyllables\par \citep{guo2022s0252} & fNIRS (44 channels; 7.41 Hz)\par N=20 & Imagined Mandarin vowels with four lexical tones & 1,600 trials; 4.44 h of protocol-derived imagery windows & \textbf{Data on reasonable request}\par 80 trials/\allowbreak person; 16 combinations, five repetitions each. Imagery lasts 10 s within 34-s trials. Raw data are not public for legal/\allowbreak ethical reasons; the supplementary channel map is public. \\
Kolisnyk fNIRS narratives\par \citep{dataset_kolisnyk2024_d078f} & fNIRS\par N=30 recruited; 26 analyzed & Intact and scrambled audiovisual/\allowbreak audio narratives & 11.23 h of stimulus windows across retained participants & \textbf{Data on request with ethics agreement; public code}\par Four stimulus conditions; acquisition failures and quality exclusions omitted. \\
MindSpeech\par \citep{suyi2024s0450} & HD-fNIRS\par N=4 analyzed & Cued sentence imagination with typed reports & 2,646 sentence trials; 5.15 h of imagery windows & \textbf{Study corpus; public raw release not identified}\par Seven-second imagery excludes untimed sentence typing. \\
CI/\allowbreak NH multimodal speech fNIRS\par \citep{balint2025multimodalfnirs,dataset_balint2025_a7afc} & fNIRS\par N=72 recorded (46 cochlear implant, 26 normal hearing) & Speech in quiet/\allowbreak noise, audiovisual speech and lipreading & 72 recordings; 10.40 h of stimulus windows & \textbf{Public raw SNIRF archive; Dryad}\par 40 stimulus windows per participant; appointment time excluded. \\
Natural conversations fNIRS\par \citep{dataset_hu2025_efe67} & fNIRS\par N=31 released & Fat talk and control conversations & 31 recordings; 13.87 h continuous fNIRS & \textbf{Public raw and processed signals; OSF}\par Conversation windows total 7.75 h; only one partner wore fNIRS. \\
\end{longtable}
\endgroup

\end{document}